%% file: draft.tex
\documentclass[10pt,twocolumn,letterpaper]{article}

\usepackage{iccv}
\usepackage{times}
\usepackage{epsfig}
\usepackage{graphicx}
\usepackage{amsmath}
\usepackage{amssymb}

\usepackage[nosort,compress]{cite}

\usepackage[bf,small]{caption}

\usepackage{eucal}

\usepackage[leftcaption]{sidecap}

\input{macros.tex}

\iccvfinalcopy

\begin{document}

\title{Contextual Hypergraph Modelling  for Salient Object Detection\thanks{Appearing in Proc.\ International Conference on Computer Vision, Sydney, Australia, 2013.}
}

\author{Xi Li, Yao Li, Chunhua Shen,  Anthony Dick,
Anton van den Hengel\\
The University of Adelaide, Australia\\
}

\maketitle
\begin{abstract}

Salient object detection aims to locate objects
that capture human attention within images.
 Previous approaches often pose
 this as a problem of image contrast
 analysis.
In this work, we model an image as a hypergraph that
utilizes a set of hyperedges to
capture the contextual properties of image pixels or regions.
As a result,
the problem of salient object detection becomes one of
finding salient vertices and hyperedges in the
hypergraph.
The main advantage of hypergraph modeling is that it takes into
account each pixel's (or region's) affinity with its neighborhood
as well as its separation from image background.
Furthermore,
we propose an alternative approach based on center-versus-surround contextual
contrast analysis,
which performs salient object detection by optimizing a cost-sensitive
support vector machine (SVM) objective function.
Experimental results on four challenging datasets
demonstrate the effectiveness of the proposed
approaches against the state-of-the-art approaches to salient object detection.
\end{abstract}

\section{Introduction}

Image saliency detection
aims to effectively identify important and informative
regions in images.
Early approaches in this area focus mainly on
predicting where humans look,
and thus work only on human eye fixation data~\cite{itti1998model,DBLP:conf/nips/BruceT05,DBLP:conf/nips/HouZ08}.
Recently, a large body of work concentrates on
{\em salient object detection}~\cite{achanta2009frequency,WeiWZ012,
FengWTZS11,cheng2011global,liu2011learning,KleinF11,
AlexeDF10,perazzi2012saliency,shen2012unified,jiang2011automatic,
GofermanZT10,ChangLCL11,RahtuKSH10,yangsaliency_cvpr2013}, whose goal is
to discover the most salient and attention-grabbing object in an image.
This has
a wide range of applications such as
image retargeting~\cite{sun2011scale},
image classification~\cite{SharmaJS12},
and image segmentation~\cite{WangXZH11,lu2011salient}.
Because it is difficult to define saliency analytically, the performance of salient object detection
is evaluated on datasets containing
human-labeled bounding boxes or foreground masks.

Salient object detection
is typically accomplished by image contrast
computation, either on a local or a global scale.
estimates the saliency degree of an image region by
computing the contrast against its local neighborhood.
Various contrast measures
have been proposed, including
mutual information~\cite{gao2007discriminant}, incremental coding length~\cite{DBLP:conf/nips/HouZ08}, and
center-versus-surround feature discrepancy~\cite{FengWTZS11, liu2011learning, KleinF11,AlexeDF10,
jiang2011automatic, GofermanZT10, ChangLCL11}.

\begin{figure}[t]
\begin{center}
\begin{tabular}{@{}c@{}c@{}c}
\includegraphics[width=0.3180003\linewidth]{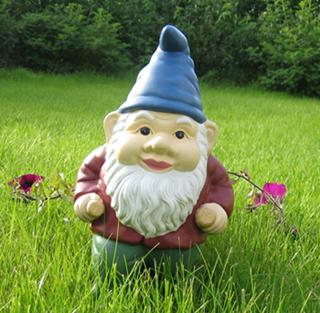} \ &
\includegraphics[width=0.3180003\linewidth]{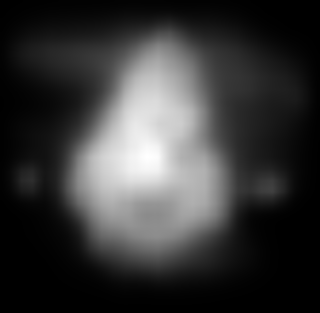} \ &
\includegraphics[width=0.3180003\linewidth]{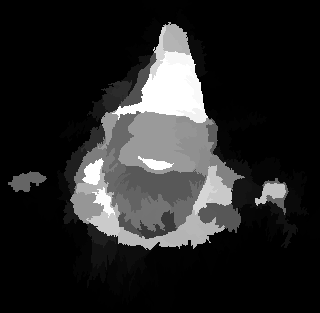}\\
\includegraphics[width=0.3180003\linewidth]{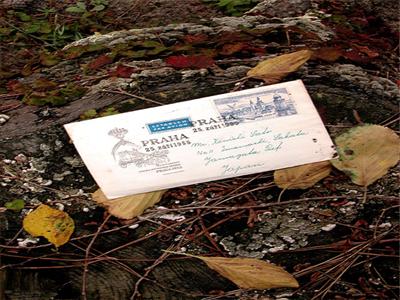} \ &
\includegraphics[width=0.3180003\linewidth]{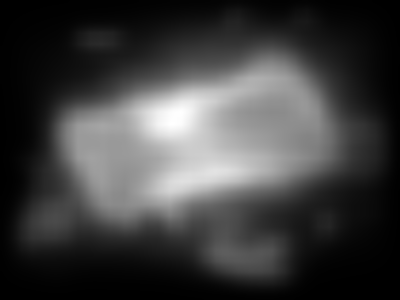} \ &
\includegraphics[width=0.3180003\linewidth]{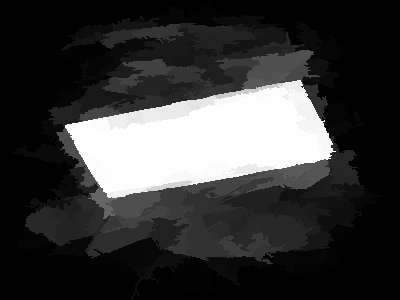}\\
\ {\footnotesize Image} & {\footnotesize SVM saliency} & {\footnotesize Hypergraph saliency}\\
\end{tabular}
\end{center}
\caption{Illustration of our approaches to salient object detection.
}
\label{fig:Intro_Saliency_Example}
\end{figure}

Global salient object detection approaches~\cite{achanta2009frequency,WeiWZ012,
cheng2011global,perazzi2012saliency,shen2012unified}
estimate the saliency of
a particular image region
by measuring its uniqueness
in the entire image.
These approaches model uniqueness by exploiting the
global statistical properties of
the image, including
frequency spectrum analysis~\cite{achanta2009frequency},
color-spatial distribution modeling~\cite{cheng2011global},
high-dimensional Gaussian filtering~\cite{perazzi2012saliency},
low-rank matrix decomposition~\cite{shen2012unified}, and
geodesic distance computation~\cite{WeiWZ012}.

Therefore,
the definition of object saliency depends on the choice of context. Global saliency defines the
context as the entire image, whereas local saliency requires the definition of a local context.
In this work, we first show that within a fixed context, a cost-sensitive SVM can accurately measure saliency by capturing centre-surround contrast information.
We then show that the use of a hypergraph captures more comprehensive contextual information,
and therefore enhances the accuracy of
salient object detection.

Here, we propose two approaches to salient object detection
based on hypergraph modeling and imbalanced max-margin learning.
Our main contributions are as follows.
\begin{enumerate}
\setlength{\itemsep}{-0.86mm}
\item We introduce hypergraph modeling into the process of
image saliency detection for the first time.
A hypergraph is a rich, structured image representation
modeling pixels (or superpixels) by their contexts rather than their individual values.
This additional structural information enables more accurate saliency measurement.
The problem of saliency detection is naturally
cast as that of
detecting salient vertices and hyperedges in a hypergraph
at multiple scales.

\item We formulate the centre-surround contrast approach to saliency
as a cost-sensitive
max-margin classification problem.
Consequently,
the saliency degree of an image region is measured by
its associated normalized SVM coding length.
\end{enumerate}
Example results of our approaches to salient object detection
are shown in Fig.~\ref{fig:Intro_Saliency_Example}.
We describe each approach in the following two sections, before evaluating them in Sec.~\ref{sec:results}.

\section{Cost-sensitive SVM saliency detection}

\begin{figure}[]
\begin{center}
\includegraphics[width=.5\textwidth]{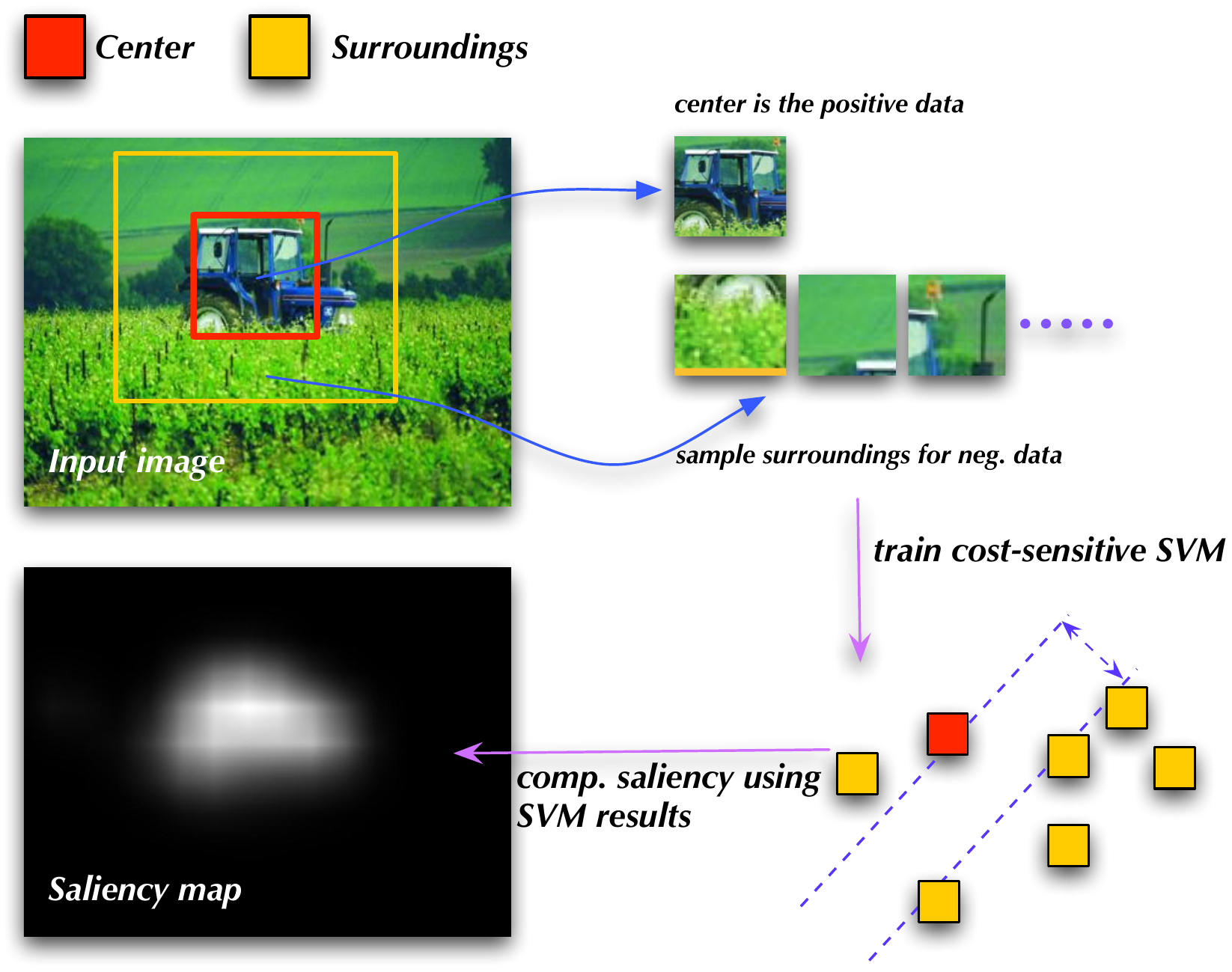}
\end{center}
\caption{Illustration of cost-sensitive SVM for saliency detection.
The saliency score is computed using Equ.~\eqref{eq:local_saliency} based on the SVM classification results.
}
\label{fig:Local_SVM_Saliency}
\end{figure}

\begin{SCfigure*}
\centering
\includegraphics[width=.65\textwidth]{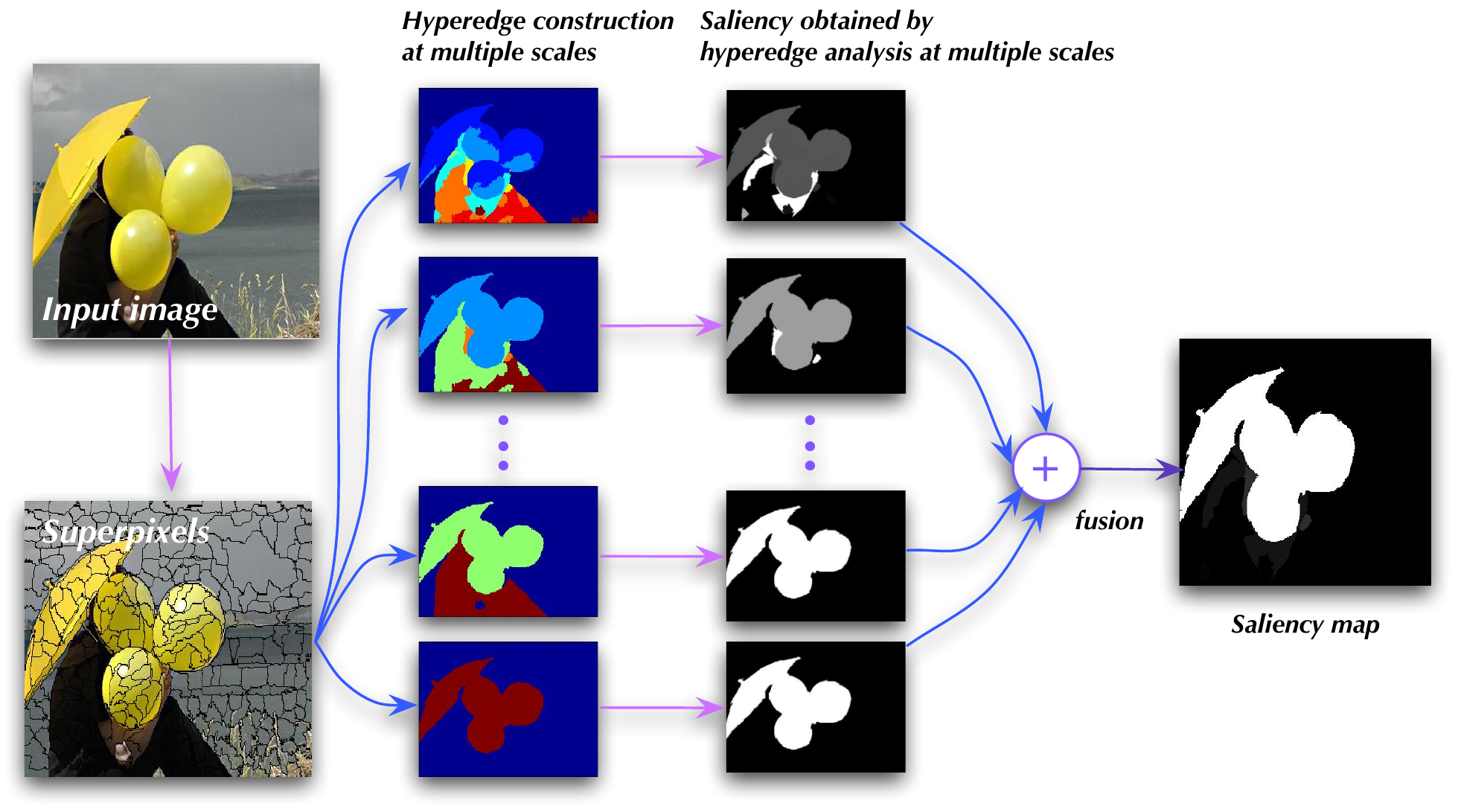}
\caption{Illustration of hypergraph modeling for  saliency  detection
using nonparametric clustering.
The first column shows an input image and its associated over-segmented
image with a set of superpixels.
    The middle columns display the multi-scale hyperedges (constructed
    by nonparametric clustering on the superpixels) and their corresponding results of
    hyperedge saliency evaluation.  The rightmost image shows the
    final
    saliency map $HSa$ generated by multi-scale hyperedge
    saliency fusion.  Note that the regions highlighted in different
    colors correspond to different hyperedges (i.e., superpixel cliques
    having some common visual properties). In theory, our hypergraph modeling
    can also work on pixels in a similar way.
}
\label{fig:Hyperedge_ms}
\end{SCfigure*}

As illustrated in~\cite{KleinF11,hubel1965receptive},
saliency detection is typically
posed as the problem of center-versus-surround contextual contrast analysis.
To address this problem, we propose a saliency detection method based on imbalanced
max-margin learning, which is capable of effectively
discovering the local salient image
regions that significantly differ from their
surrounding image regions.
In this case, the image is divided into
overlapping rectangular windows which are tested
for saliency. The context for each window is the windows that overlap it.

Before describing the method, we first introduce some notation
used hereinafter.
Let $ \bx_1 $
denote the feature vector associated with a center image patch,
and
$\{\bx_{\ell}\}_{\ell=2\dots N}$
denote the feature vectors associated with
the spatial overlapping surrounding patches of the
center image patch.
Using these patches,
the proposed method explores their
 inter-class separability in a max-margin
classification framework.

As shown in the top-right part of Fig.~\ref{fig:Local_SVM_Saliency},
the center image patch $\bx_1$ is
thought of as a positive sample
while the surrounding patches
$\{\bx_{\ell}\}_{\ell=2\dots N}$ are used
as the negative samples.
The saliency degree of $\bx_1$
is determined by its inter-class separability
from $\{\bx_{\ell}\}_{\ell=2\dots N}$.
In other words, if $\bx_1$ could be easily
separated from $\{\bx_\ell\}_{\ell=2\dots N}$,
then it is deemed to be salient;
otherwise, its saliency degree is low.
This is a binary classification problem,
which is associated with a cost-sensitive
classification objective function~\cite{suykens2002weighted}:
\begin{equation}
\begin{array}{cc}
\underset{\bw, b, \bepsilon}{\min} & J(\bw, b, \boldsymbol \epsilon) =
\frac{1}{2}\|\bw\|_{2}^{2} + \frac{1}{2}C\sum_{\ell=1}^{N}\nu_{ \ell
}\epsilon_{ \ell }^{2},\\
\mbox{s.t.} & y_{\ell} = f(\bx_{\ell}) + \epsilon_{\ell},
\end{array}
\label{eq:ls_svm}
\end{equation}
where
$\|\cdot\|_{2}$ is the $L_2$ norm,
$f(\bx) = \bw^{\T}\bx + b$ is the classifier
to learn; $\bepsilon $ is the
residual vector;
$C$ is the regularization parameter;
and $\nu_{\ell}$ is the corresponding weight of $\bx_{\ell}$
such that $\nu_{1}\gg \nu_{\ell}$ for $ \ell= 2 \dots N $.
We set all the negative samples to have the same weight
$ \nu_\ell $, $ \ell = 2 \dots N$.
According to the KKT condition, we have the following
linear system:
\begin{equation}
\left[
\begin{array}{cc}
0 & \bone^{\T}_{N}\\
\bone_{N} & \Omega + V_{C}
\end{array}
\right]
\left[
\begin{array}{c}
b\\
\balpha
\end{array}
\right]=
\left[
\begin{array}{c}
0\\
\by
\end{array}
\right],
\label{eq:w_ls_svm_linear_system}
\end{equation}
where $\bone_{N}\in \mathcal{R}^{N}$ is
the all-one column vector,
$\by = (y_{1}, y_{2}, \ldots, y_{N})^{\T}$ is the label
vector, $\Omega = (\Omega_{ij})_{N\times N}$ is the kernel matrix
$\Omega_{ij} = \bx_{i}^{\T}\bx_{j}$, and
$V_{C}$ is a diagonal matrix such that
$V_{C} = \mbox{diag}(\frac{1}{C\nu_{1}}, \frac{1}{C\nu_{2}}, \ldots, \frac{1}{C\nu_{N}})$.
Based on the solution $(\balpha^{\ast}, b^{\ast})$ to the linear system~\eqref{eq:w_ls_svm_linear_system},
we have the weighted LS-SVM classifier $f(\bx) = (\bw^{\ast})^{\T}\bx + b^{\ast}$
with
$\bw^{\ast} = (\bx_{1}, \bx_{2}, \ldots, \bx_{N})\balpha^{\ast}$.

Using the weighted LS-SVM classifier $f(\bx)$, we define the
saliency score as:
\begin{equation}
SSa(\bx_{1}) = \frac{1}{N-1}\sum_{ \ell =2}^{N}\frac{1 -
\mbox{sgn}(f(\bx_{ \ell }))}{2},
\label{eq:local_saliency}
\end{equation}
where $\mbox{sgn}(\cdot)$ is a sign function and the term $\sum_{\ell=2 } ^ N   \frac{1 - \mbox{sgn}(f(\bx_{ \ell }))}{2}$ counts
the number of correctly classified surrounding samples.
Loosely speaking,
the saliency score $SSa(\bx_{1})$ can be viewed as
a normalized SVM coding length (i.e., training accuracy for the surrounding samples), which characterizes the inter-class separability
between $\bx_{1}$ and its surroundings $\{\bx_{ \ell }\}_{ \ell =2
\dots N}$.
As shown in the bottom-left part of Fig.~\ref{fig:Local_SVM_Saliency},
the harder $\bx_{1}$ is to separate from
$\{\bx_{ \ell }\}_{ \ell =2 \dots N}$, the smaller $SSa(\bx_{1})$ will be.
    In other words, the center patch looks similar to its
    surroundings.
Conversely, the larger $SSa(\bx_{1})$ indicates the lower
similarity between $\bx_{1}$  and $\{\bx_{ \ell }\}_{ \ell =2 \dots N}$,
and hence a higher saliency degree.
Note that, here the cost-sensitive LS-SVM is not the only choice.
We can use other classifiers such as the exemplar SVM
\cite{malisiewicz2011ensemble}, where the standard hinge-loss SVM is
used. We have used LS-SVM for its simplicity (it has a closed-form
solution). Note that, this max-margin learning framework can
be easily extended to perform saliency detection on
a global scale. Namely, the image boundary patches
can be treated as negative samples while the remaining
image patches are used as positive samples.
By running the max-margin learning procedure over
such training samples, the saliency degree of each image patch
can be measured by computing its distance
to the separating hyperplane.

Example saliency maps derived from this measure are
shown in Figs.~\ref{fig:Intro_Saliency_Example} and~\ref{fig:Local_SVM_Saliency}.
Although they accurately locate the salient object in each case,
they also suffer from ``fuzziness'' or lack of precision around
object boundaries and in locally homogeneous regions. This is mainly
due to the center-surround local context that they are based on.
In the next section, we describe an alternative approach based on segmentation based context that alleviates these problems.

\begin{figure}[t]
\begin{center}
\begin{tabular}{@{}c@{}c@{}c}
\includegraphics[width=0.33\linewidth]{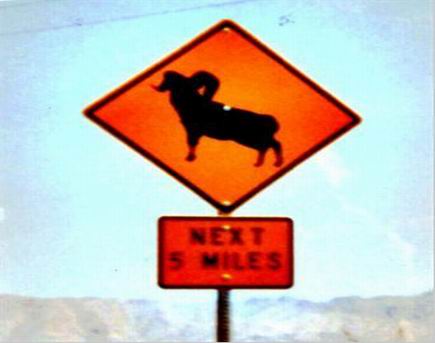} \ &
\includegraphics[width=0.33\linewidth]{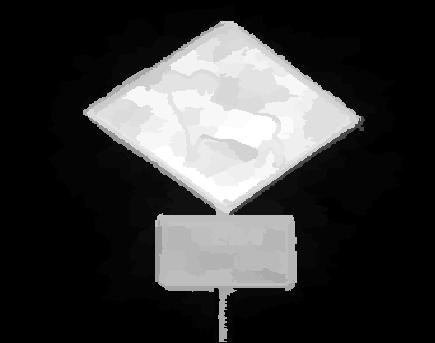} \ &
\includegraphics[width=0.33\linewidth]{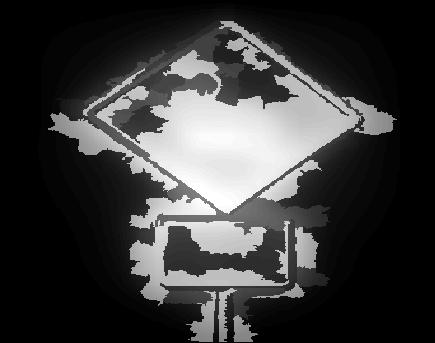}\\
\ {\footnotesize Image} & {\footnotesize Hypergraph saliency} & {\footnotesize Standard graph saliency}\\
\end{tabular}
\end{center}
\caption{Illustration of salient object detection using two different types of graphs
(i.e., hypergraph and standard pairwise graph). Clearly, our hypergraph saliency
measure is able to accurately capture the intrinsic structural properties of
the salient object.
}
\label{fig:graph_comparison}
\end{figure}

\section{Hypergraph modeling for saliency  detection}
\label{sec:hypergraph_modeling}

To more effectively find salient object regions,
we propose a hypergraph modeling based saliency detection method that
forms contexts of superpixels to capture both internal consistency and external separation.
Fig.~\ref{fig:Hyperedge_ms} shows the high-level flowchart of the
proposed method.

As illustrated in~\cite{zhou2007learning}, a
hypergraph is a graph comprising
a set of vertices and hyperedges.
In contrast to the pairwise edge in a standard graph,
the hyperedge in a hypergraph is a high-order edge
associated with
a vertex clique linking more than two vertices.
Effectively constructing such hyperedges is crucial for
encoding the intrinsic contextual information
on the vertices in the hypergraph.

\paragraph{Hypergraph modeling}
In our method, an image $I$ is modeled by a hypergraph
$\mathcal{G}=(\mathcal{V},\mathcal{E})$,
where $\mathcal{V}=\{v_{i}\}$ is the vertex set corresponding to the
image superpixels and
$\mathcal{E} = \{e_{j}\}$ is the hyperedge set comprising a family of subsets of $\mathcal{V}$
such that $\bigcup_{e \in \mathcal{E}} = \mathcal{V}$~\cite{zhou2007learning}.
As shown in Fig.~\ref{fig:Hyperedge_ms}, these hyperedges are constructed by
multi-scale clustering, which
groups the image
superpixels
into a set of superpixel cliques.
Each clique corresponds to
a collection of superpixels having some
common visual properties,
and works as a hyperedge of the hypergraph $\mathcal{G}$.
The process of hyperedge construction
implicitly encodes intrinsic affinity information
on superpixels.
Namely, if two superpixels have
a higher co-occurrence frequency in the
hyperedges, they tend to share
more visual properties and have
a higher visual similarity.

A hyperedge can also be viewed as
a high-order context that enforces the contextual constraints on
each superpixels in the hyperedge.
As a result, the saliency of each superpixel, as measured by the hyperedges it belongs to,
is not only determined by the superpixel itself
but also influenced by its associated contexts.
Due to such contextual constraints on
each superpixel, we simply convert
the original saliency detection problem
to that of detecting
salient vertices and hyperedges in the hypergraph $\mathcal{G}$.
Mathematically, the hypergraph $\mathcal{G}$ is associated with a $|\mathcal{V}|\times |\mathcal{E}|$
incidence matrix $\bH=(H(v_{i}, e_{j}))_{|\mathcal{V}|\times
|\mathcal{E}|}$:
\begin{equation}
H(v_{i}, e_{j}) =
\left\{
\begin{array}{ll}
1, & \mbox{if} \thickspace v_{i}\in e_{j},\\
0, & \mbox{otherwise}.
\end{array}
\right.
\label{eq:pairwise_hypergraph_incidence_matrix}
\end{equation}
The saliency value of any vertex $v_{i}$ in $\mathcal{G}$
is defined as: %
\begin{equation}
HSa(v_{i}) = \sum_{e \in \mathcal{E}} \Gamma(e)H(v_{i}, e),
\label{eq:hypergraph_saliency_degree}
\end{equation}
where $\Gamma(e)$ encodes the saliency information on
the hyperedge $e$.
In essence, our hypergraph saliency measure~\eqref{eq:hypergraph_saliency_degree}
is a generalization of the standard pairwise saliency measure
defined as:
\begin{equation}
PSa(v_{i}) = \sum_{v_j \in \mathcal{N}_{v_i}}\hspace{-0.2cm}  d_{(v_i,v_j)} = \hspace{-0.42cm} \sum_{e \in \{(v_i,
v_j)|j \neq i\}} \hspace{-0.65cm}  \mathbb{I}_{e}d_{e}H(v_{i}, e), \hspace{-0.2cm}
\end{equation}
where $\mathcal{N}_{v_i}$ stands for the neighborhood
of $v_i$,
$d_{(v_i,v_j)}$ measures the saliency degree
of the pairwise edge $(v_i,v_j)$,
and $\mathbb{I}_{e}$
is the pairwise adjacency indicator
(s.t. $\mathbb{I}_{e} = 1$ if $v_j \in \mathcal{N}_{v_i}$;
otherwise, $\mathbb{I}_{e} = 0$).
Instead of using simple pairwise edges, our hypergraph saliency measure
takes advantage of the higher-order hyperedges (i.e., superpixel cliques) to effectively
capture the intrinsic structural properties of the salient object, as shown in Fig.~\ref{fig:graph_comparison}.
To implement this approach, we need to address the following two key issues: 1) how to adaptively construct the hyperedge set $\mathcal{E}$;
and 2) how to accurately measure the saliency
degree $\Gamma(e)$ of each hyperedge.

\paragraph{Adaptive hyperedge construction}
A hyperedge in the hypergraph $\mathcal{G}$ is actually a superpixel clique whose elements
have some common visual properties. To capture the
hierarchial visual saliency information, we construct a set of hyperedges
by adaptively grouping the superpixels according to
their visual similarities at multiple scales. In theory,
this can be carried out in many ways using any number of established segmentation
and clustering techniques.

\begin{figure*}[t]
\begin{center}
\includegraphics[width=0.6769\linewidth]{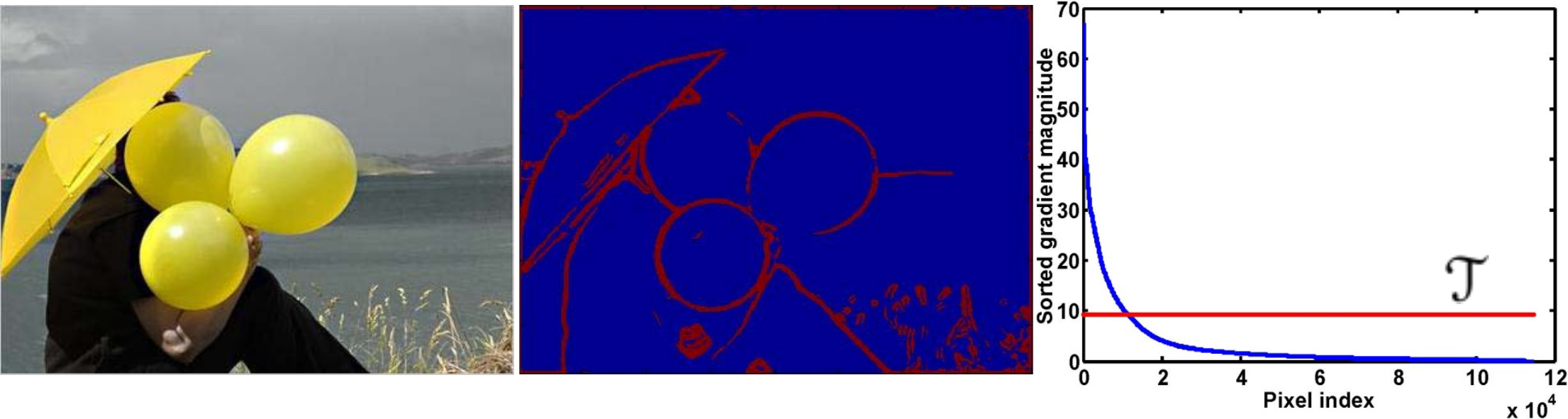}
\end{center}
\caption{Illustration of the gradient magnitude information for hyperedge
saliency evaluation. The left subfigure shows the original image, and
the middle subfigure displays the gradient magnitude map $I_{g}^{\ast}$
obtained by binarizing $I_{g}$ using the adaptive threshold $\mathcal{T}$,
as illustrated in the right subfigure.
}
\label{fig:Hyperedge_gradient_maps}
\end{figure*}

\emph{Non-parametric clustering}
is typically associated with a kernel density estimator:
\begin{equation}
\widehat{f}_{k}(\bp) = \frac{C_{k}}{Q|\bSigma|^{\frac{1}{2}}}\sum_{i=1}^{Q}k(M^2(\bp, \bp_{i}, \bSigma)),
\end{equation}
where $\bp_{i}$ is a feature vector associated with the $i$-th superpixel (generated
from image oversegmentation),
$k(\cdot)$ is a kernel profile ($k(x) = \exp(-x/2)$ in our case),
$\bSigma$ is a symmetric positive definite bandwidth matrix
(in the experiments, $\bSigma = \gamma^{2}\bI$ with $\gamma$ being a scaling factor
and $\bI$ being an identity matrix),
$M^2(\bp, \bp_{i}, \bSigma) = (\bp - \bp_{i})^{\T}\bSigma^{-1}(\bp - \bp_{i})$
stands for the Mahalanobis distance,
and $C_{k}$ is a normalization constant.
Therefore, the superpixel cliques can be discovered by seeking
the modes of $\widehat{f}_{k}(\bp)$.
Mathematically, the mode-seeking problem can
be converted to that of locating the zeros
of the gradient $\nabla \widehat{f}_{k}(\bp) = 0$,
which leads to the following iterative procedure:
\begin{equation}
\bp^{t+1} = \frac{\sum_{i=1}^{Q}g(M^2(\bp^{t}, \bp_{i}, \bSigma))\bp_{i}}{\sum_{i=1}^{Q}g(M^2(\bp^{t}, \bp_{i}, \bSigma))},
\label{eq:ms_iteration}
\end{equation}
where $g(x) = -k'(x)$ and the superscript $t$ indexes the iteration number.
To accelerate the optimization process~\eqref{eq:ms_iteration},
we adopt a fast agglomerative mean-shift clustering method based on
iterative query set compression~\cite{yuan2012agglomerative}.

Each mode is associated with a hyperedge,
containing all the superpixels that
converge to it after running
the iterative procedure~\eqref{eq:ms_iteration}.
The bandwidth matrix $\bSigma = \gamma^{2}\bI$ controls the
scaling properties of the hyperedge.
Consequently, using different values of $\gamma$ for nonparametric
clustering can
generate the hyperedges at different scales, as shown in Fig.~\ref{fig:Hyperedge_ms}.
By using different configurations of $\gamma$,
we obtain
a set of multi-scale
hyperedges $\{e_{i}\}$
with $e_{i}$ being the $i$-th
hyperedge.

\begin{figure}[t]
\vspace{-0.0cm}
\begin{center}
\includegraphics[width=0.99\linewidth]{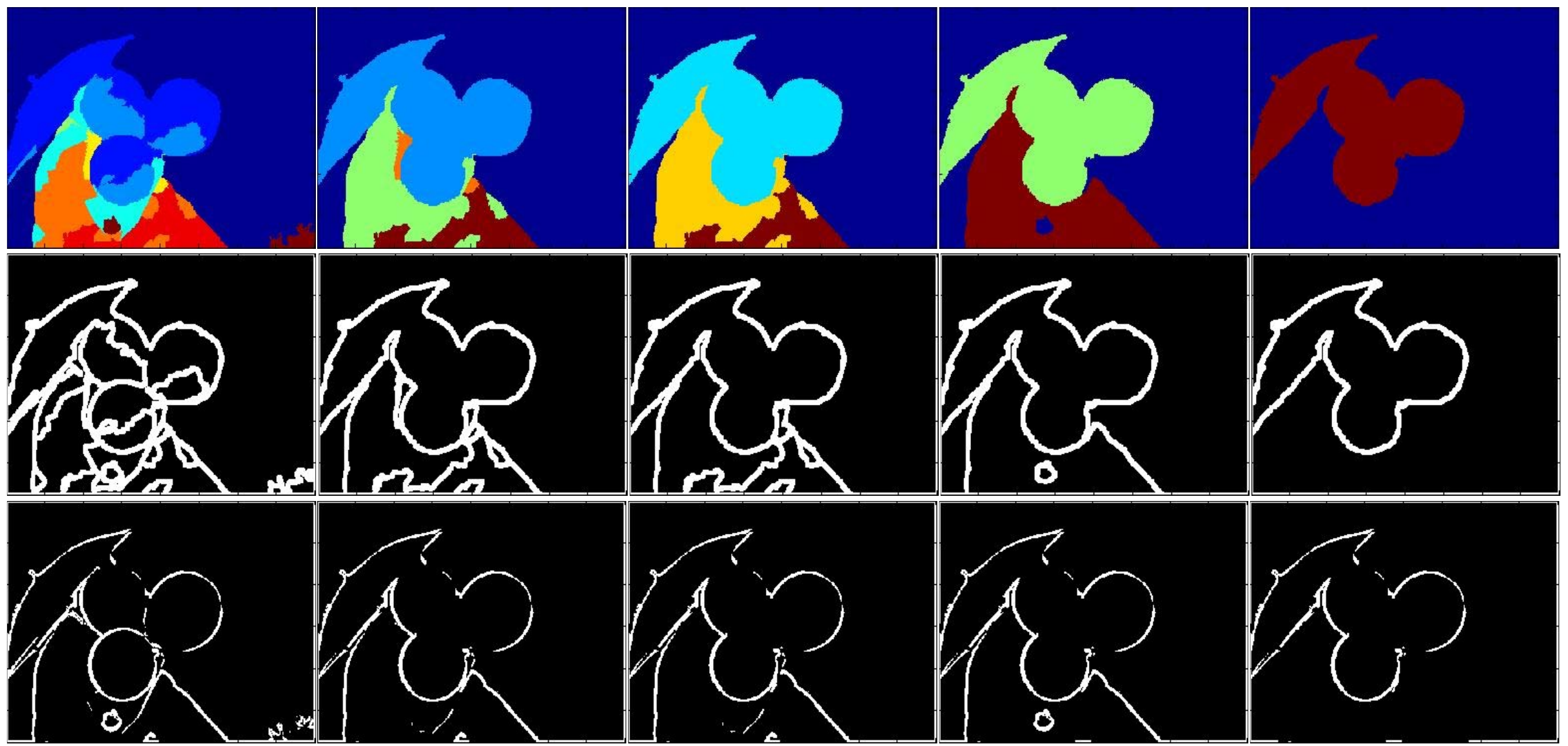}
\end{center}
\caption{Illustration of $M_{g}$ and $I_{g}^{\ast}\circ M_{g}$ for hyperedge saliency evaluation.
The top row shows the multi-scale hyperedges;
the middle row displays
the scale-specific $M_{g}$ that
indicates the pixels (within a narrow band) along the boundary of
the scale-specific hyperedge;
and the bottom row exhibits the filtered gradient magnitude map $I_{g}^{\ast}\circ M_{g}$.
}
\label{fig:Hyperedge_binary_mask}
\end{figure}

\begin{figure*}[t]
\vspace{0.05cm}
\begin{center}
\begin{tabular}{@{}c@{}c@{}c@{}c}
\includegraphics[width=4.2cm,height=3.15cm]{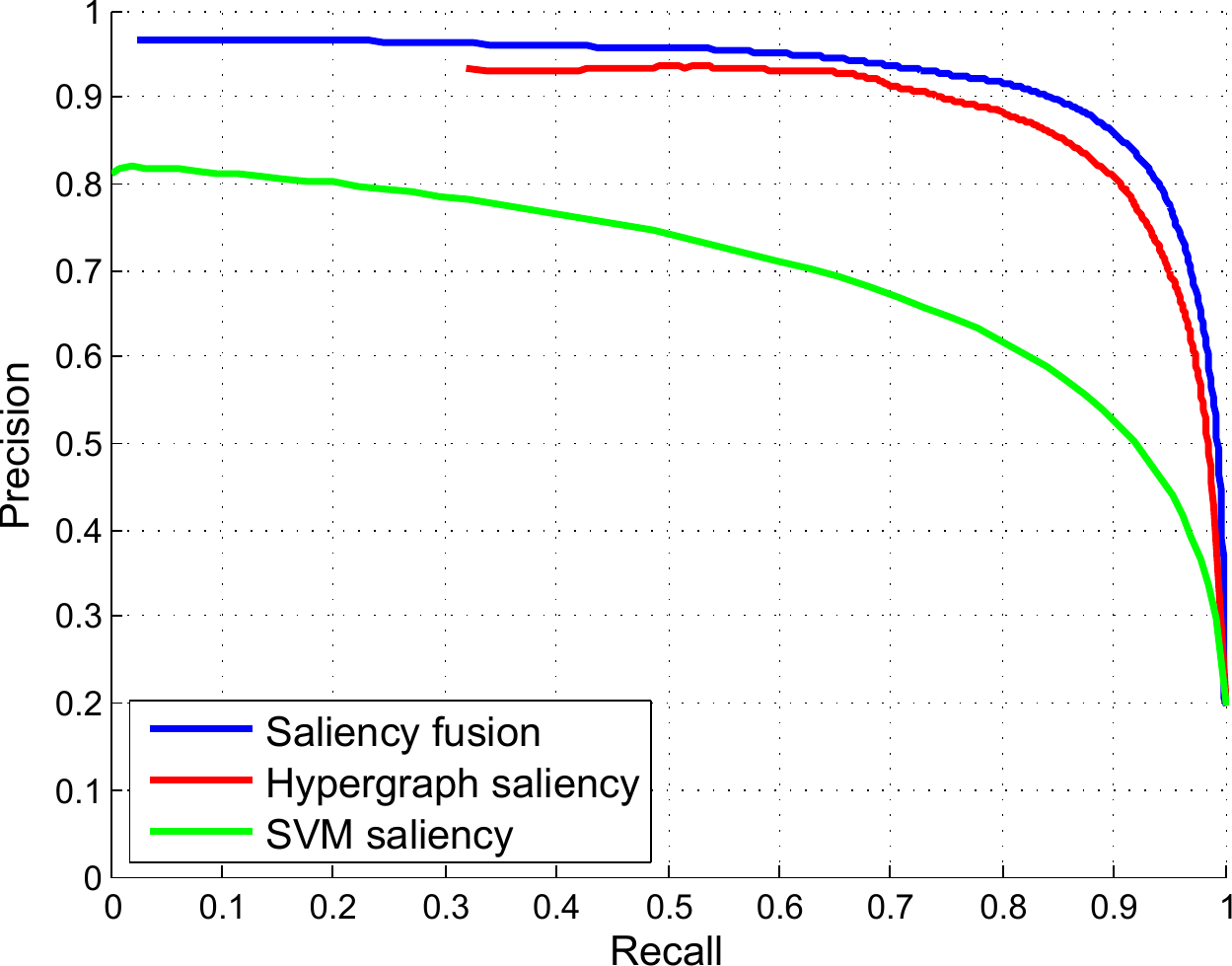} \ &
\includegraphics[width=4.2cm,height=3.15cm]{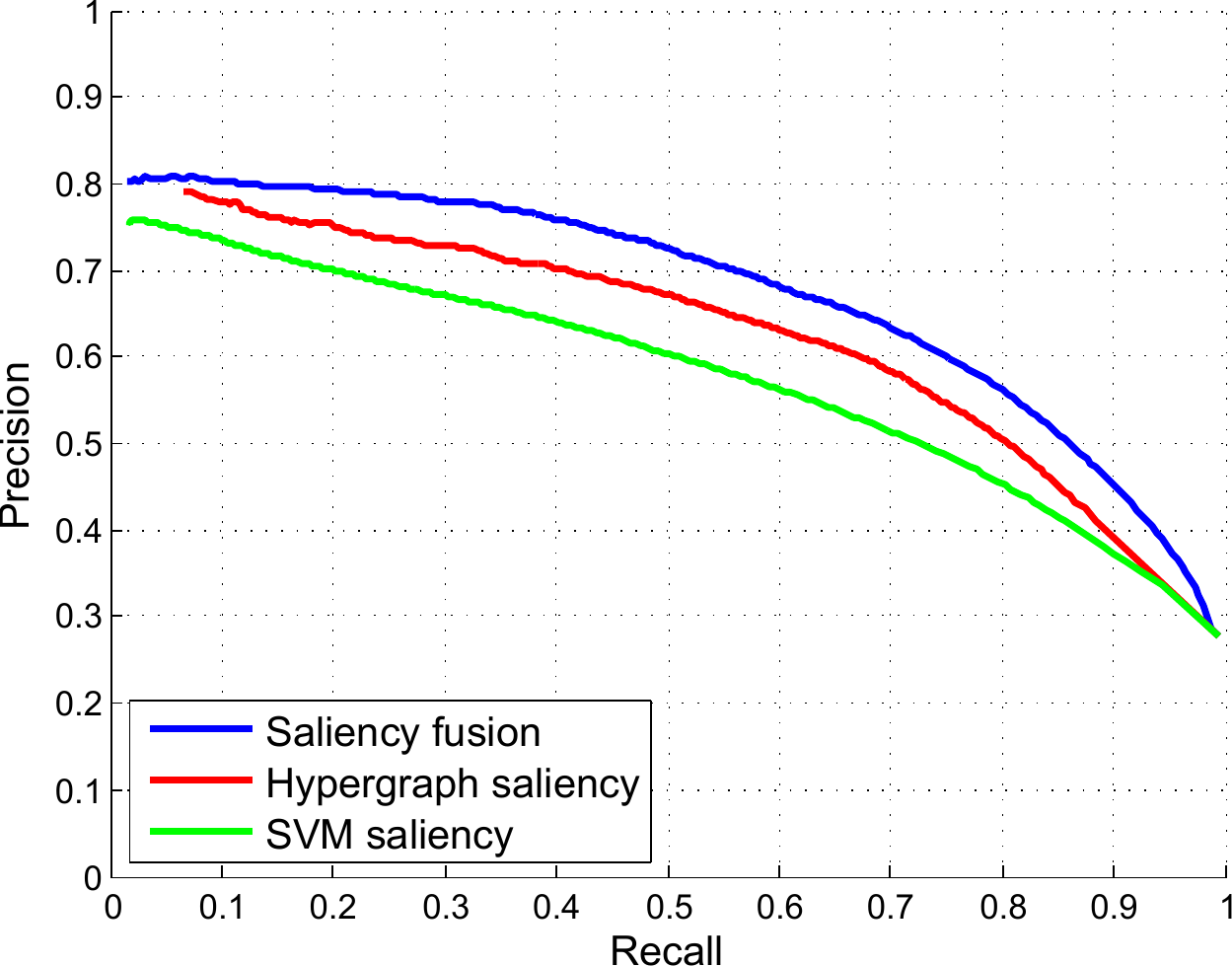} \ &
\includegraphics[width=4.2cm,height=3.15cm]{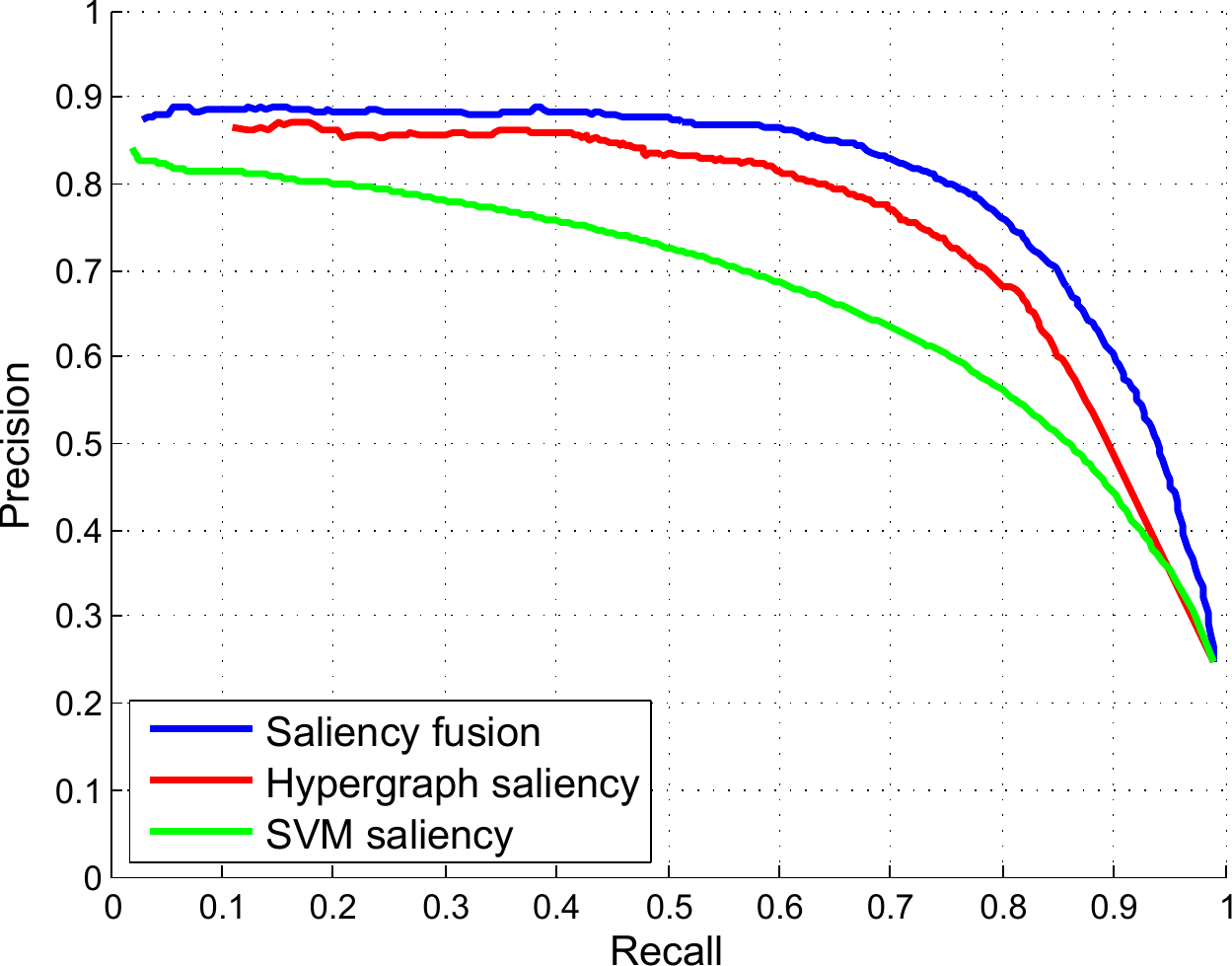} \ &
\includegraphics[width=4.2cm,height=3.15cm]{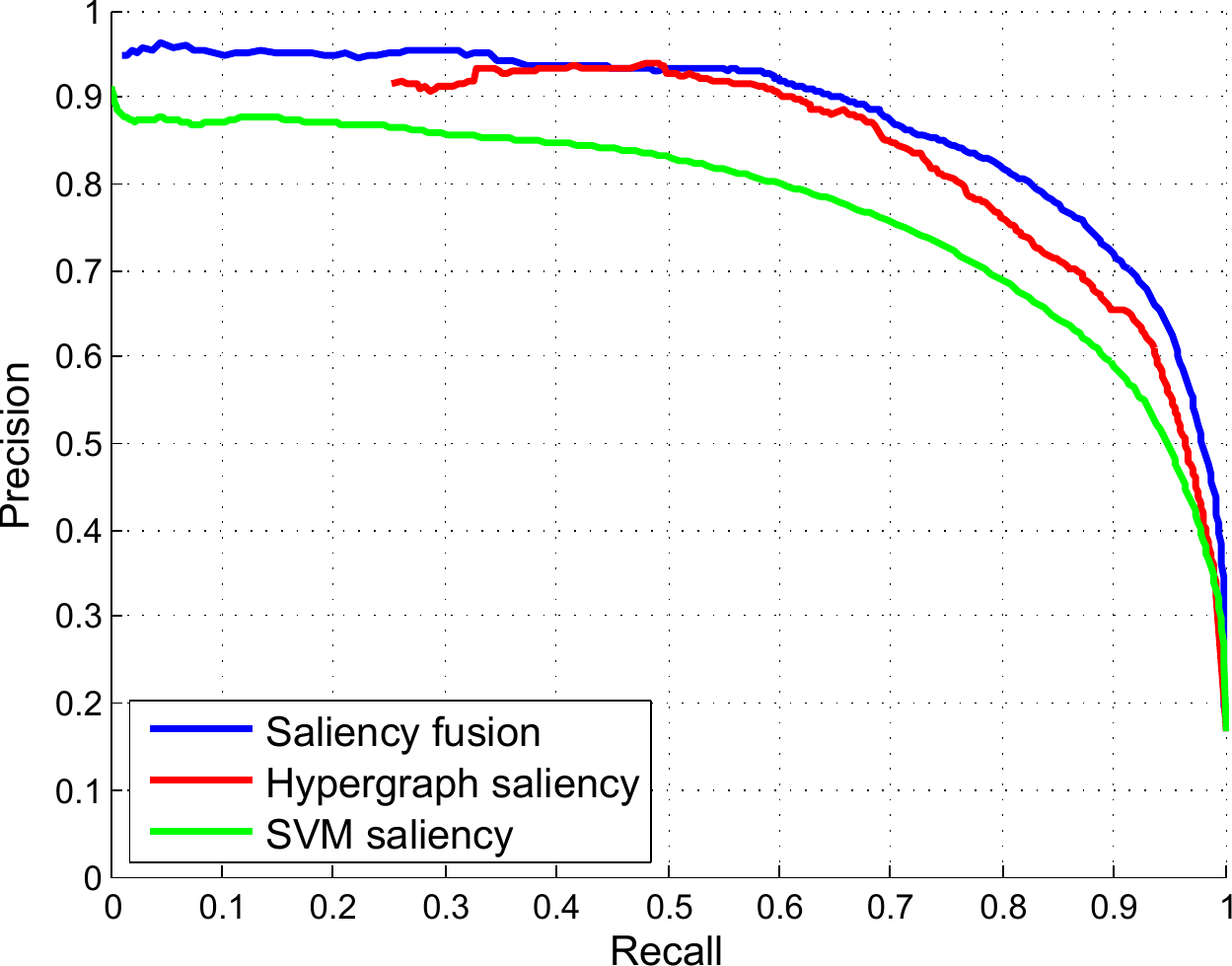} \ \\
\end{tabular}
\end{center}
\caption{PR curves based on three different configurations:
1) using the SVM saliency approach only; and 2) using the hypergraph saliency approach only; 3)
combining the SVM and hypergraph saliency approaches.
Clearly, the saliency detection performance of using the third configuration
outperform that of using the first and second configurations.
From left to right: MSRA-1000, SOD, SED-100, and Imgsal-50.
}
\label{fig:Local_vs_Global} \end{figure*}

\paragraph{Hyperedge saliency evaluation}
By construction, a hyperedge defines a group of pixels that is internally consistent. In addition, a salient hyperedge should
have the following two properties: 1) it
should be enclosed by strong image
edges; and 2) its intersection with the image boundaries
ought to be small~\cite{WeiWZ012,jiang2011automatic}.
Therefore, we measure the saliency degree of
a scale-specific hyperedge $e$ by summing up
the corresponding gradient magnitudes
of the pixels (within a narrow band) along the boundary of
the hyperedge. If the hyperedge  touches
the image boundaries, we decrease
its saliency degree by a
penalty factor.

More specifically,
edge detection (using the Sobel operator in our case) is carried out
for image $I$. Let $I_{x}$ and $I_{y}$ denote
the x-axis and y-axis
gradient magnitude maps, respectively.
Thus, the final gradient magnitude
map $I_{g}$ has the following entry:
$I_{g}(m,n) = \sqrt{I_{x}^{2}(m,n) + I_{y}^{2}(m,n)}$.
To obtain a robust gradient map,
we introduce the following criterion:
$I_{g}^{\ast}(m, n) = 1$ if $I_{g}(m, n)> \mathcal{T}$;
otherwise, $I_{g}^{\ast}(m, n) = 0$, as shown in Fig.~\ref{fig:Hyperedge_gradient_maps}.
Here, $\mathcal{T}$ is a threshold
(picking out the top 10\%
of the $I_{g}$'s elements in our case).
As a result, the saliency value of
the hyperedge $e$ is computed as:
\begin{equation}
\Gamma(e) = \omega_{e}\left[\|I_{g}^{\ast}\circ M_{g}(e)\|_{1} - \rho(e)\right].
\label{eq:shape_border_saliency} \end{equation}
Here,
$\omega_{e}$ is a scale-specific hyperedge weight
(a larger scale leads to a larger weight),
$\|\cdot\|_{1}$ is the 1-norm,
$M_{g}(e)$ is a binary mask (illustrated in Fig.~\ref{fig:Hyperedge_binary_mask})
indicating the pixels (within a narrow band) along the boundary of
the hyperedge $e$,
$\circ$ is the elementwise dot product operator,
and $\rho(e)$ is a penalty factor
that is equal to the number of
the image boundary pixels shared
by the hyperedge $e$.
Based on  Equ.~\eqref{eq:hypergraph_saliency_degree},
we obtain the hypergraph saliency measure
$HSa(v_{i})$ for any vertex $v_{i}$ in the hypergraph $\mathcal{G}$.

After both SVM and hypergraph saliency detection, we obtain
the corresponding saliency maps.
Each element of these saliency maps is mapped into
[0, 255] by linear normalization, leading to
the normalized saliency maps.
Finally, the final saliency map
is obtained by linearly
combining the SVM and hypergraph saliency detection results.

\section{Experiments}
\label{sec:results}

\subsection{Experimental setup}

\paragraph{Datasets}
As a subset of the MSRA dataset~\cite{liu2011learning},
MSRA-1000~\cite{achanta2009frequency} is a commonly used benchmark dataset
for salient object detection.
SOD~\cite{movahedi2010design} is composed of
300 challenging images.
SED-100 is a subset of the SED dataset~\cite{alpert2007image,BorjiSI12},
and consists of 100 images. Each image in SED-100 contains only
one salient object.
Imgsal-50 is a subset of the Imgsal dataset~\cite{li2012visual},
and comprises 50 images with
large salient objects for evaluation.
Each image in the aforementioned datasets contains a human-labelled
foreground mask used as ground truth for salient object detection.

\begin{figure*}[t]
\vspace{-0.0cm}
\begin{center}
\begin{tabular}{@{}c@{}c@{}c@{}c}
\includegraphics[width=4cm,height=3cm]{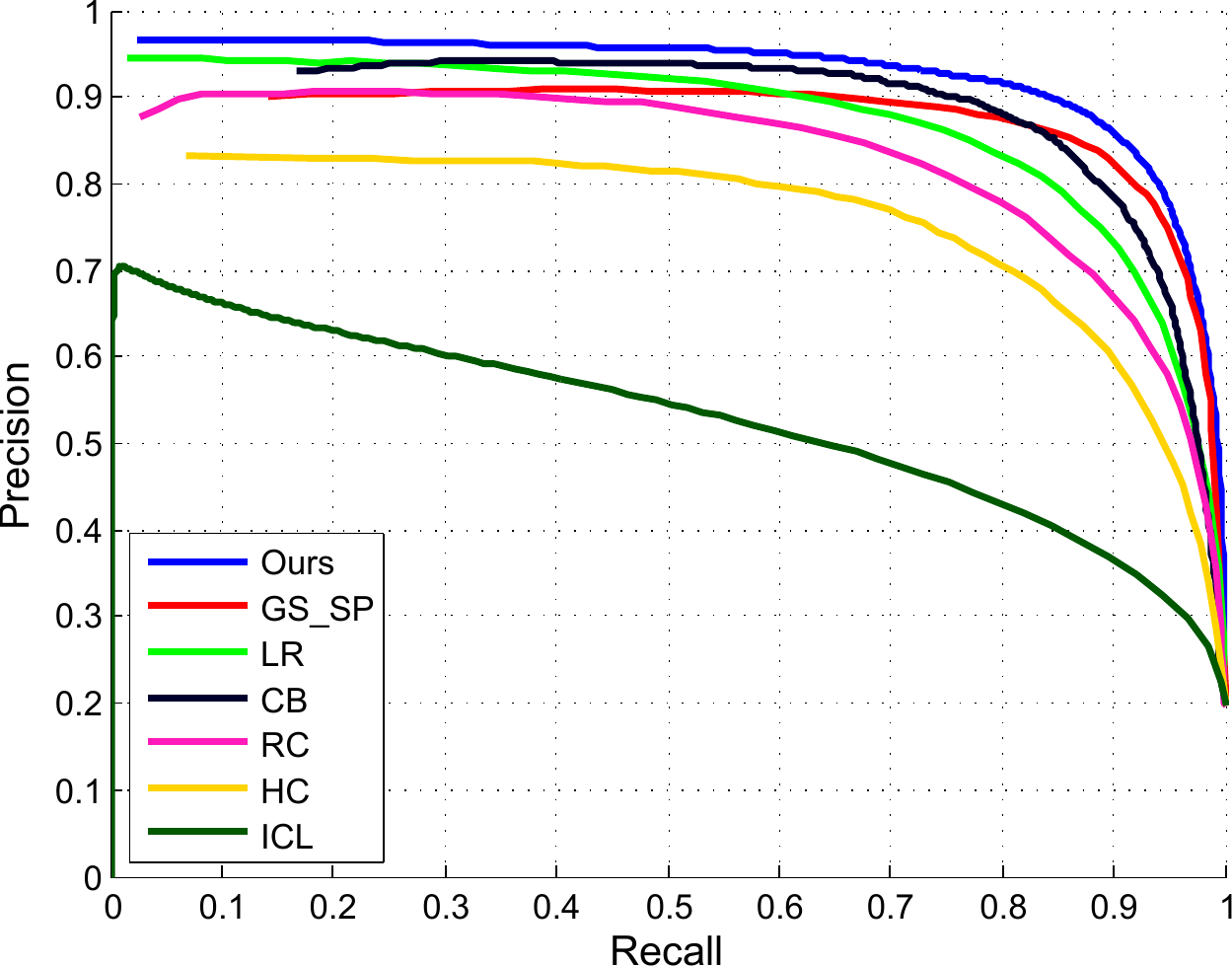} \ &
\includegraphics[width=4cm,height=3cm]{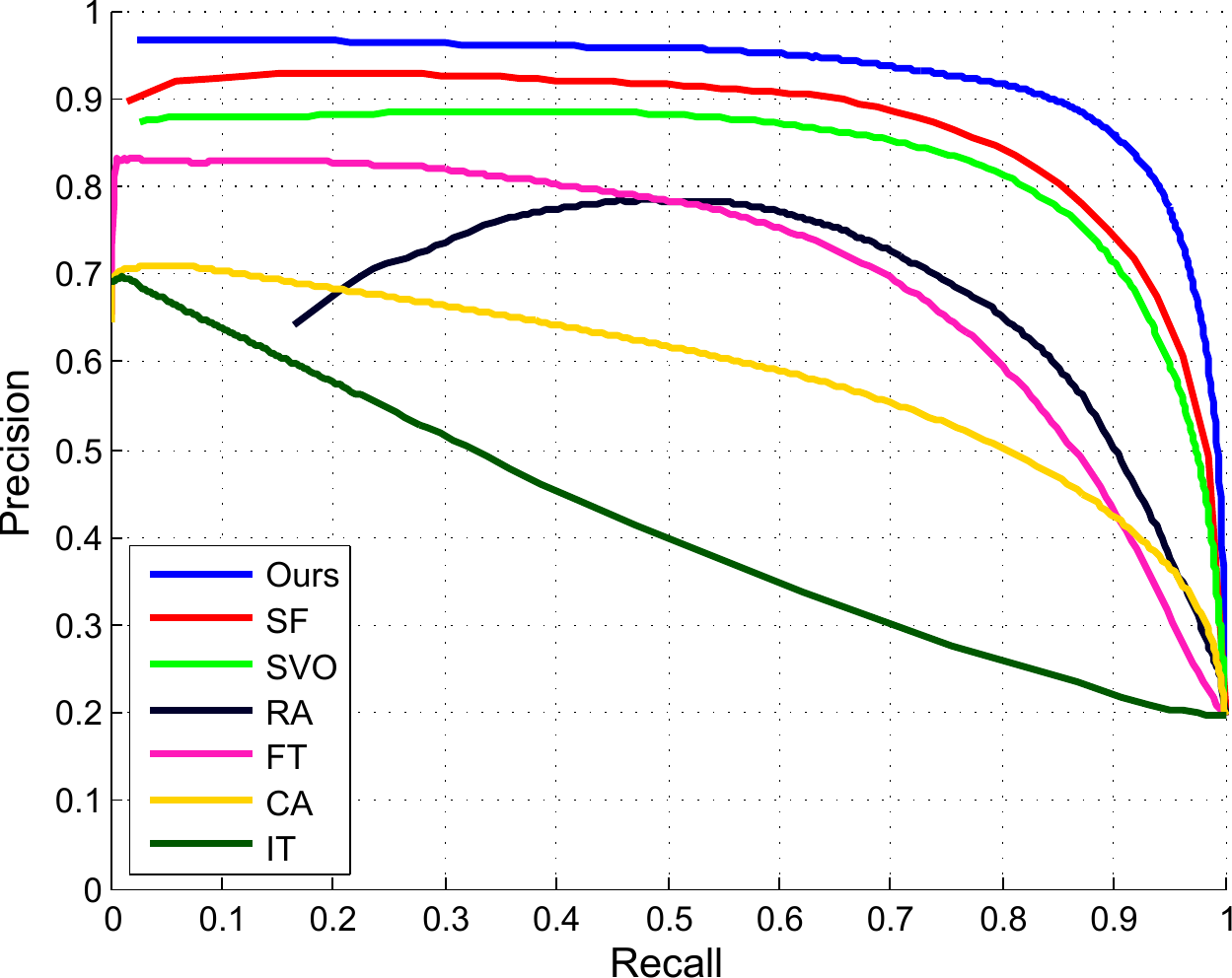} \ &
\includegraphics[width=4cm,height=3cm]{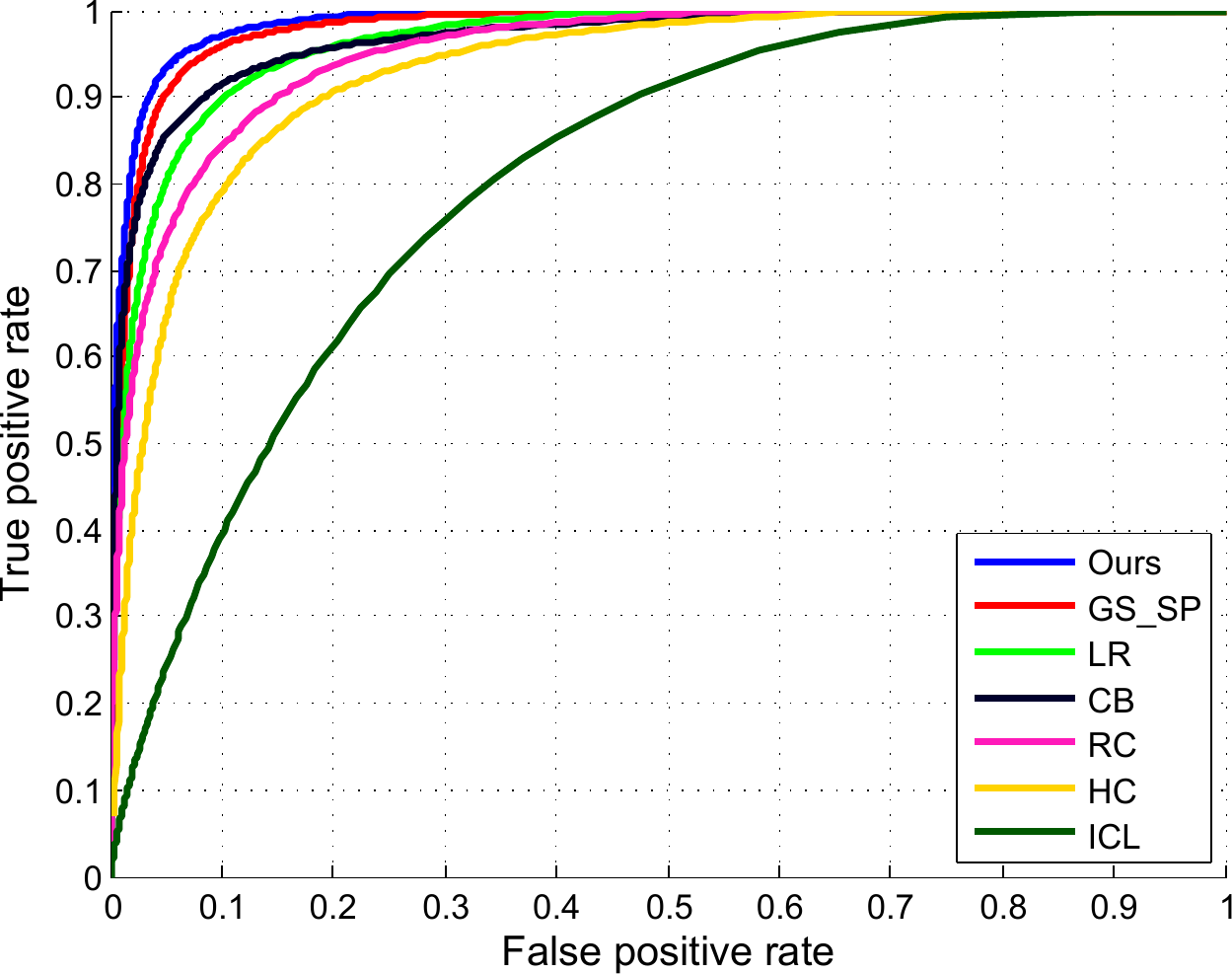} \ &
\includegraphics[width=4cm,height=3cm]{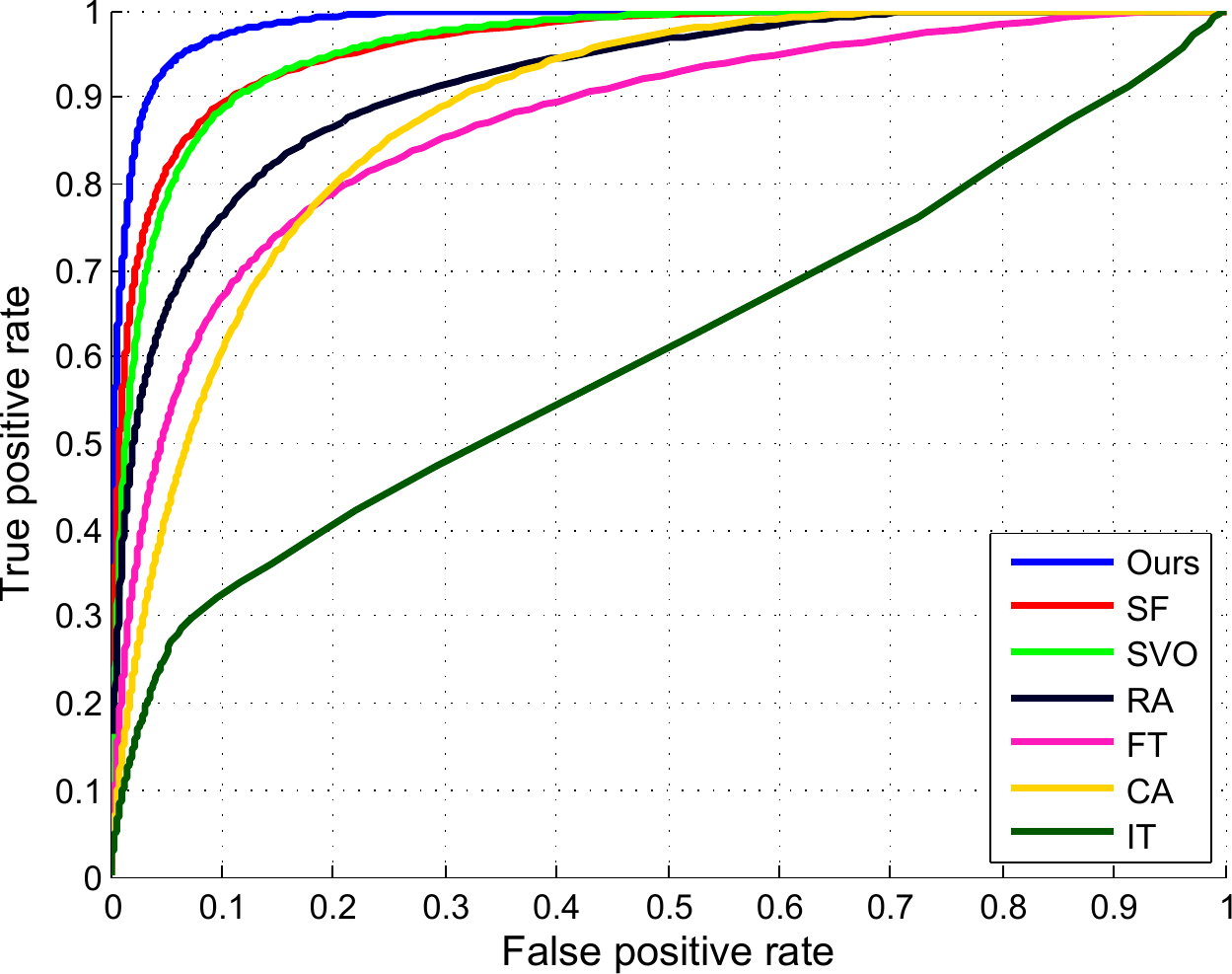} \ \\
\includegraphics[width=4cm,height=3cm]{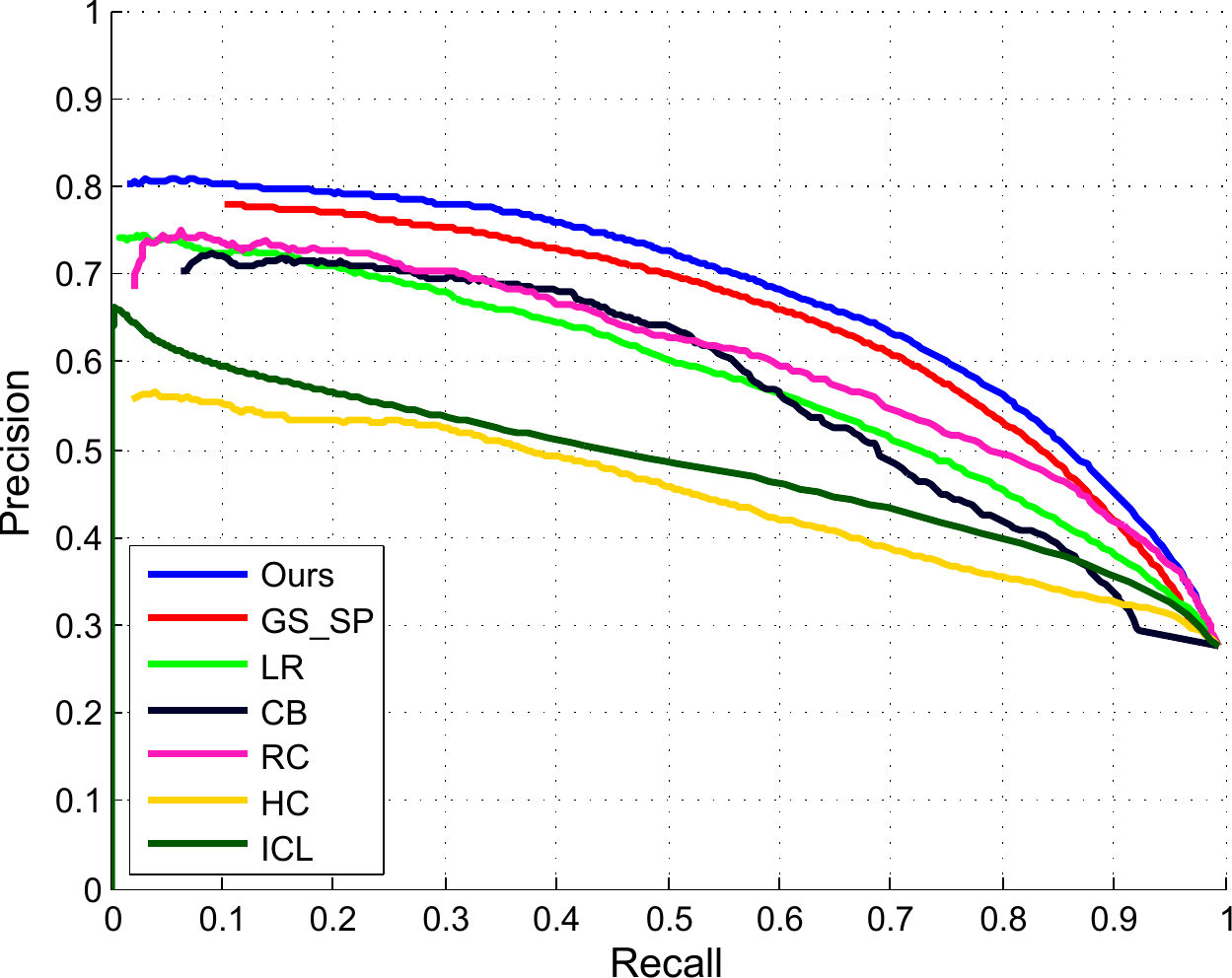} \ &
\includegraphics[width=4cm,height=3cm]{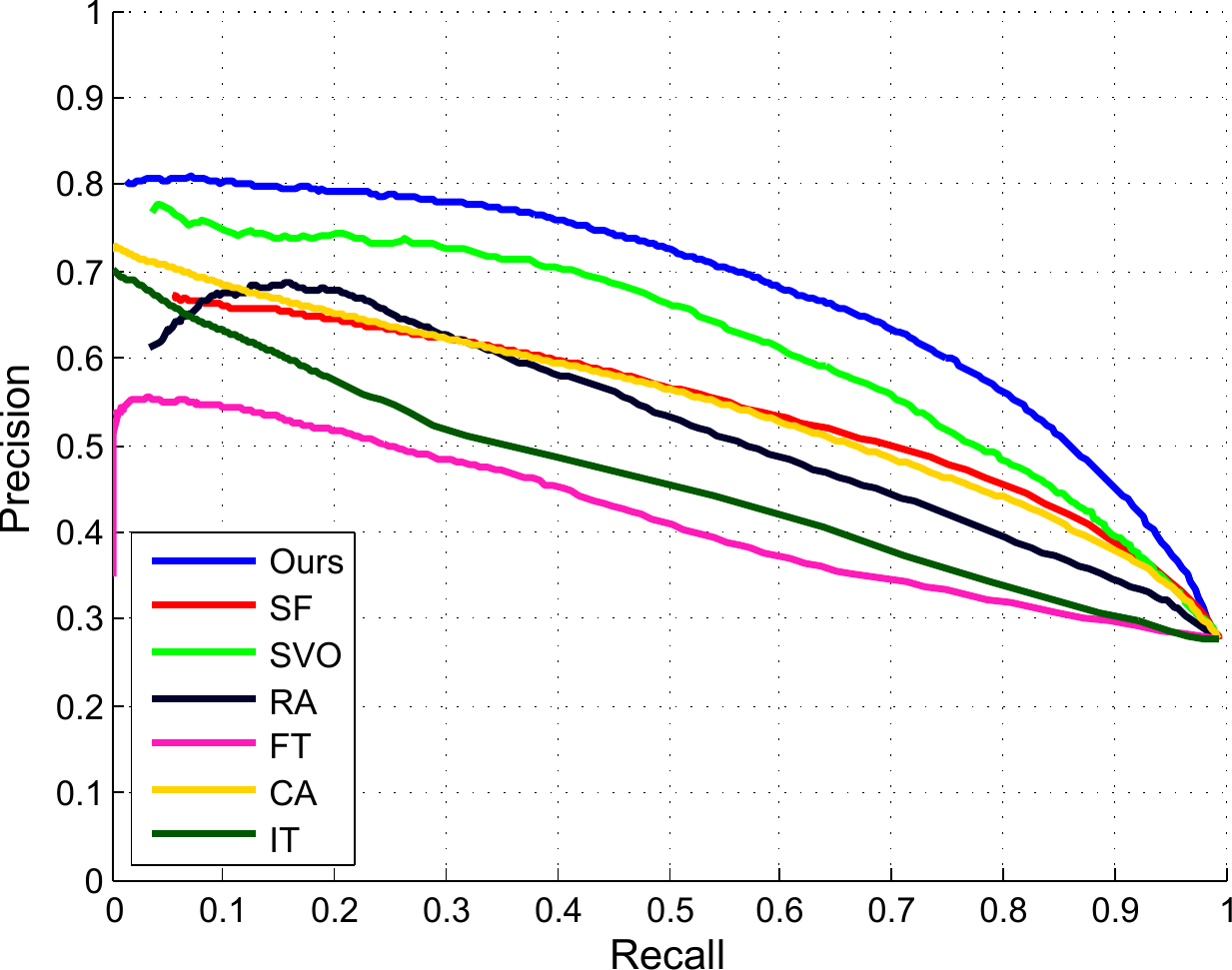} \ &
\includegraphics[width=4cm,height=3cm]{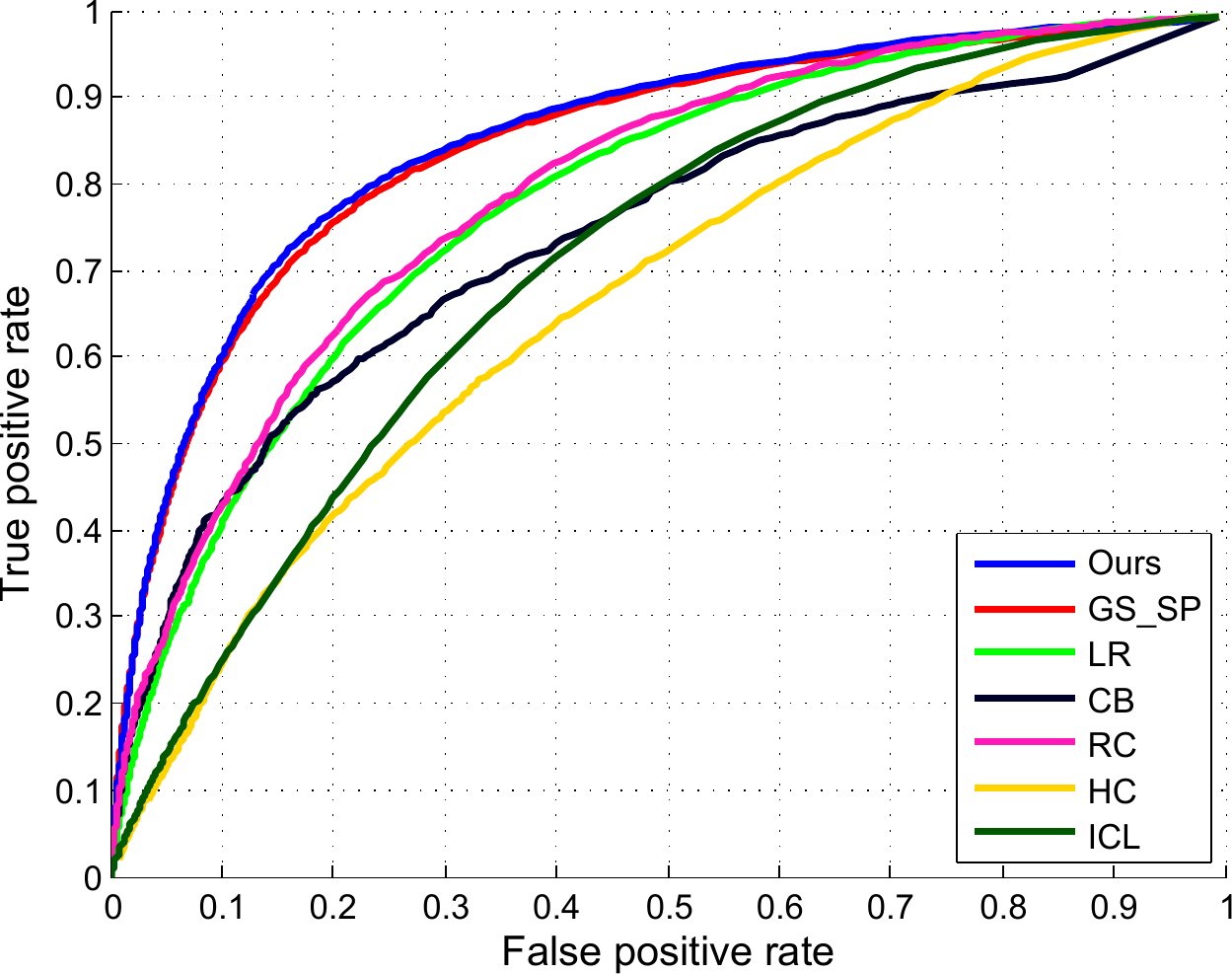} \ &
\includegraphics[width=4cm,height=3cm]{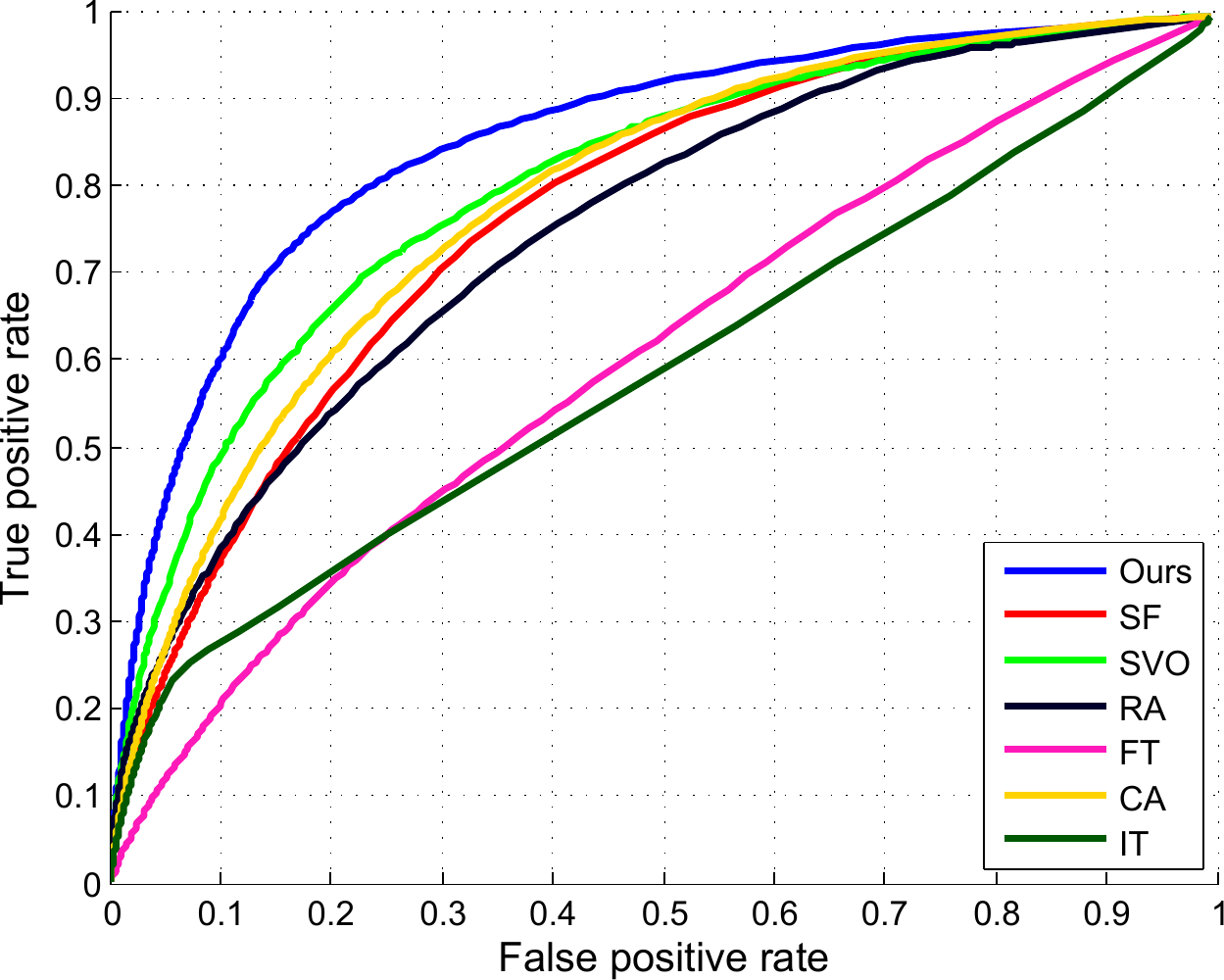} \ \\
\includegraphics[width=4cm,height=3cm]{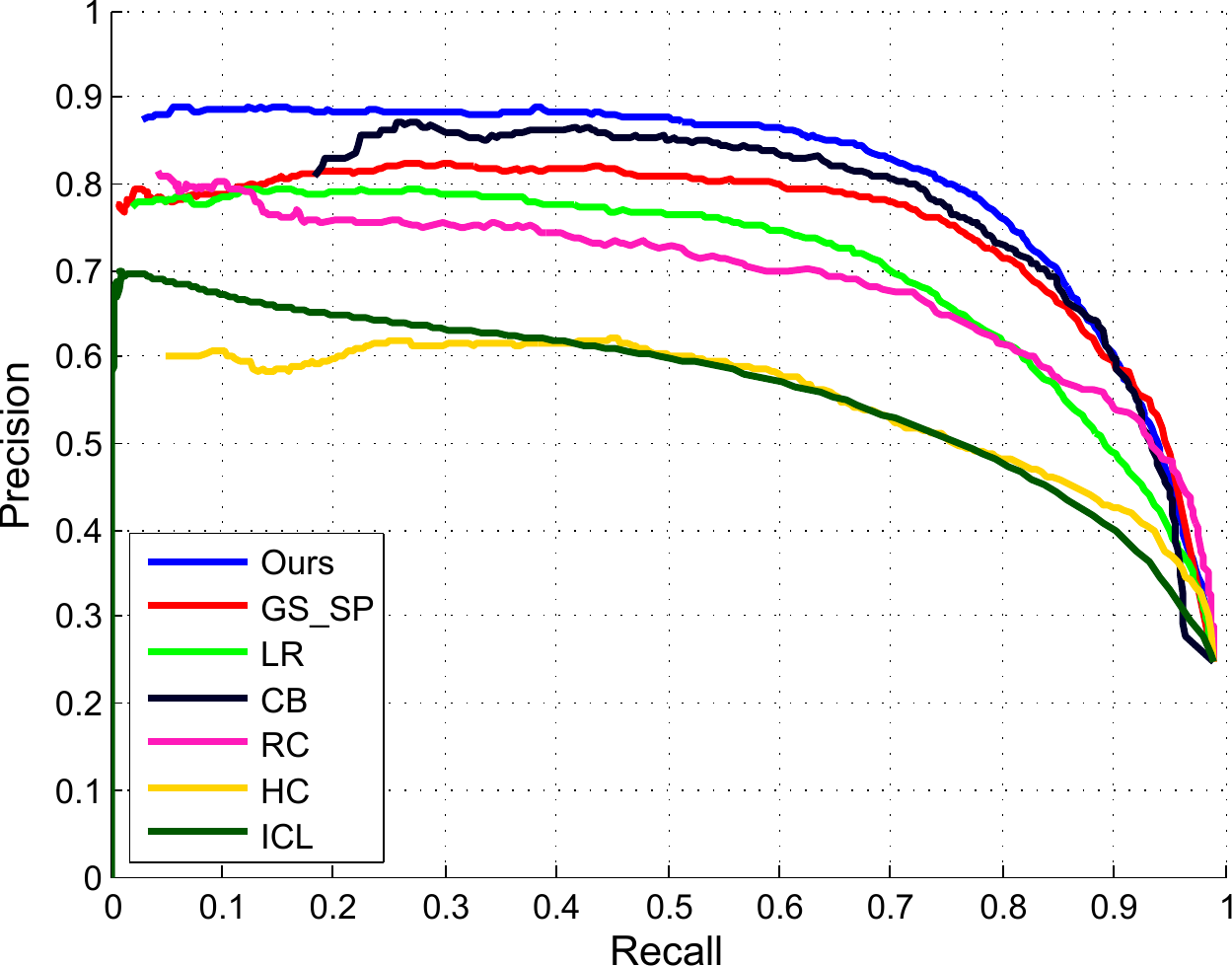} \ &
\includegraphics[width=4cm,height=3cm]{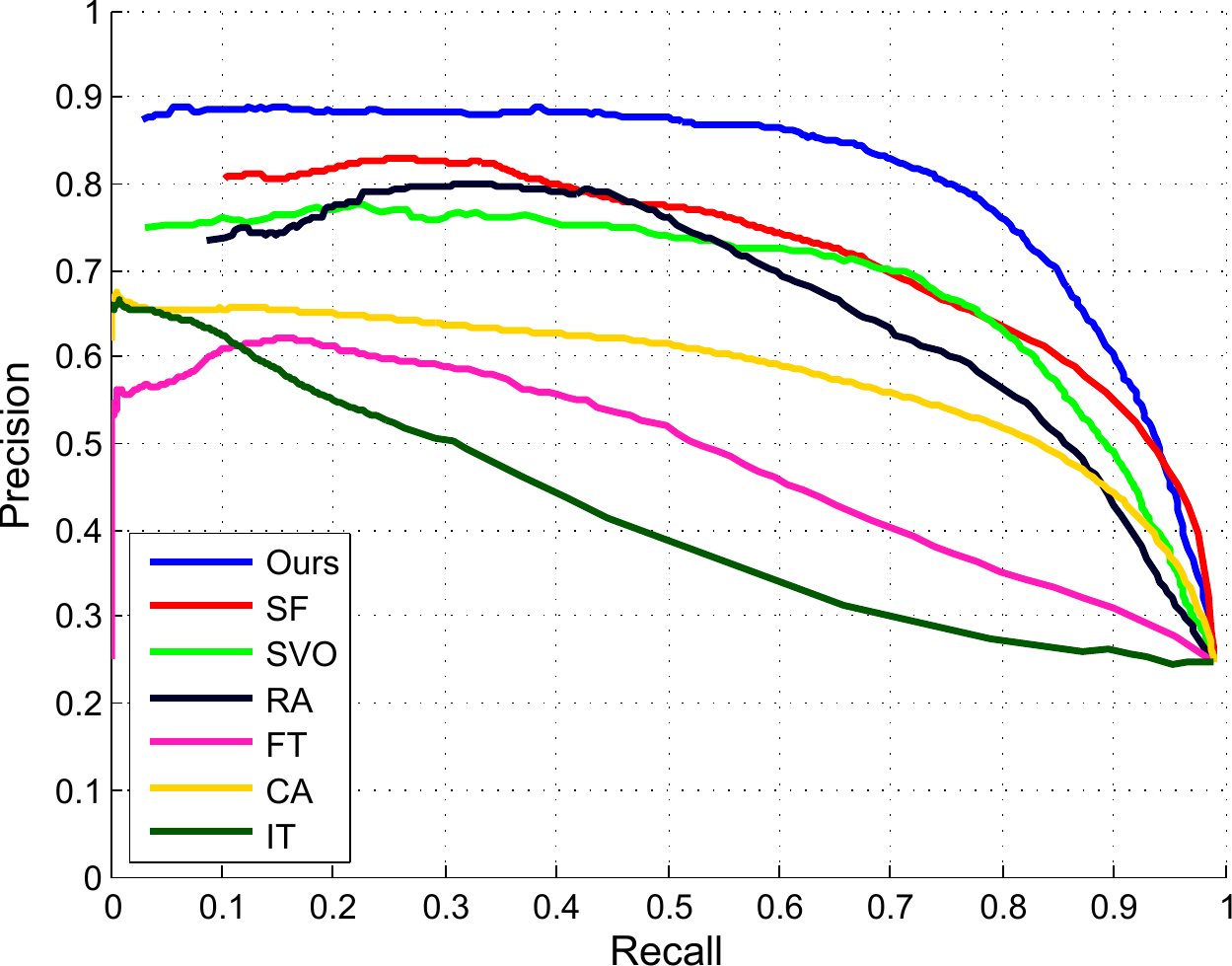} \ &
\includegraphics[width=4cm,height=3cm]{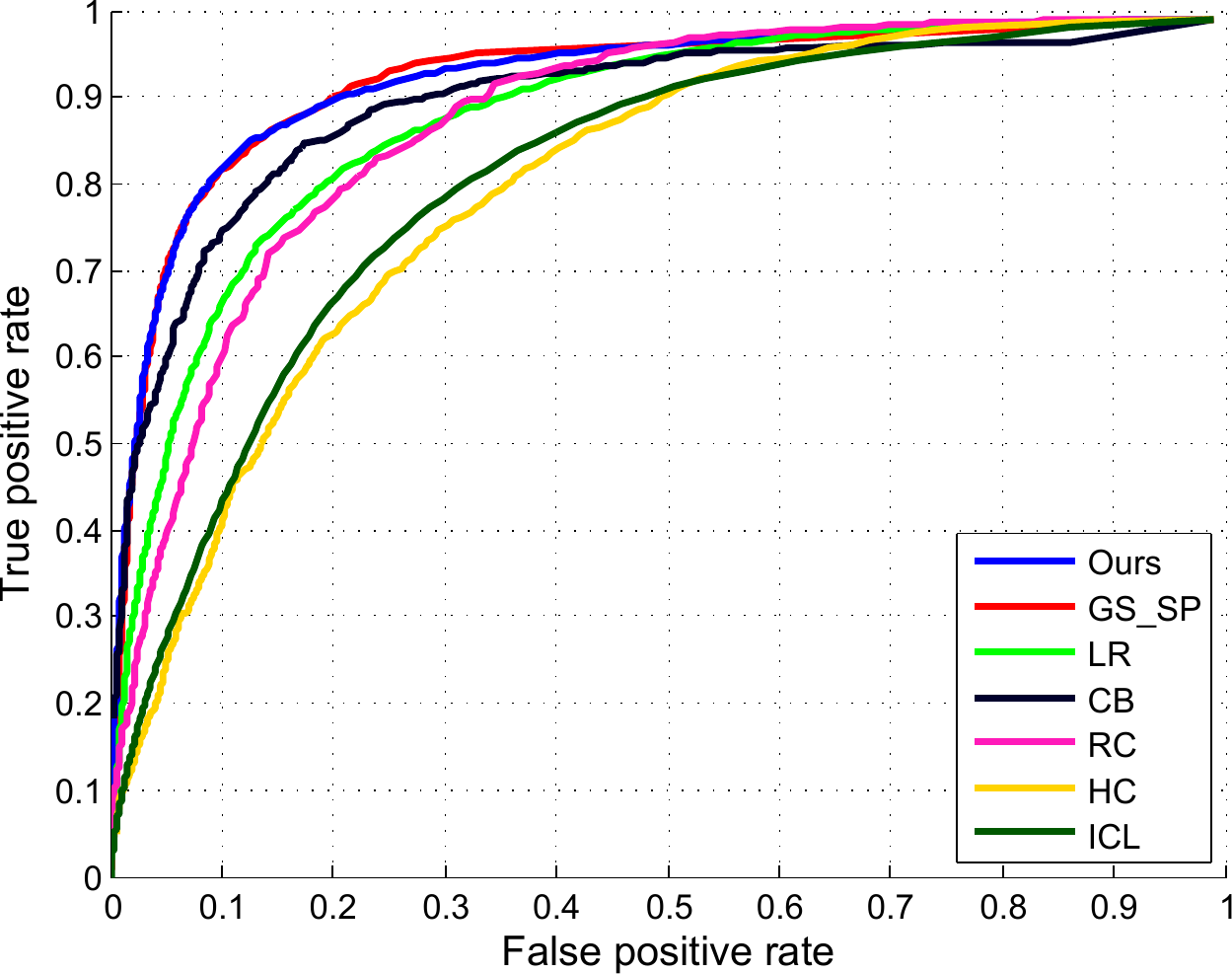} \ &
\includegraphics[width=4cm,height=3cm]{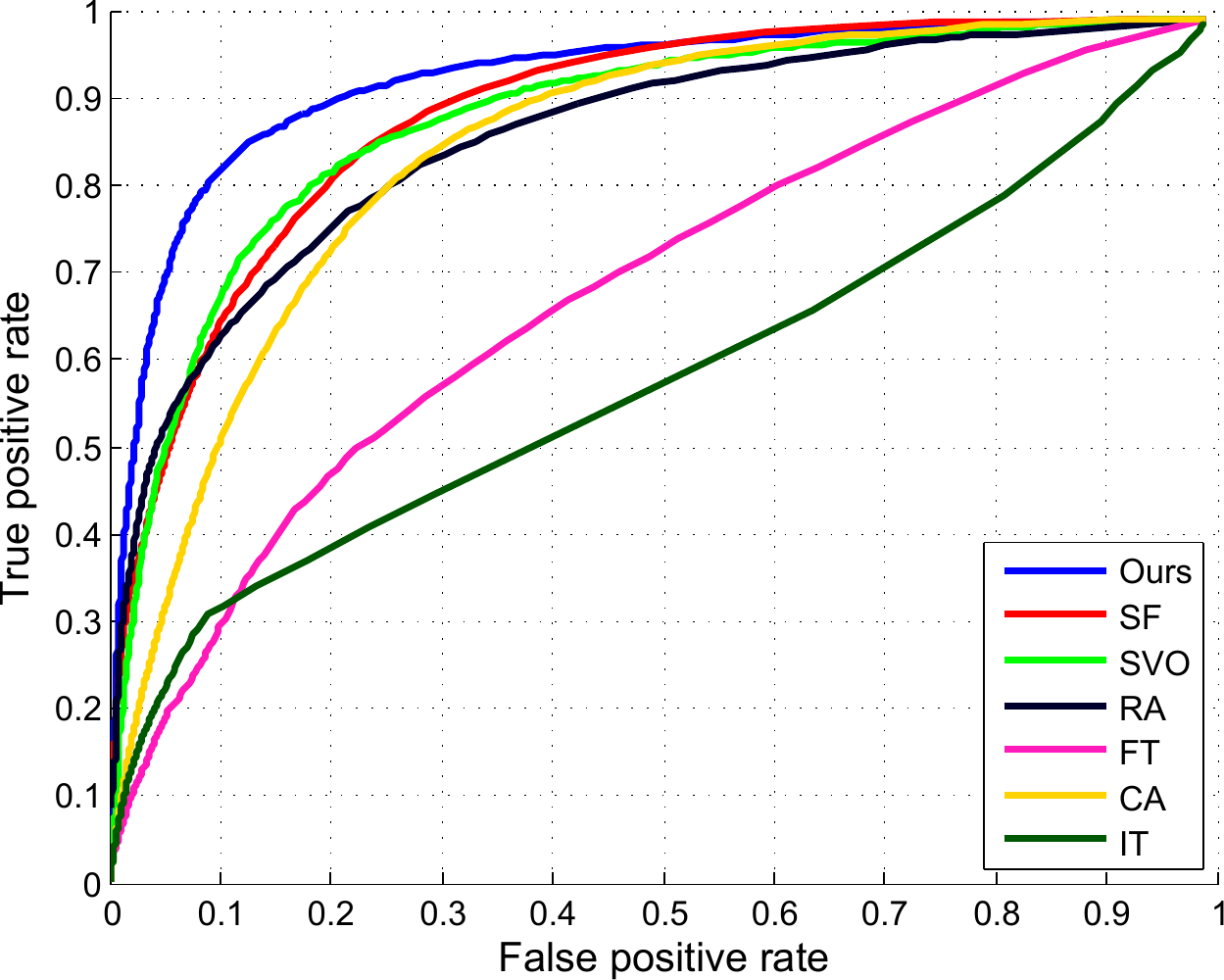} \ \\
\includegraphics[width=4cm,height=3cm]{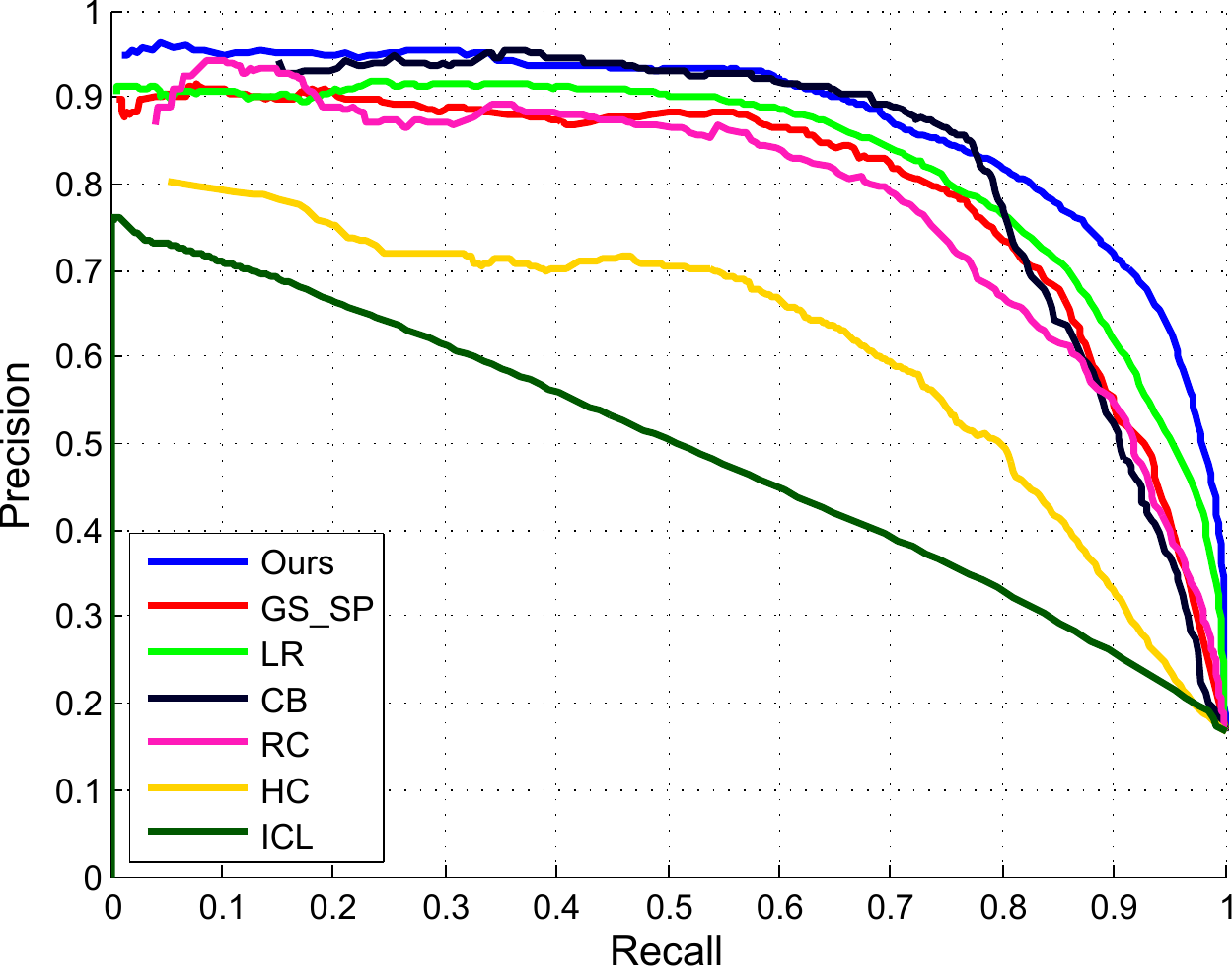} \ &
\includegraphics[width=4cm,height=3cm]{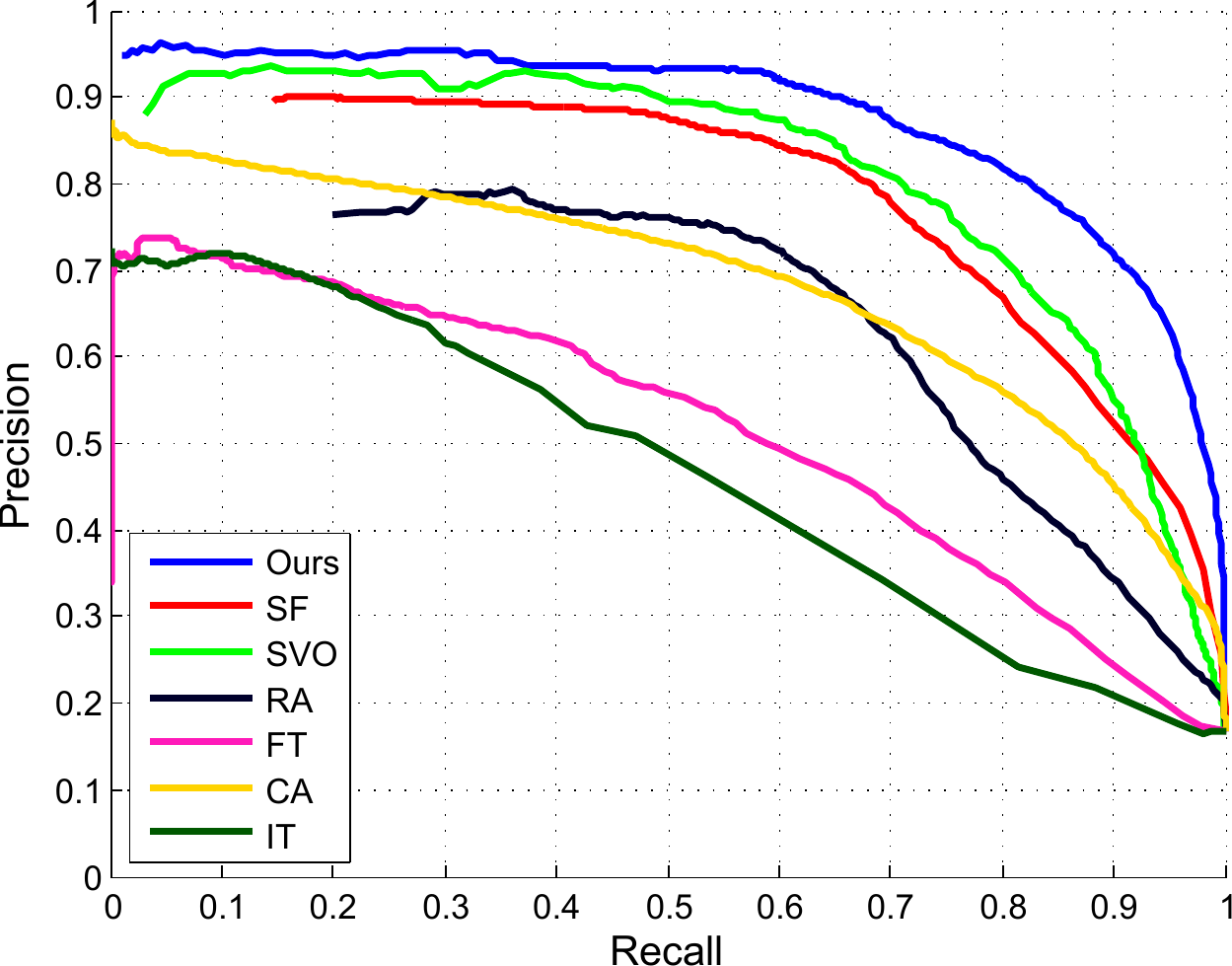} \ &
\includegraphics[width=4cm,height=3cm]{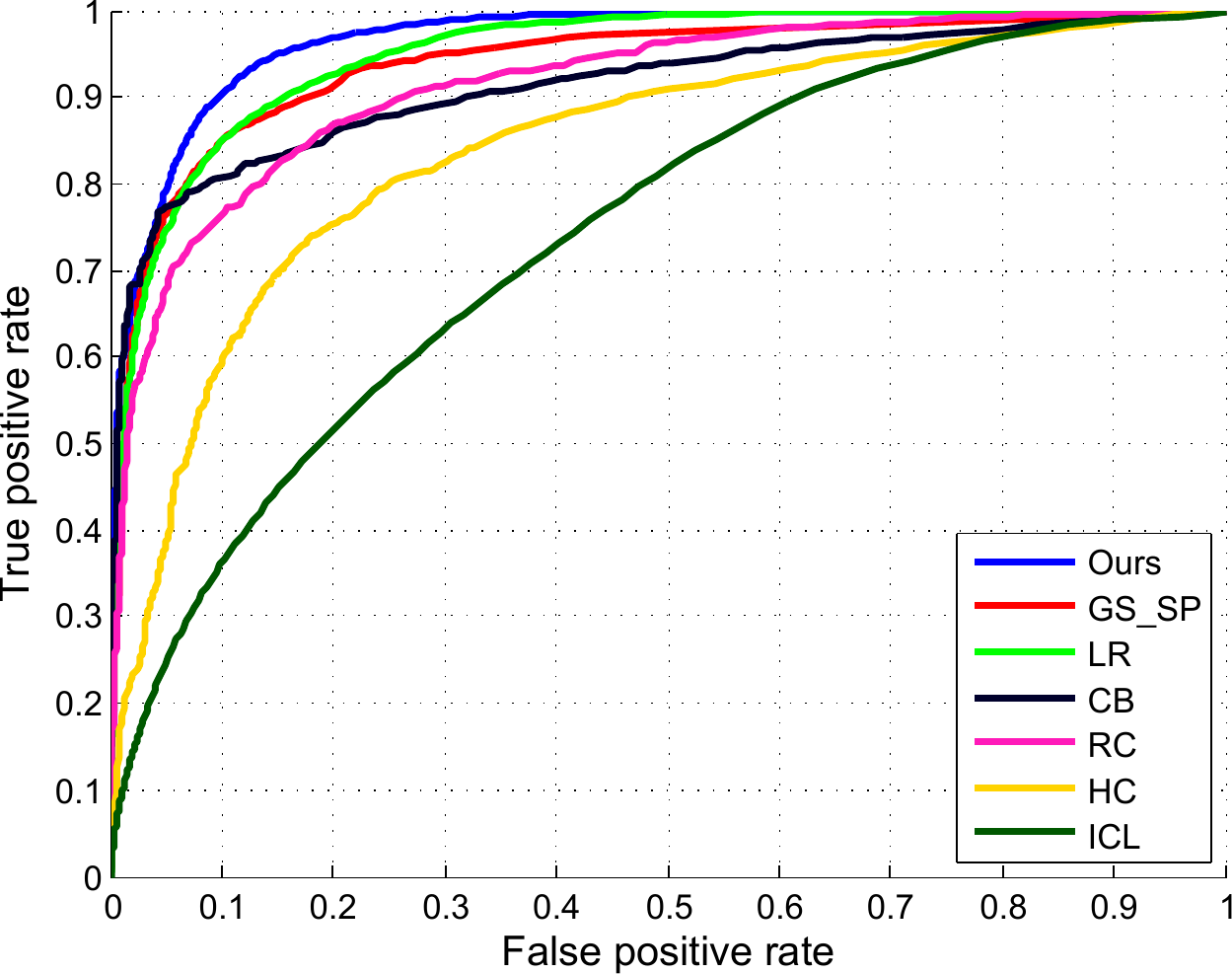} \ &
\includegraphics[width=4cm,height=3cm]{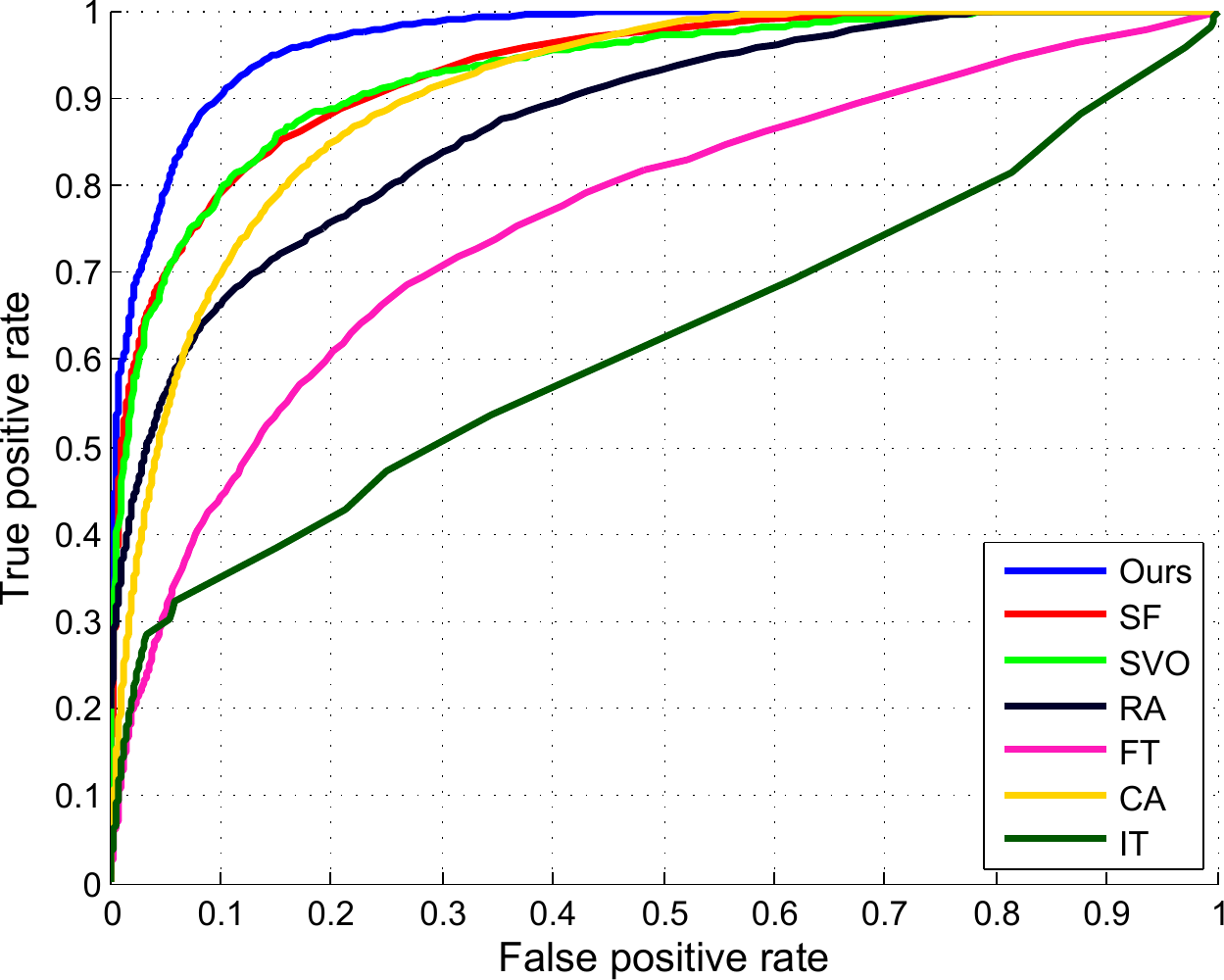} \ \\
\end{tabular}
\end{center}
\caption{Quantitative PR and ROC performance of all the thirteen approaches on the
four datasets. The left two columns show the PR curves while the right two columns display the ROC curves.
The rows from top to bottom correspond to
MSRA-1000, SOD, SED-100, and Imgsal-50, respectively.
Clearly, our approach achieve a better PR and ROC performance than the other competing approaches
in most cases.
}
\label{fig:cmparison} \end{figure*}

\begin{figure*}[t]
\begin{center}
\begin{tabular}{@{}c@{}c@{}c@{}c}
\includegraphics[width=4cm,height=3.5cm]{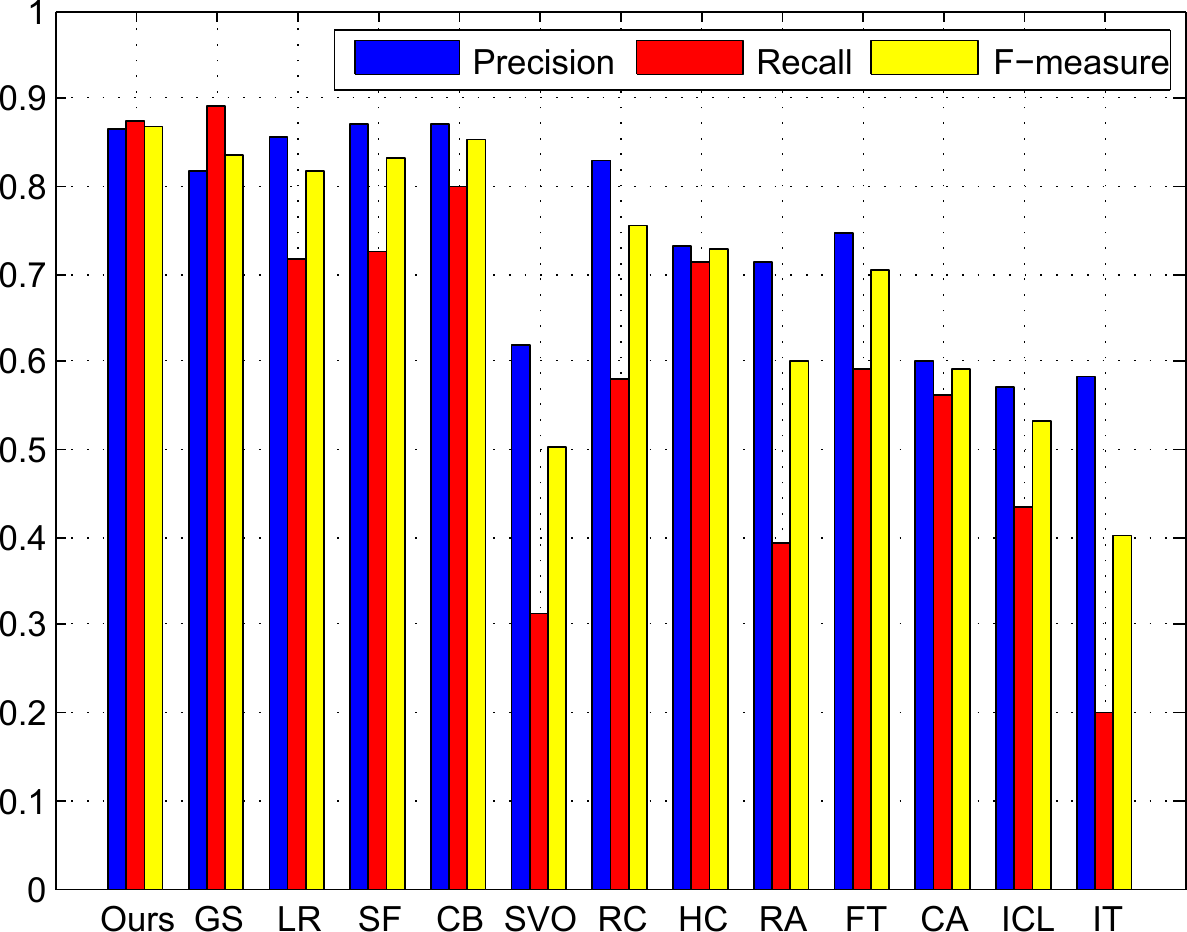} \ &
\includegraphics[width=4cm,height=3.5cm]{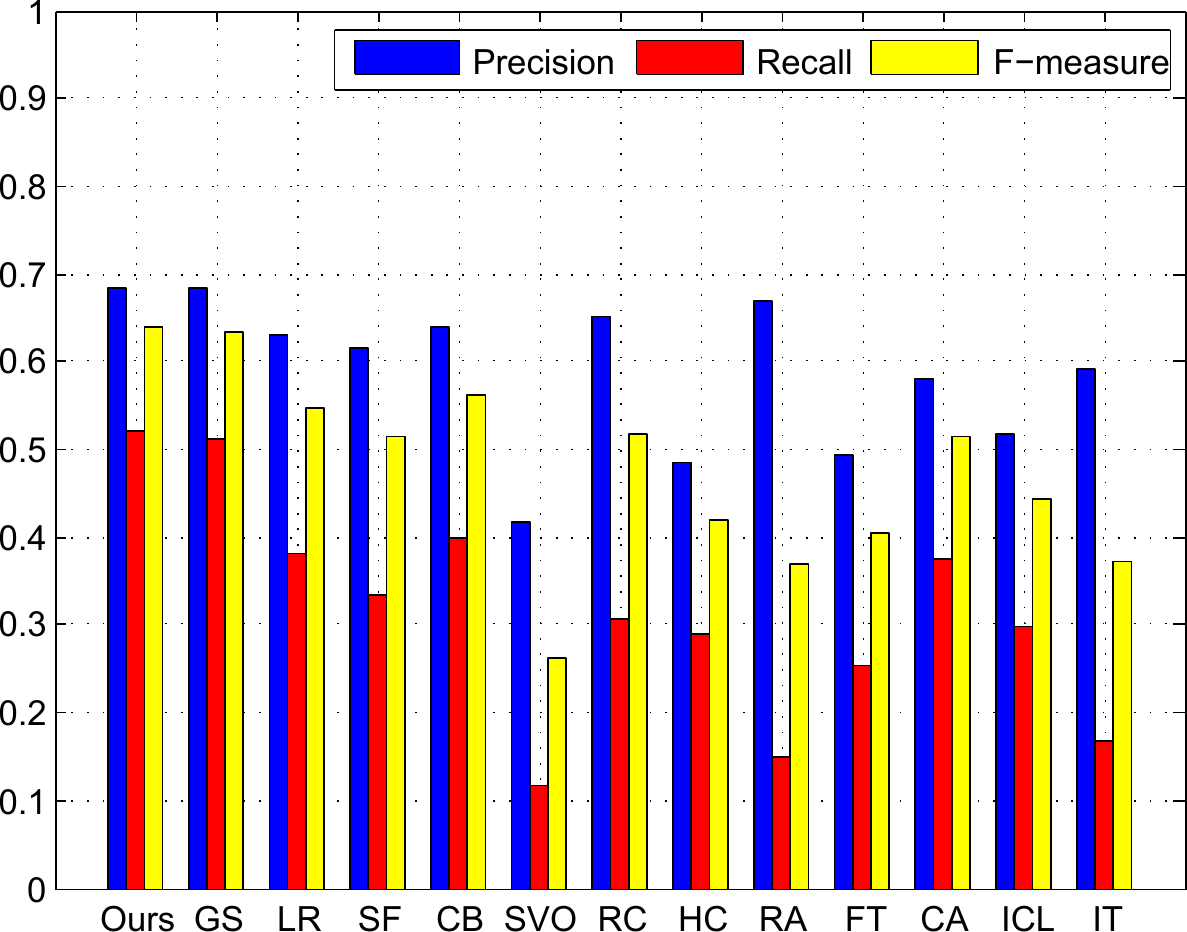} \ &
\includegraphics[width=4cm,height=3.5cm]{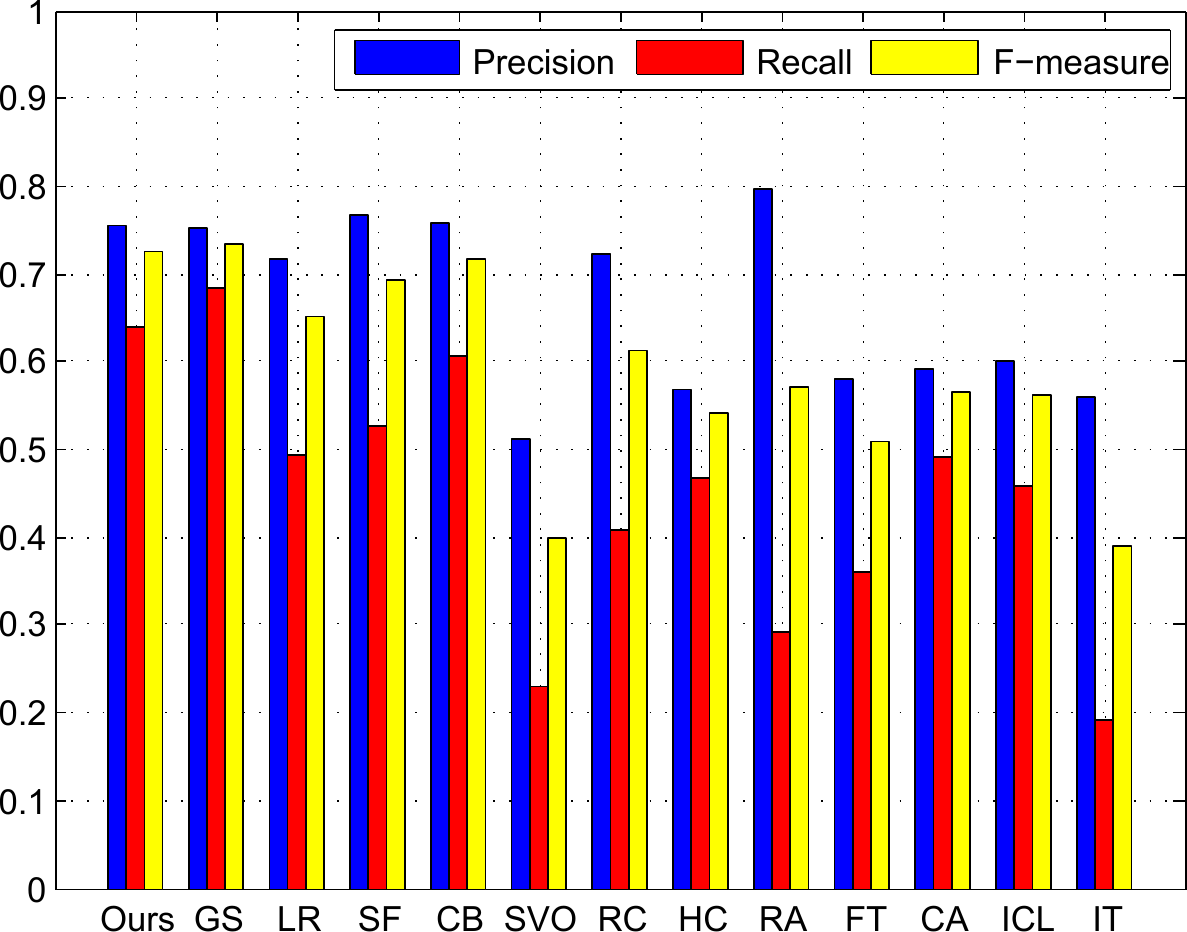} \ &
\includegraphics[width=4cm,height=3.5cm]{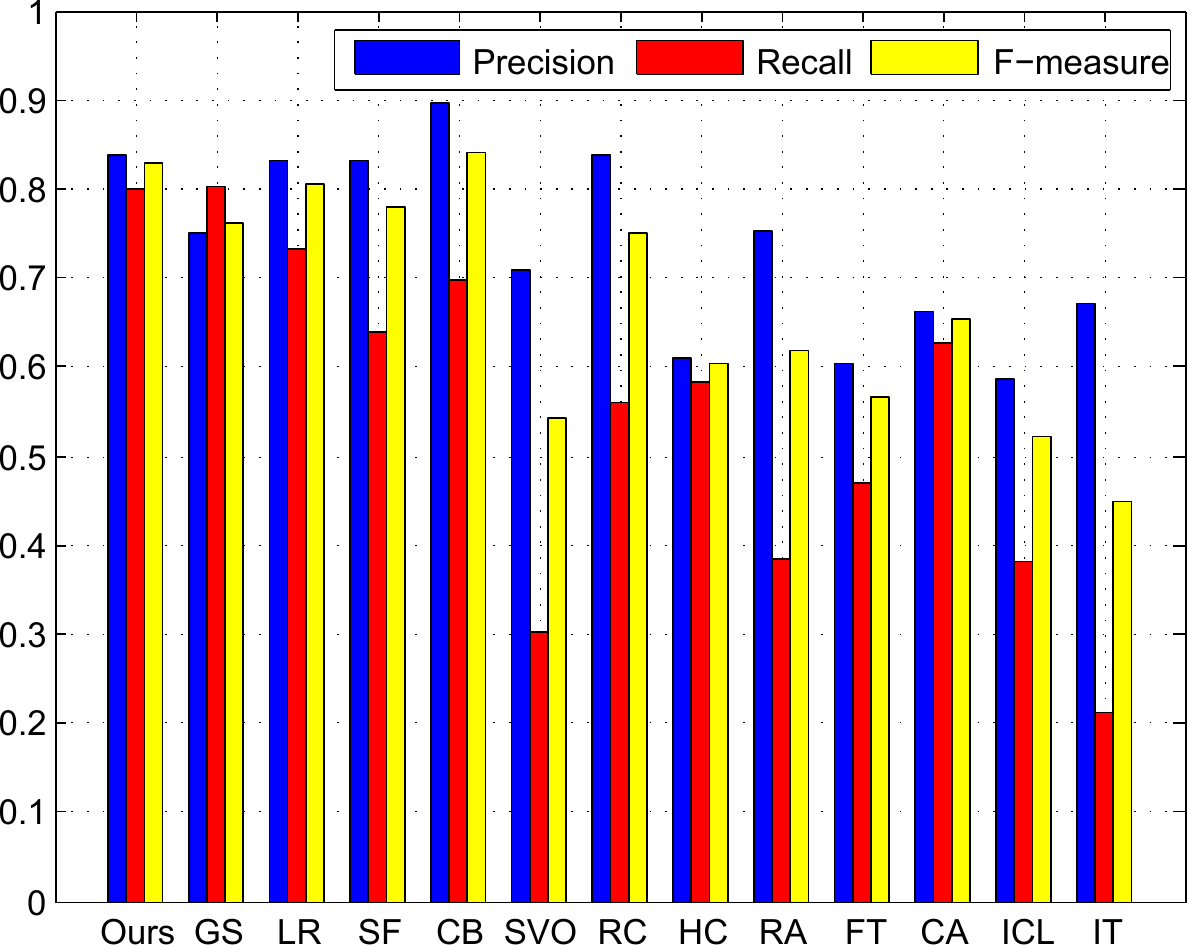} \ \\
\end{tabular}
\end{center}
\caption{Quantitative F-measure performance of all the thirteen approaches on the
four datasets. The columns from left to right correspond to MSRA-1000, SOD, SED-100, and Imgsal-50, respectively.
Here, GS is a shorthand form of GS\_SP.
It is clear that our approach achieve a good F-measure performance on the four datasets.
}
\label{fig:cmparison_Fmeasure}
\end{figure*}

\paragraph{Evaluation criterion}
For a given saliency map, we adopt four criteria
to evaluate the quantitative performance of
different approaches: precision-recall (PR) curves,
F-measures,  receiver operating characteristic (ROC) curves,
and VOC overlap scores. Specifically,
the PR curve is obtained by binarizing
the saliency map using a number of thresholds ranging from 0 to 255, as
in~\cite{achanta2009frequency,cheng2011global,shen2012unified,perazzi2012saliency}.
As described in~\cite{achanta2009frequency}, F-measure is computed as $F = ((\beta^{2}+1)P\cdot R)/(\beta^{2}P + R)$.
Here, $P$ and $R$
are the precision and recall rates obtained by
binarizing the saliency map using an adaptive threshold
that is twice the overall mean saliency value~\cite{achanta2009frequency}.
$\beta^2=0.3$ is the same as that in~\cite{achanta2009frequency}.
Identical to~\cite{BorjiSI12},
the ROC curve is generated
from true
positive rates and false positive rates
obtained during the calculation of the corresponding PR curve.
The VOC Overlap score~\cite{RosenfeldW11} is defined
as $\frac{|S \cap S'|}{|S \cup S'|}$.
Here, $S$ is the ground-truth foreground mask,
and $S'$ is the object segmentation mask
obtained by binarizing
the saliency map using the same adaptive threshold
during the calculation of F-measure.

\begin{figure*}[t]
\begin{center}
\scalebox{0.95}{
\begin{tabular}{@{}c@{}c@{}c@{}c@{}c@{}c@{}c@{}c@{}c@{}c@{}c@{}c@{}c@{}c@{}c}
{\scriptsize Image} & {\scriptsize GT} & {\scriptsize Ours}
& {\scriptsize GS\rule[-2pt]{1mm}{0.5pt}SP\cite{WeiWZ012}}& {\scriptsize LR\cite{shen2012unified}}
& {\scriptsize SF\cite{perazzi2012saliency}} & {\scriptsize CB\cite{jiang2011automatic}} &
{\scriptsize SVO\cite{ChangLCL11}}  & {\scriptsize RC\cite{cheng2011global}} &
{\scriptsize HC\cite{cheng2011global}} &
{\scriptsize RA\cite{RahtuKSH10}} &  {\scriptsize FT\cite{achanta2009frequency}} &
{\scriptsize CA\cite{GofermanZT10}} & {\scriptsize ICL\cite{DBLP:conf/nips/HouZ08}}
& {\scriptsize IT\cite{itti1998model}}\\

\includegraphics[width=0.065\linewidth]{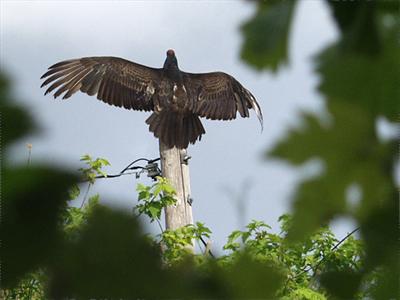} \ &
\includegraphics[width=0.065\linewidth]{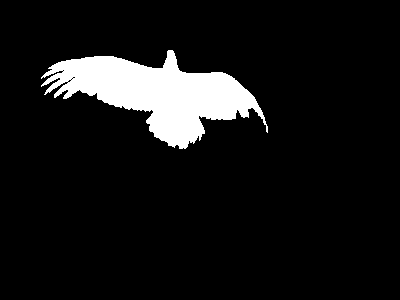} \ &
\includegraphics[width=0.065\linewidth]{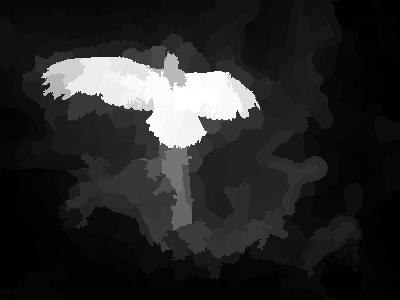} \ &
\includegraphics[width=0.065\linewidth]{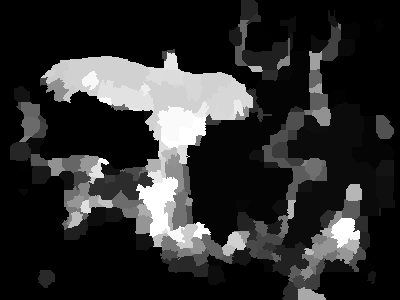} \ &
\includegraphics[width=0.065\linewidth]{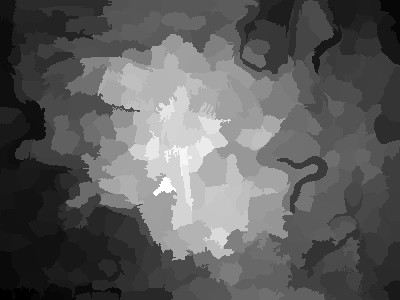} \ &
\includegraphics[width=0.065\linewidth]{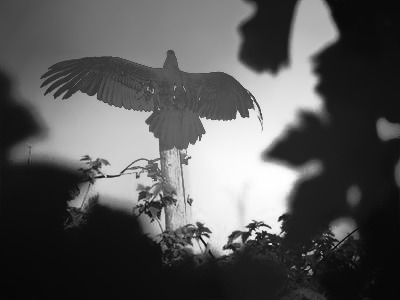} \ &
\includegraphics[width=0.065\linewidth]{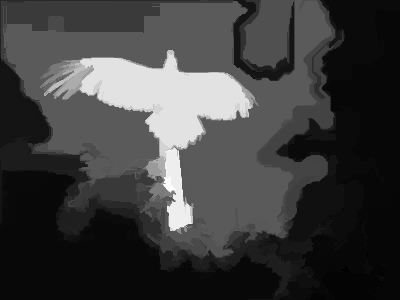} \ &
\includegraphics[width=0.065\linewidth]{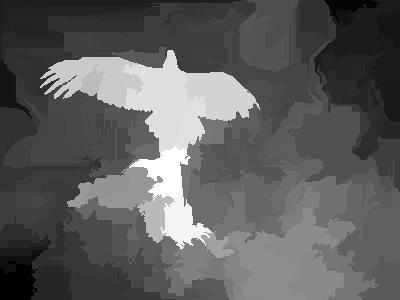} \ &
\includegraphics[width=0.065\linewidth]{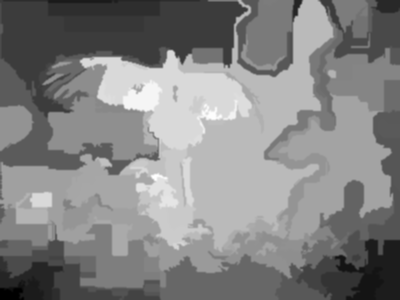} \ & \includegraphics[width=0.065\linewidth]{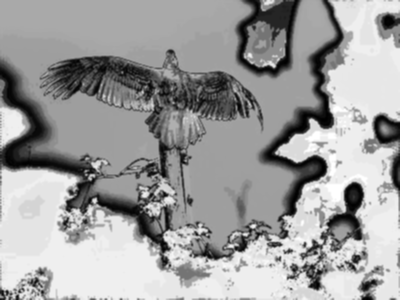} \ &
\includegraphics[width=0.065\linewidth]{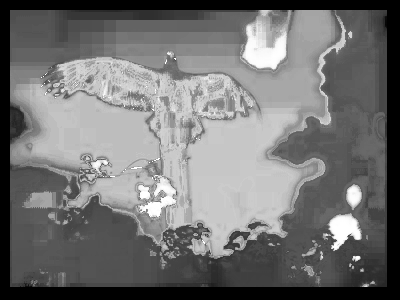} \ &
\includegraphics[width=0.065\linewidth]{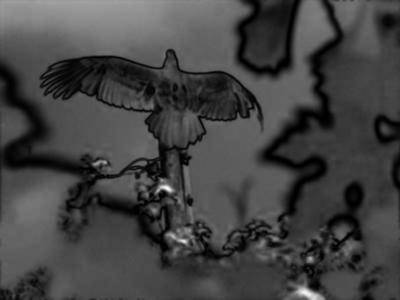} \ &
\includegraphics[width=0.065\linewidth]{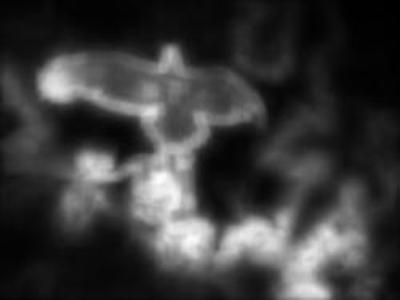} \ &
\includegraphics[width=0.065\linewidth]{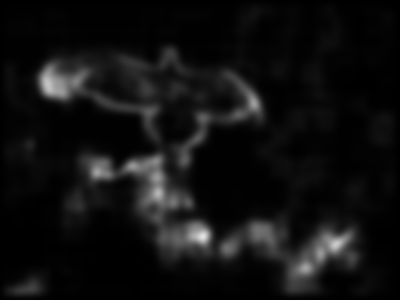} \ &
\includegraphics[width=0.065\linewidth]{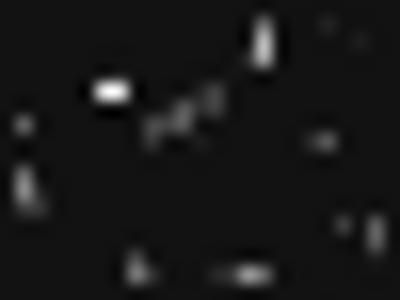}\\
\includegraphics[width=0.065\linewidth]{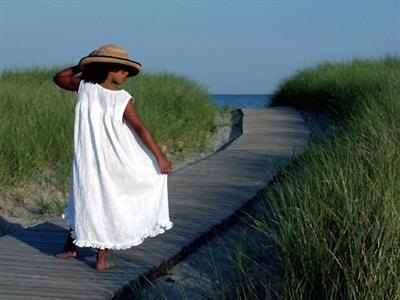} \ &
\includegraphics[width=0.065\linewidth]{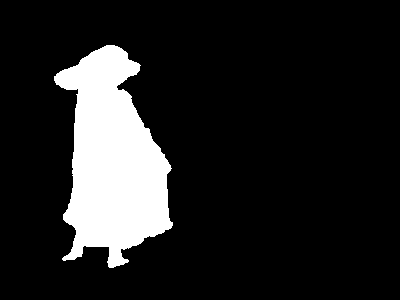} \ &
\includegraphics[width=0.065\linewidth]{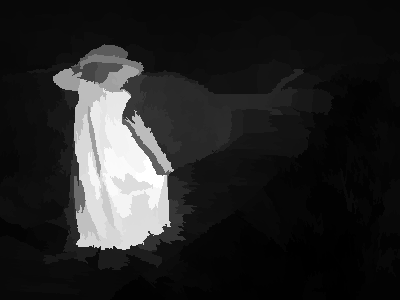} \ &
\includegraphics[width=0.065\linewidth]{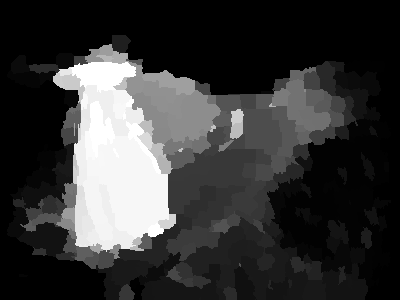} \ &
\includegraphics[width=0.065\linewidth]{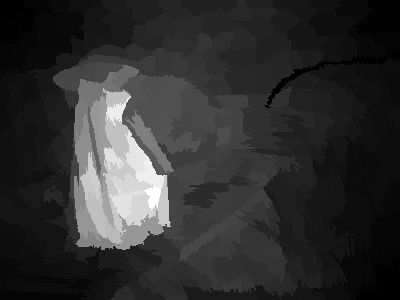} \ &
\includegraphics[width=0.065\linewidth]{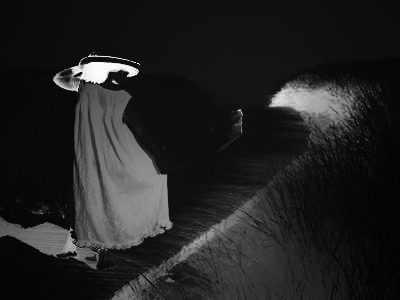} \ &
\includegraphics[width=0.065\linewidth]{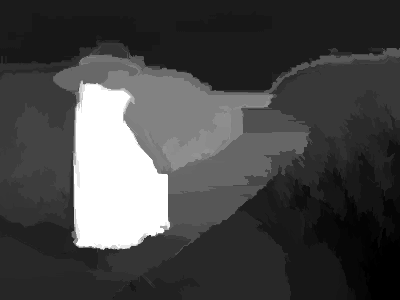} \ &
\includegraphics[width=0.065\linewidth]{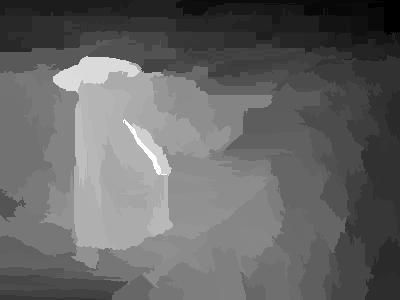} \ &
\includegraphics[width=0.065\linewidth]{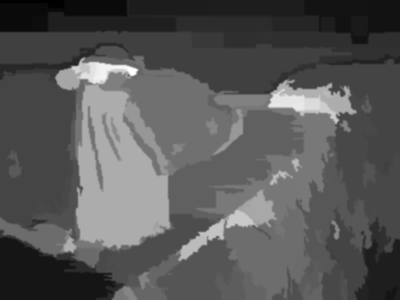} \ & \includegraphics[width=0.065\linewidth]{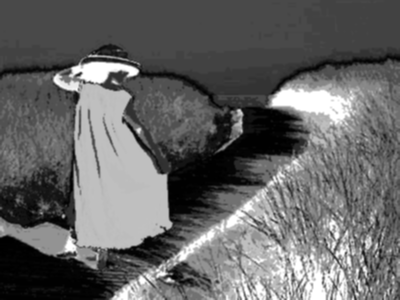} \ &
\includegraphics[width=0.065\linewidth]{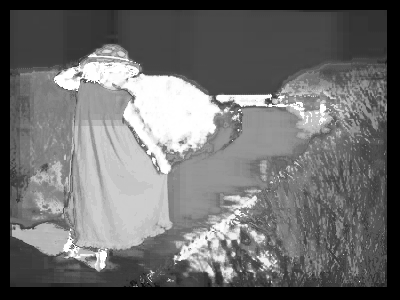} \ &
\includegraphics[width=0.065\linewidth]{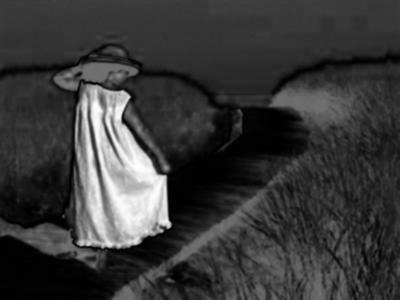} \ &
\includegraphics[width=0.065\linewidth]{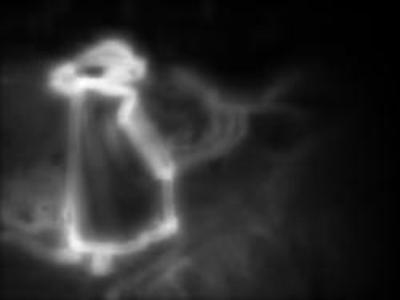} \ &
\includegraphics[width=0.065\linewidth]{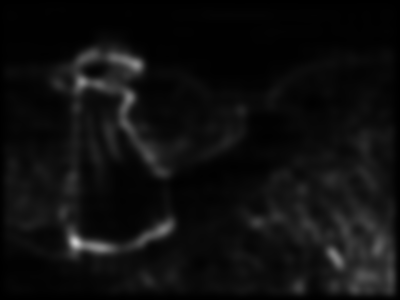} \ &
\includegraphics[width=0.065\linewidth]{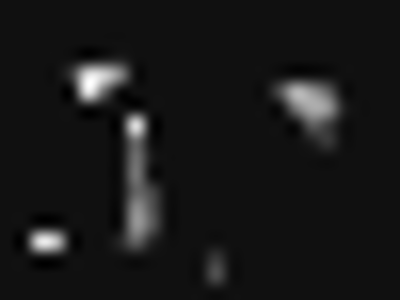}\\
\includegraphics[width=0.065\linewidth]{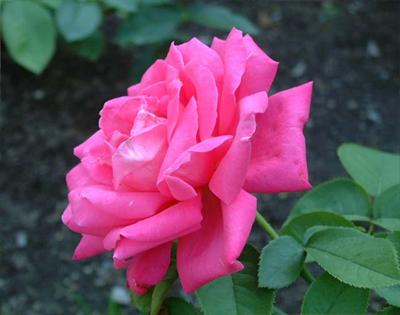} \ &
\includegraphics[width=0.065\linewidth]{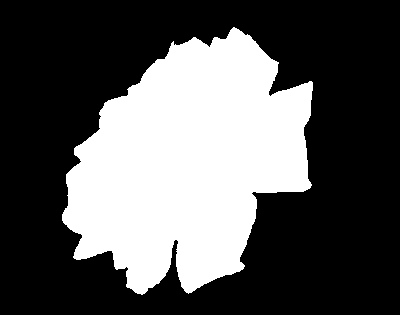} \ &
\includegraphics[width=0.065\linewidth]{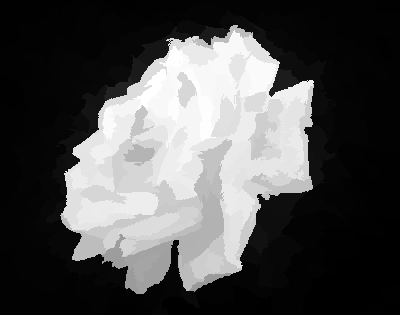} \ &
\includegraphics[width=0.065\linewidth]{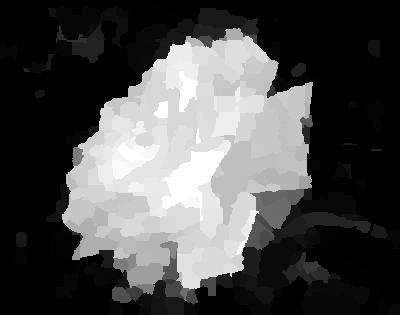} \ &
\includegraphics[width=0.065\linewidth]{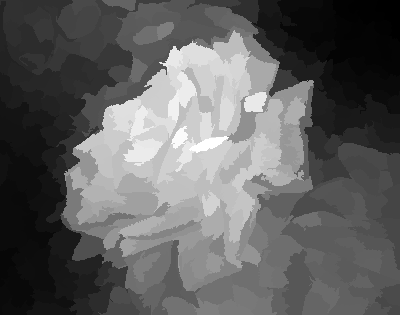} \ &
\includegraphics[width=0.065\linewidth]{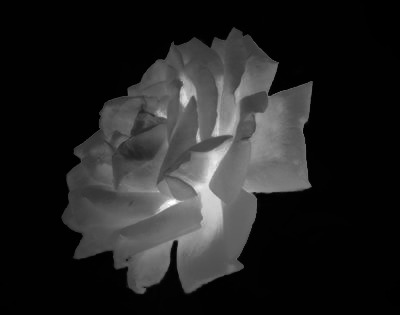} \ &
\includegraphics[width=0.065\linewidth]{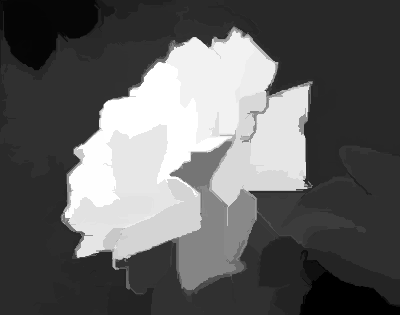} \ &
\includegraphics[width=0.065\linewidth]{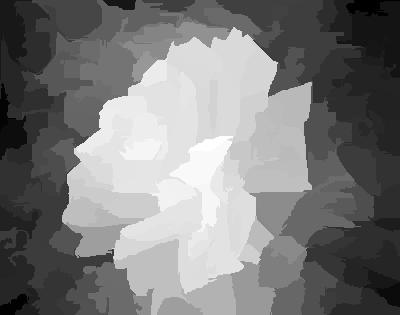} \ &
\includegraphics[width=0.065\linewidth]{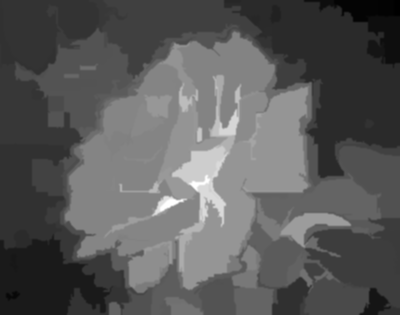} \ &  \includegraphics[width=0.065\linewidth]{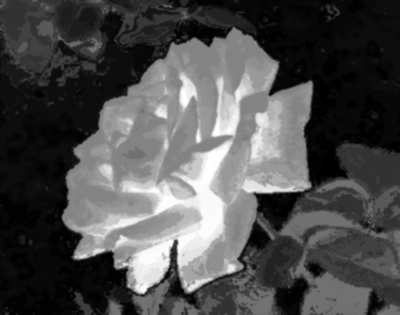} \ &
\includegraphics[width=0.065\linewidth]{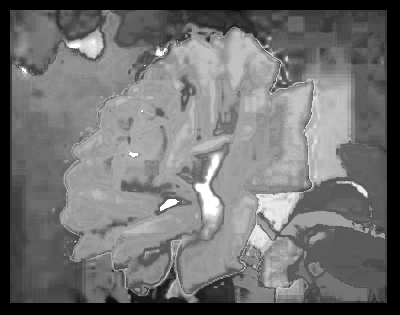} \ &
\includegraphics[width=0.065\linewidth]{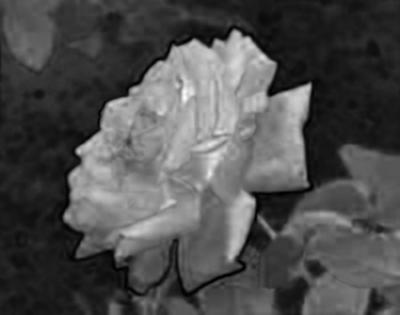} \ &
\includegraphics[width=0.065\linewidth]{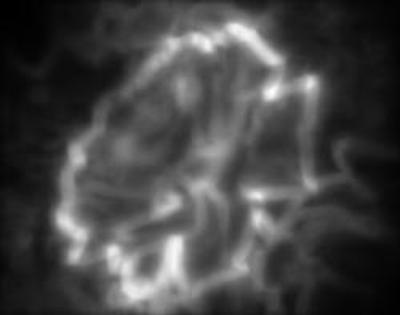} \ &
\includegraphics[width=0.065\linewidth]{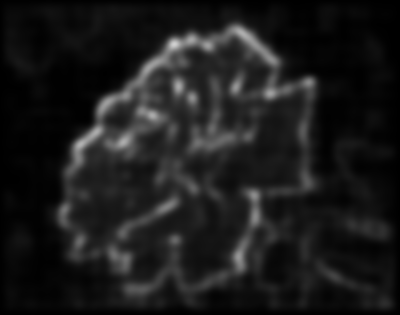} \ &
\includegraphics[width=0.065\linewidth]{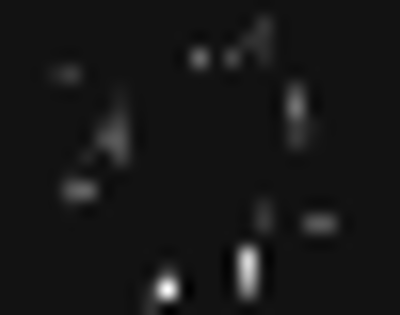}\\
\includegraphics[width=0.065\linewidth]{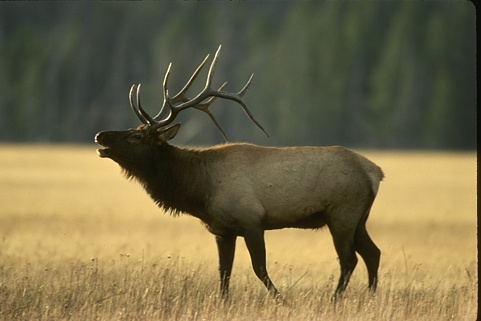} \ &
\includegraphics[width=0.065\linewidth]{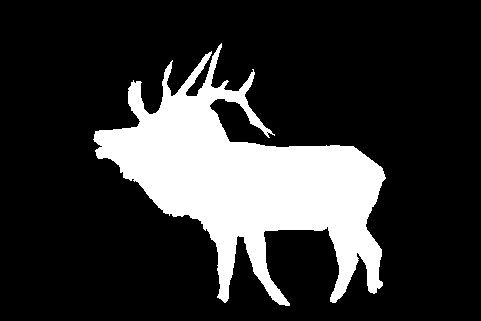} \ &
\includegraphics[width=0.065\linewidth]{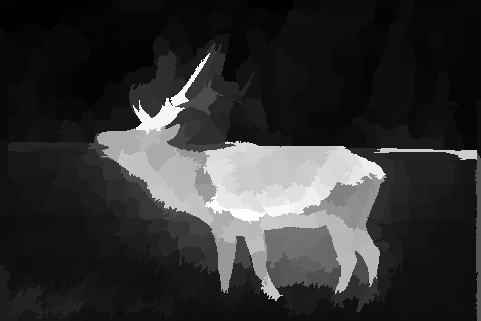} \ &
\includegraphics[width=0.065\linewidth]{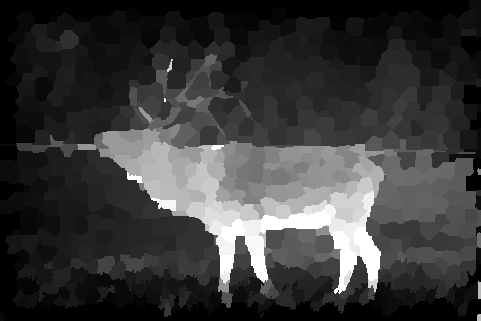} \ &
\includegraphics[width=0.065\linewidth]{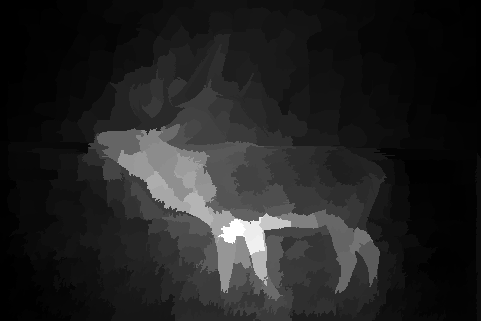} \ &
\includegraphics[width=0.065\linewidth]{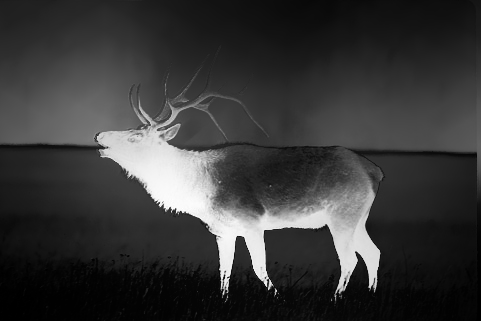} \ &
\includegraphics[width=0.065\linewidth]{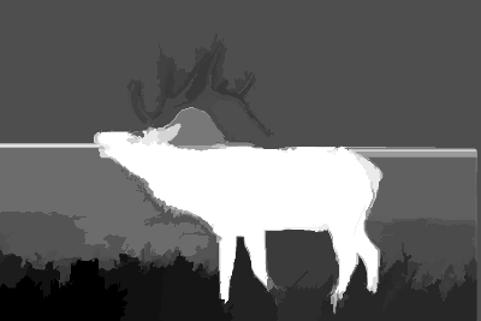} \ &
\includegraphics[width=0.065\linewidth]{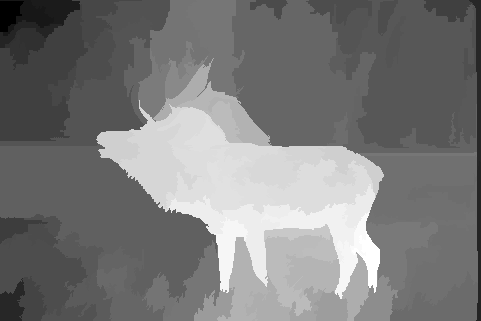} \ &
\includegraphics[width=0.065\linewidth]{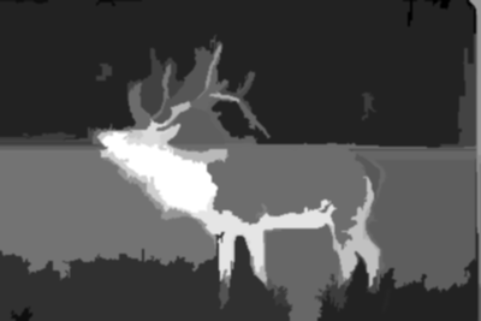} \ & \includegraphics[width=0.065\linewidth]{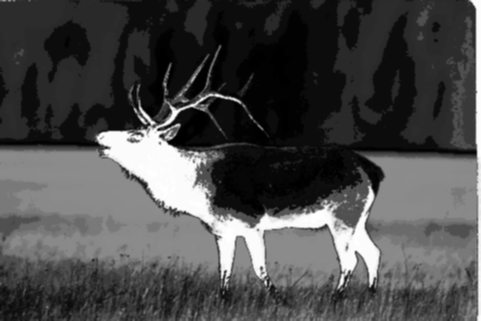} \ &
\includegraphics[width=0.065\linewidth]{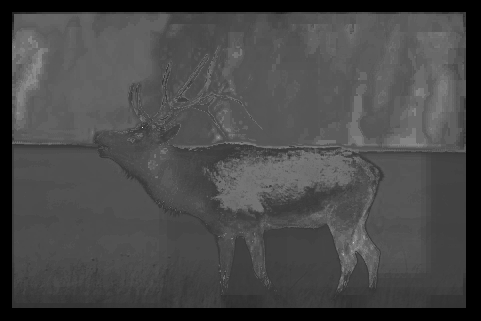} \ &
\includegraphics[width=0.065\linewidth]{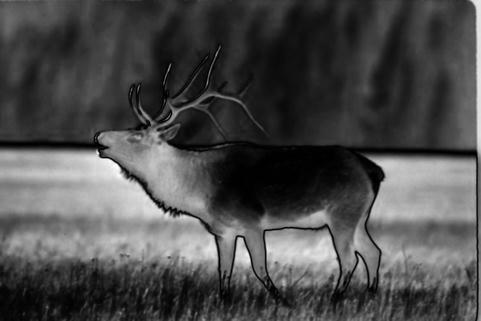} \ &
\includegraphics[width=0.065\linewidth]{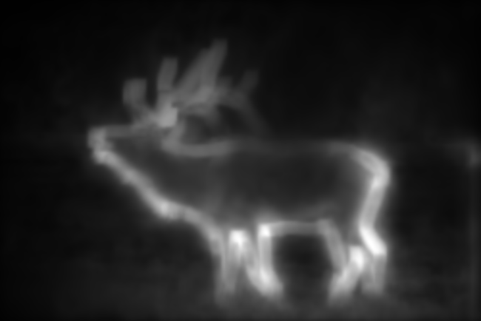} \ &
\includegraphics[width=0.065\linewidth]{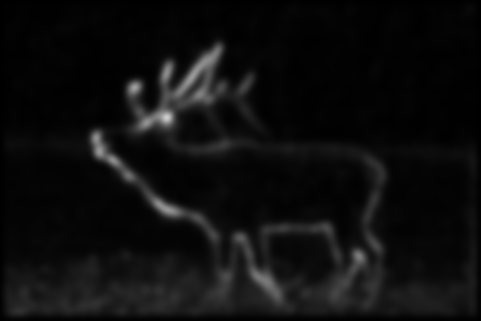} \ &
\includegraphics[width=0.065\linewidth]{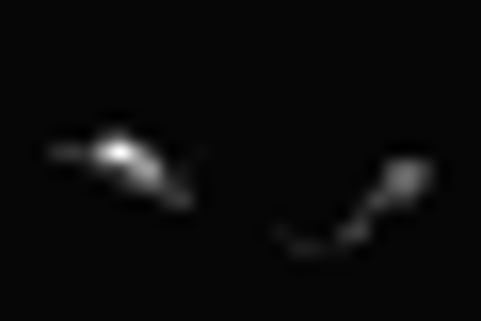}\\
\includegraphics[width=0.065\linewidth]{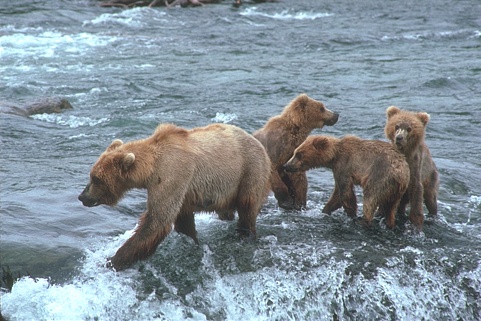} \ &
\includegraphics[width=0.065\linewidth]{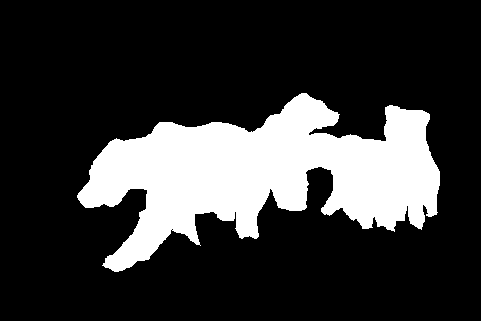} \ &
\includegraphics[width=0.065\linewidth]{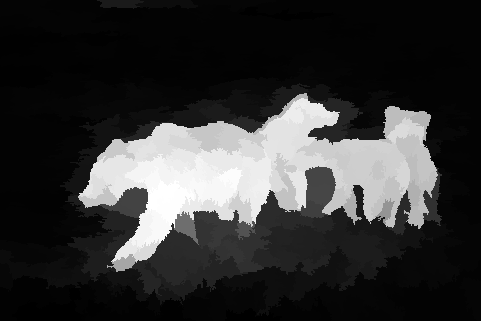} \ &
\includegraphics[width=0.065\linewidth]{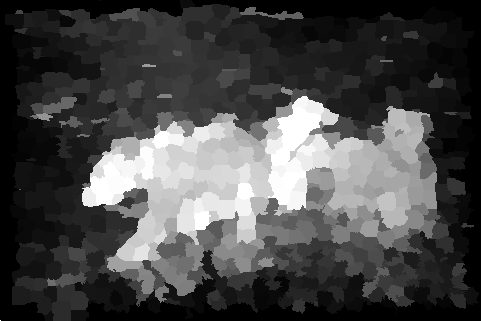} \ &
\includegraphics[width=0.065\linewidth]{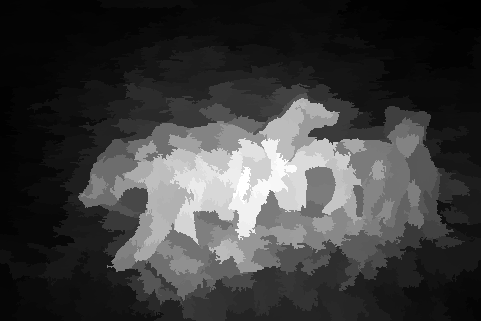} \ &
\includegraphics[width=0.065\linewidth]{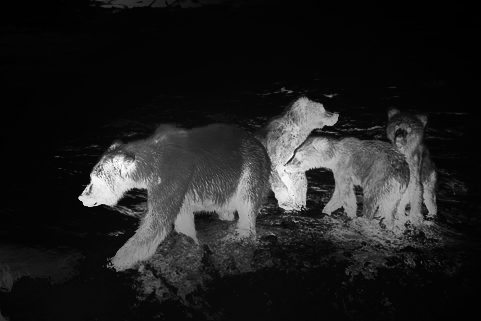} \ &
\includegraphics[width=0.065\linewidth]{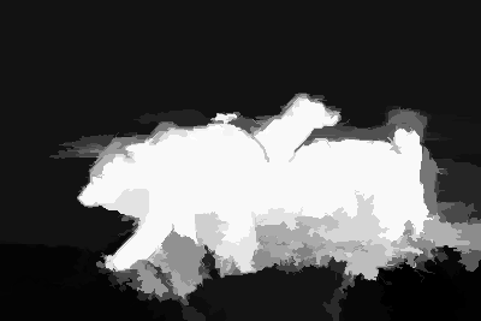} \ &
\includegraphics[width=0.065\linewidth]{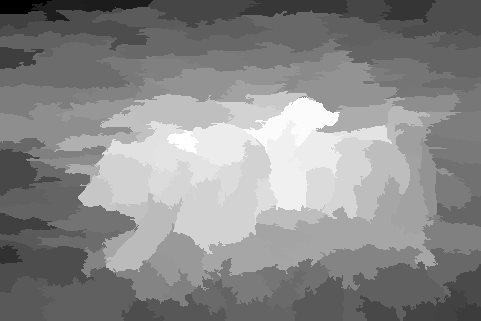} \ &
\includegraphics[width=0.065\linewidth]{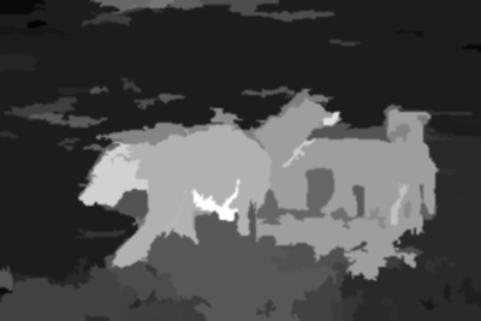} \ & \includegraphics[width=0.065\linewidth]{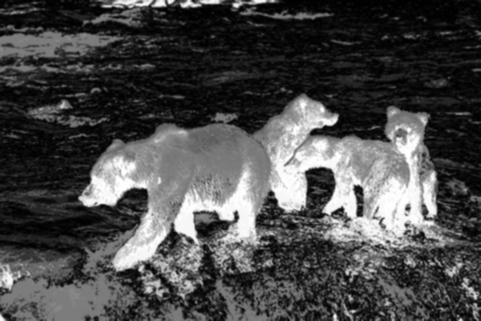} \ &
\includegraphics[width=0.065\linewidth]{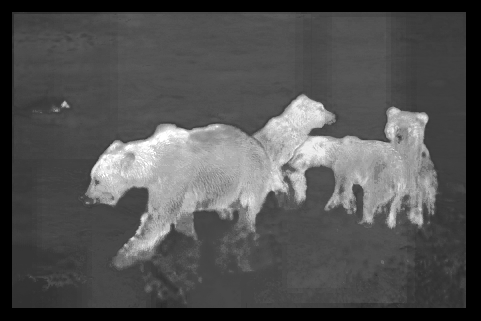} \ &
\includegraphics[width=0.065\linewidth]{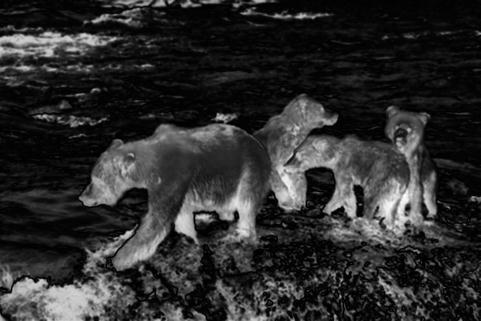} \ &
\includegraphics[width=0.065\linewidth]{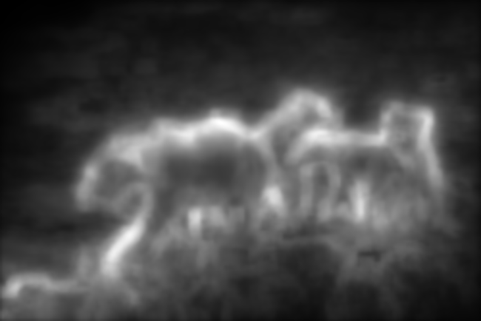} \ &
\includegraphics[width=0.065\linewidth]{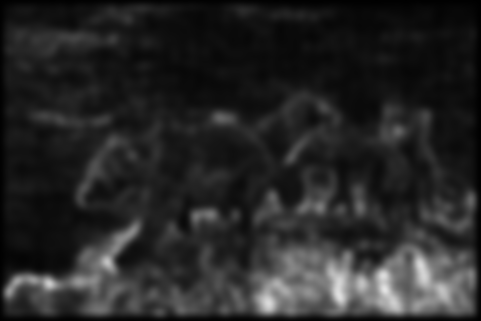} \ &
\includegraphics[width=0.065\linewidth]{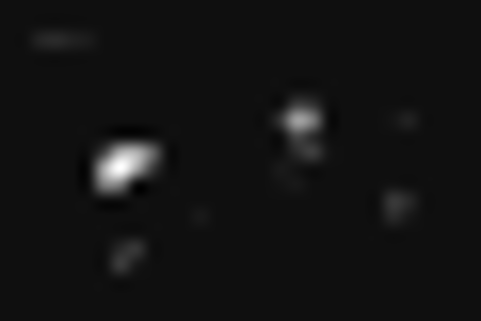}\\
\includegraphics[width=0.065\linewidth]{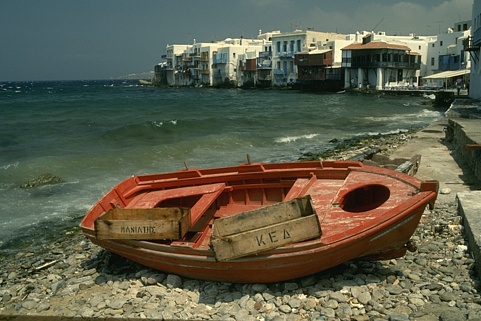} \ &
\includegraphics[width=0.065\linewidth]{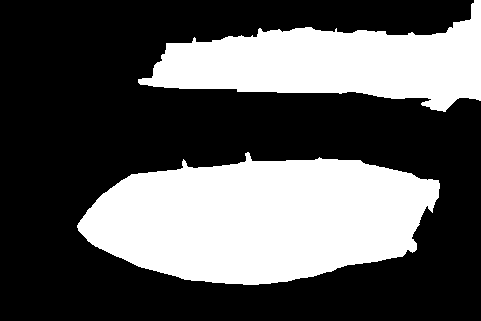} \ &
\includegraphics[width=0.065\linewidth]{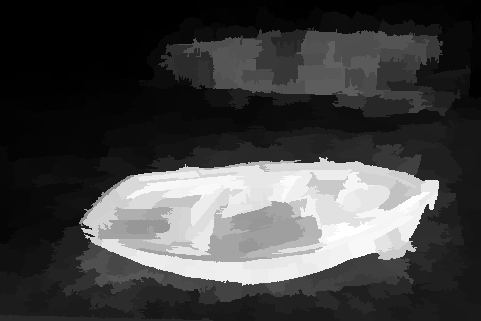} \ &
\includegraphics[width=0.065\linewidth]{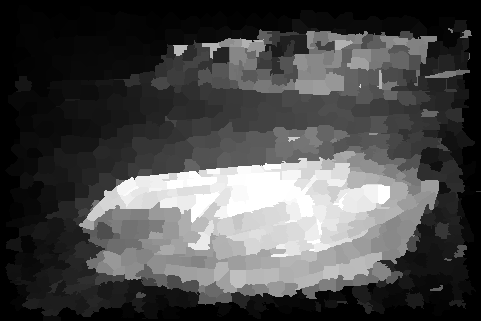} \ &
\includegraphics[width=0.065\linewidth]{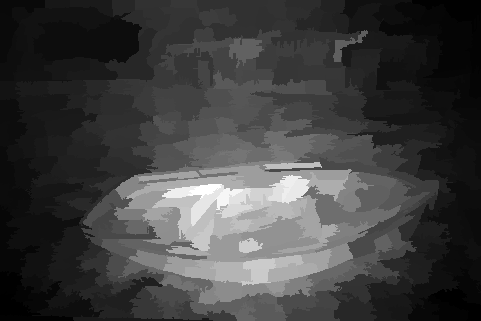} \ &
\includegraphics[width=0.065\linewidth]{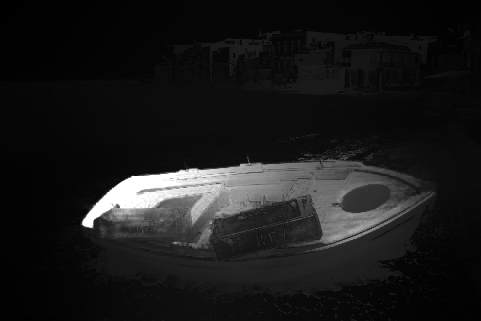} \ &
\includegraphics[width=0.065\linewidth]{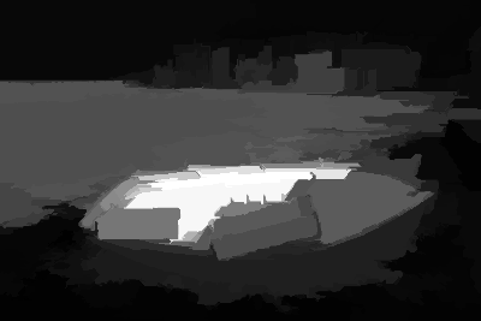} \ &
\includegraphics[width=0.065\linewidth]{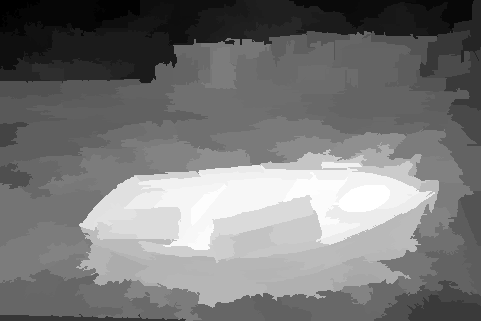} \ &
\includegraphics[width=0.065\linewidth]{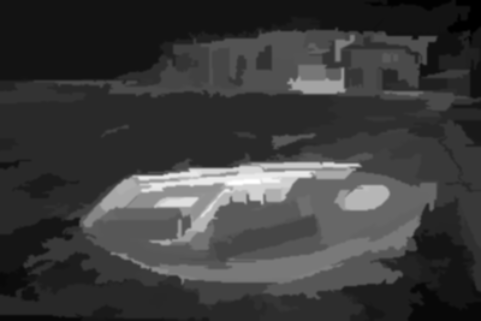} \ & \includegraphics[width=0.065\linewidth]{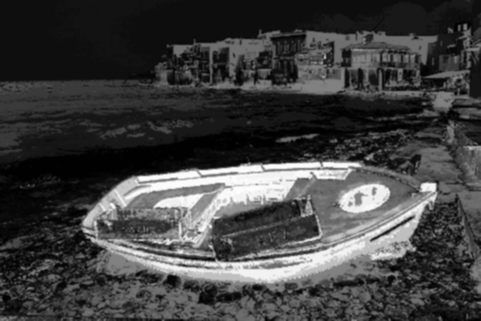} \ &
\includegraphics[width=0.065\linewidth]{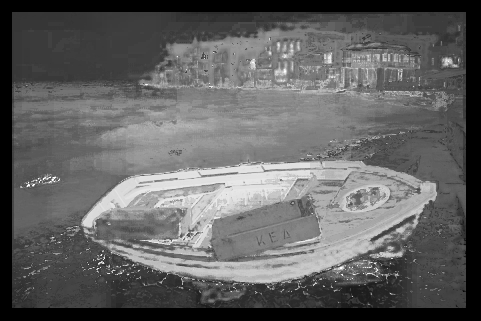} \ &
\includegraphics[width=0.065\linewidth]{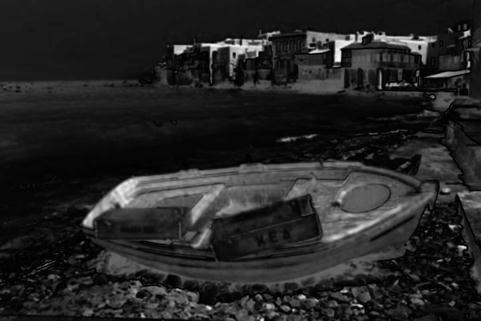} \ &
\includegraphics[width=0.065\linewidth]{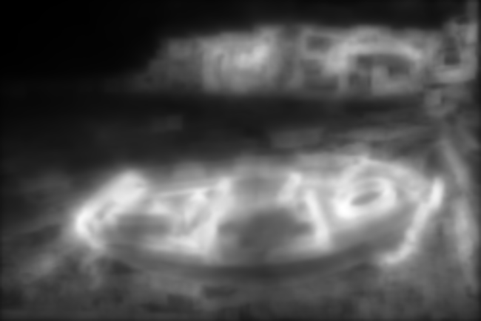} \ &
\includegraphics[width=0.065\linewidth]{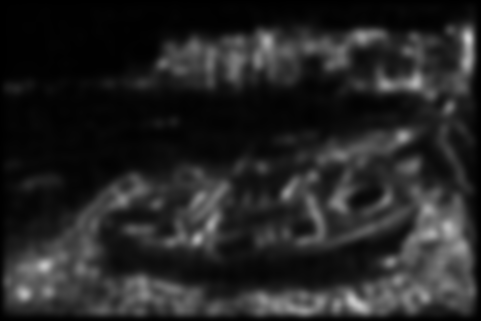} \ &
\includegraphics[width=0.065\linewidth]{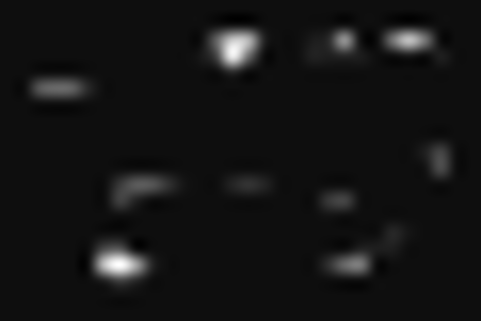}\\
\includegraphics[width=0.065\linewidth]{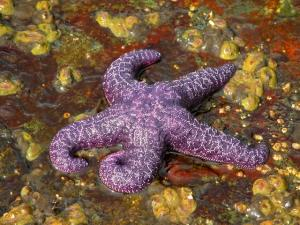} \ &
\includegraphics[width=0.065\linewidth]{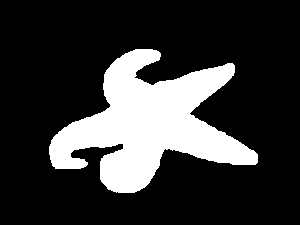} \ &
\includegraphics[width=0.065\linewidth]{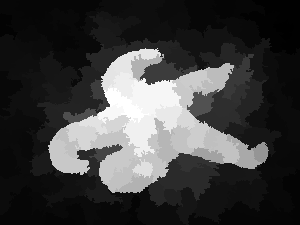} \ &
\includegraphics[width=0.065\linewidth]{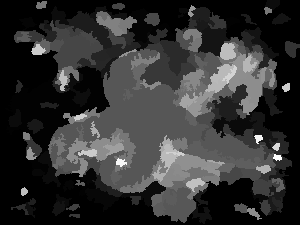} \ &
\includegraphics[width=0.065\linewidth]{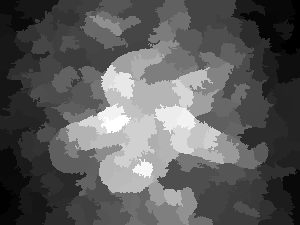} \ &
\includegraphics[width=0.065\linewidth]{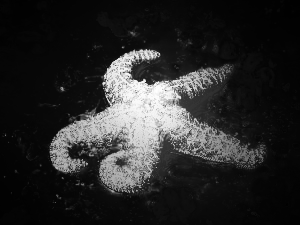} \ &
\includegraphics[width=0.065\linewidth]{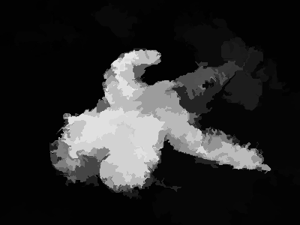} \ &
\includegraphics[width=0.065\linewidth]{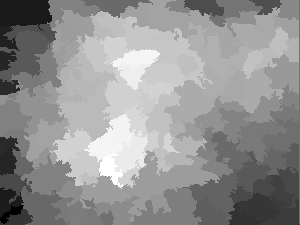} \ &
\includegraphics[width=0.065\linewidth]{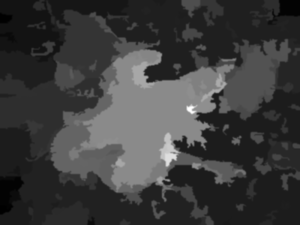} \ & \includegraphics[width=0.065\linewidth]{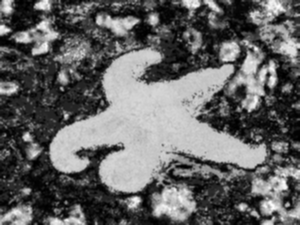} \ &
\includegraphics[width=0.065\linewidth]{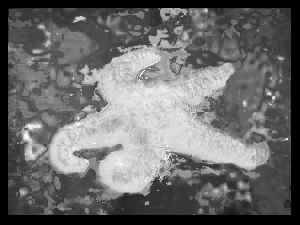} \ &
\includegraphics[width=0.065\linewidth]{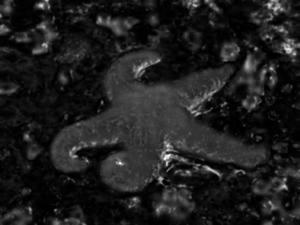} \ &
\includegraphics[width=0.065\linewidth]{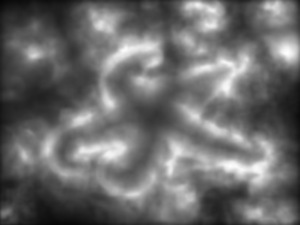} \ &
\includegraphics[width=0.065\linewidth]{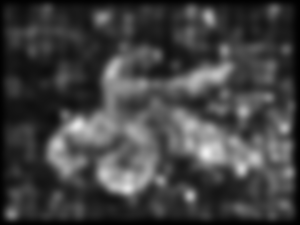} \ &
\includegraphics[width=0.065\linewidth]{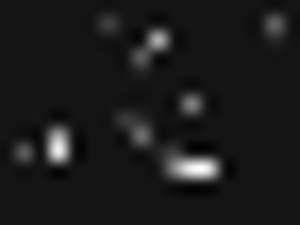}\\
\includegraphics[width=0.065\linewidth]{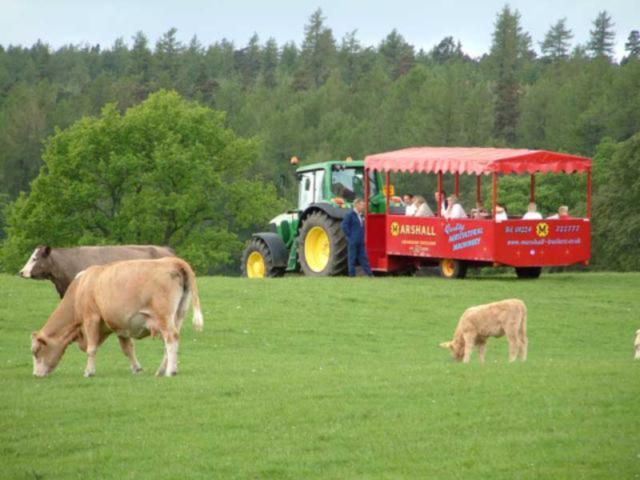} \ &
\includegraphics[width=0.065\linewidth]{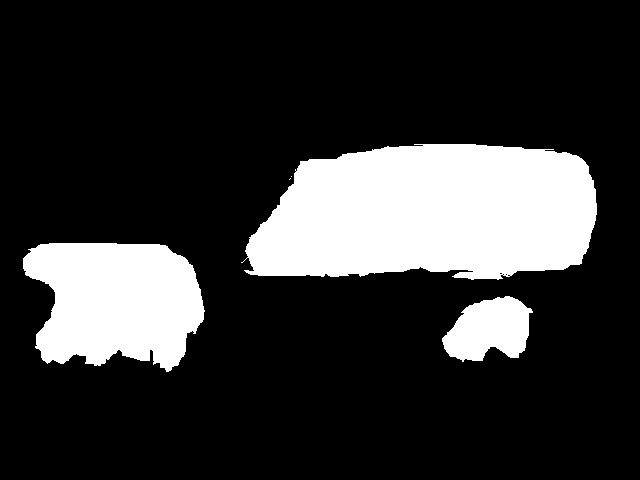} \ &
\includegraphics[width=0.065\linewidth]{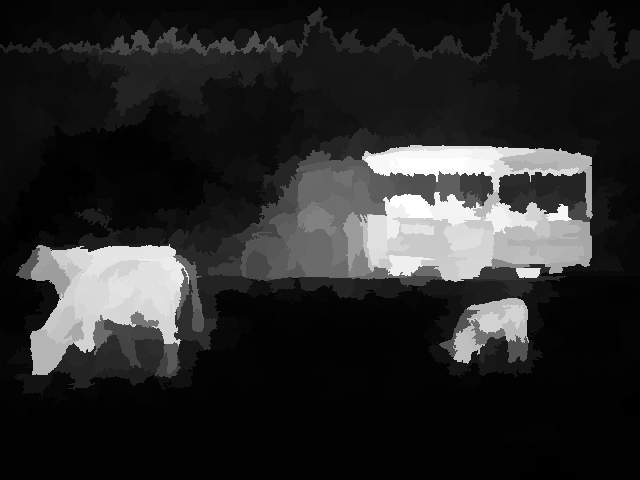} \ &
\includegraphics[width=0.065\linewidth]{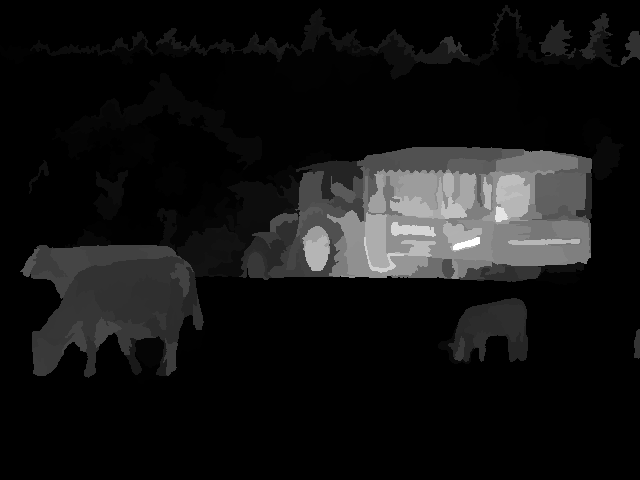} \ &
\includegraphics[width=0.065\linewidth]{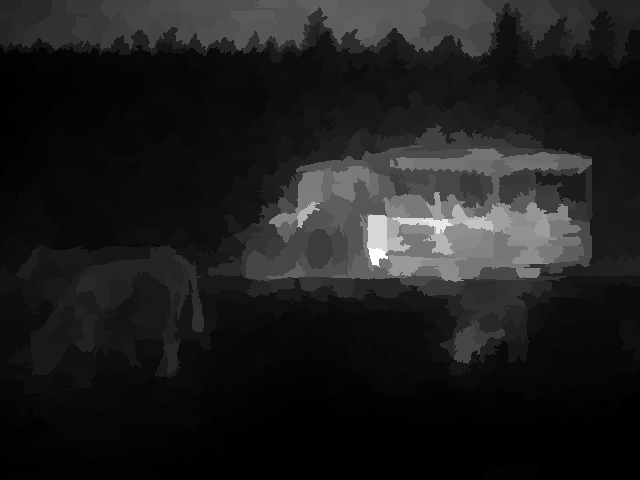} \ &
\includegraphics[width=0.065\linewidth]{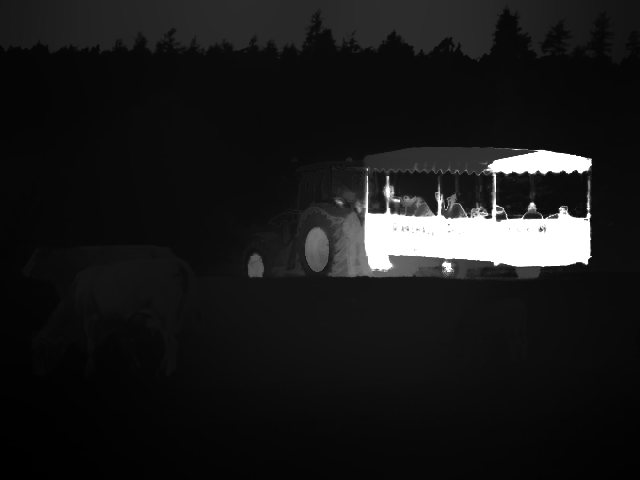} \ &
\includegraphics[width=0.065\linewidth]{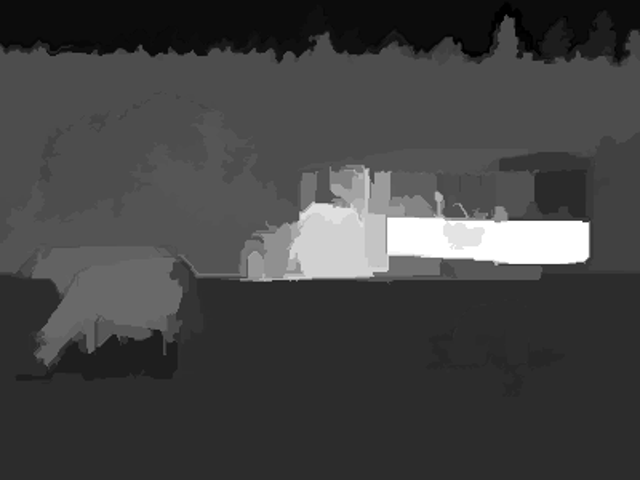} \ &
\includegraphics[width=0.065\linewidth]{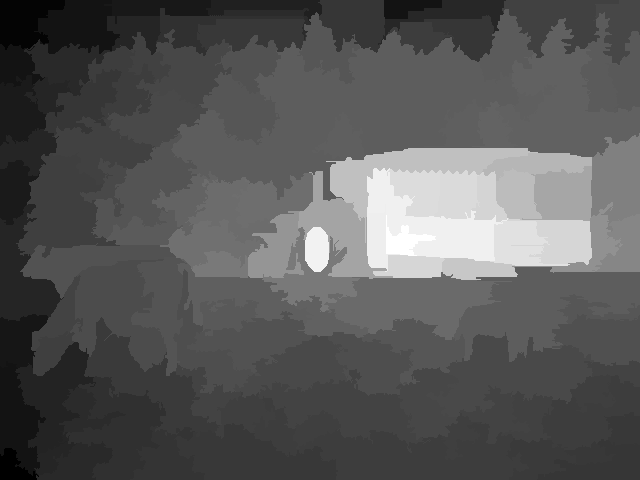} \ &
\includegraphics[width=0.065\linewidth]{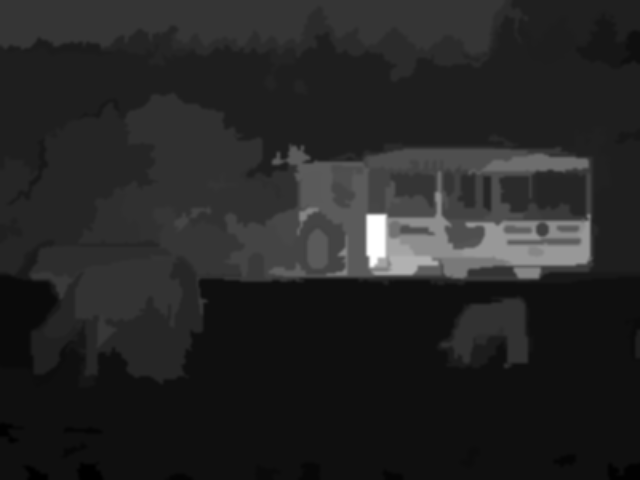} \ & \includegraphics[width=0.065\linewidth]{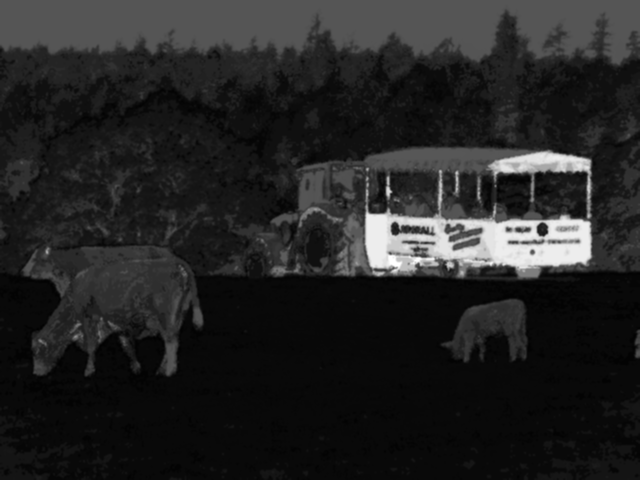} \ &
\includegraphics[width=0.065\linewidth]{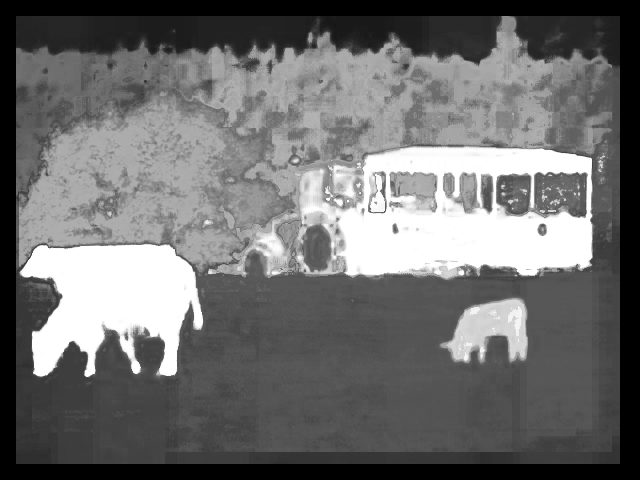} \ &
\includegraphics[width=0.065\linewidth]{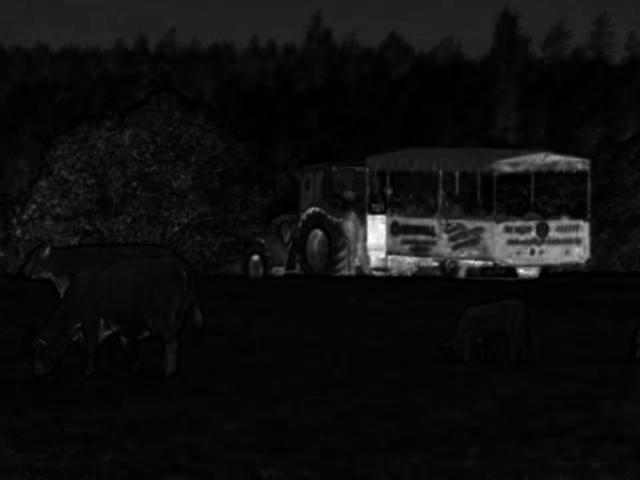} \ &
\includegraphics[width=0.065\linewidth]{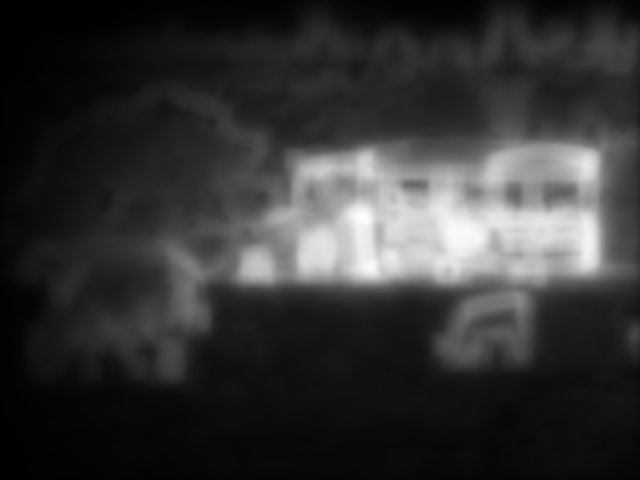} \ &
\includegraphics[width=0.065\linewidth]{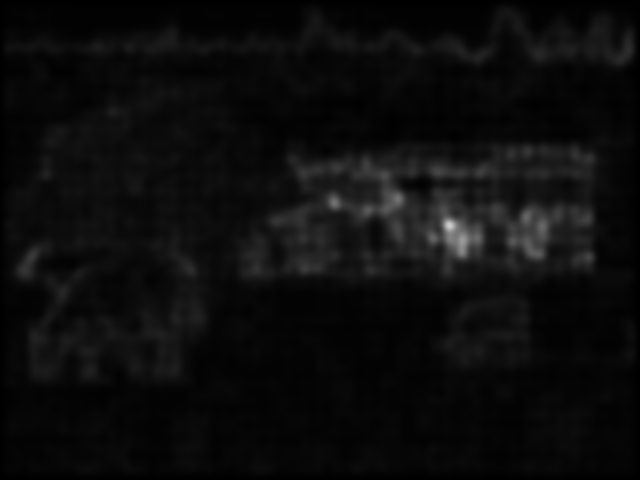} \ &
\includegraphics[width=0.065\linewidth]{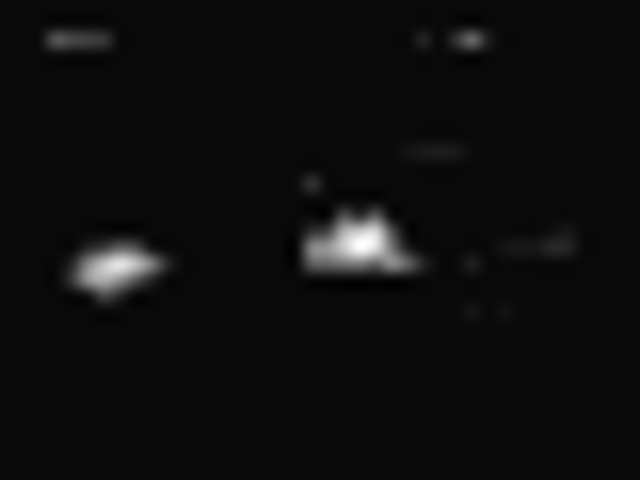}\\
\end{tabular}
}
\end{center}
\caption{Salient object detection examples of all the thirteen approaches on the four datasets:
MSRA-1000 (rows 1-3), SOD (rows 4-6), SED-100 (row 7),
and Imgsal-50 (row 8).
It is clear that our approach obtains
the visually more consistent saliency detection results than
the other competing approaches.
}
\label{fig:salmap}
\end{figure*}

\begin{table*}[t]
\hspace{-0.3cm}
\centering
\scalebox{0.63}{
\begin{tabular}{ l||c|c|c|c|c|c|c|c|c|c|c|c|c }
\hline
 & Ours & GS\rule[-2pt]{2mm}{0.5pt}SP~\cite{WeiWZ012} &  LR~\cite{shen2012unified} & SF~\cite{perazzi2012saliency} &  CB~\cite{jiang2011automatic} & SVO~\cite{ChangLCL11} & RC~\cite{cheng2011global} & HC~\cite{cheng2011global} &  RA~\cite{RahtuKSH10} & FT~\cite{achanta2009frequency} & CA~\cite{GofermanZT10} & ICL~\cite{DBLP:conf/nips/HouZ08} & IT~\cite{itti1998model}\\
\hline
MSRA-1000 & \textbf{0.77}$\pm$0.20 & 0.75$\pm$0.22 & 0.63$\pm$0.25 & 0.67$\pm$0.24 & 0.72$\pm$0.24 & 0.29$\pm$0.24 & 0.52$\pm$0.31
& 0.59$\pm$0.29 & 0.37$\pm$0.33 & 0.50$\pm$0.27 & 0.40$\pm$0.19 & 0.33$\pm$0.19 & 0.17$\pm$0.12\\
\hline
SOD & \textbf{0.40}$\pm$0.22 & 0.38$\pm$0.20 & 0.29$\pm$0.19 & 0.27$\pm$0.20 & 0.31$\pm$0.25 & 0.11$\pm$0.19 & 0.24$\pm$0.23
& 0.22$\pm$0.20 & 0.14$\pm$0.17 & 0.19$\pm$0.17 & 0.27$\pm$0.19 & 0.22$\pm$0.17 & 0.14$\pm$0.11\\
\hline
SED-100 & 0.52$\pm$0.25 & \textbf{0.56}$\pm$0.27 & 0.41$\pm$0.27 & 0.47$\pm$0.27 & 0.52$\pm$0.32 & 0.21$\pm$0.29 & 0.34$\pm$0.31 & 0.37$\pm$0.30 & 0.27$\pm$0.28 & 0.30$\pm$0.26 & 0.35$\pm$0.32 & 0.34$\pm$0.22 & 0.16$\pm$0.14\\
\hline
Imgsal-50 & \textbf{0.69}$\pm$0.18 & 0.65$\pm$0.21 & 0.64$\pm$0.18 & 0.59$\pm$0.22 & 0.64$\pm$0.19 & 0.29$\pm$0.29 & 0.52$\pm$0.25 & 0.45$\pm$0.27 & 0.37$\pm$0.30 & 0.37$\pm$0.19 & 0.47$\pm$0.19 & 0.30$\pm$0.21 & 0.19$\pm$0.10\\
\hline
\end{tabular}}
 \vspace{-0.3cm}
\caption{Quantitative performance of all the thirteen approaches in
VOC overlap scores on the four datasets. Clearly, our approach obtains
the highest VOC overlap score with a low variance
in most cases.}
\label{tal:overlapratio}
\end{table*}

\paragraph{Implementation details}
In the experiments, cost-sensitive SVM saliency detection
on an image is performed
at different scales, each of which corresponds to
a scale-specific image patch size for center-versus-surround
contrast analysis.
The final SVM saliency map is obtained by
averaging the multi-scale saliency detection results.
For computational efficiency, we first choose a
fixed-sized image $8 \times 8$ patch
and then resize the image using different
downsampling rates
to simulate the scale changes.
In addition, each image patch is represented
as a vectorized RGB feature vector.
During the optimization process~\eqref{eq:ls_svm},
the weight $\nu_{1}$ for the center image patch is chosen as
0.5 while the weights $\nu_{k}$ (s.t. $k>1$) for the surrounding image patches
are set to 0.01, as suggested in~\cite{malisiewicz2011ensemble}.
Each superpixel $\bp_{i}$ (referred to Equ.~\eqref{eq:ms_iteration})
is first generated from image over-segmentation,
and then represented by an 8-dimensional feature vector, which is
obtained by averaging the corresponding  color vectors of
all the pixels in the superpixel.
The color vector for each pixel contains four normalized color components $\bc = (l, a, b, h)$ and
their associated elementwise power transforms~\cite{xiaofeng2012discriminatively}
from the LAB and HSV color spaces.
In the experiments, the final saliency detection results
are further refined by graph-based manifold propagation.
We did not carefully tune the
aforementioned parameters in the experiments.
Note that the aforementioned parameters are fixed throughout all the
experiments.

\subsection{Evaluation of our individual approaches}

Here, we evaluate the saliency detection
performance of the proposed approaches
based on three different configurations:
1) using the SVM saliency approach only; 2)
using the hypergraph saliency approach only; and 3)
combining the SVM and hypergraph saliency approaches.
Fig.~\ref{fig:Local_vs_Global} shows their quantitative
results of salient object detection in
the aspect of PR curves.
From Fig.~\ref{fig:Local_vs_Global}, it is clearly
seen that the saliency detection performance
of only using the SVM saliency approach
is significantly enhanced after
combining the hypergraph saliency approach.
The reason is that the hypergraph saliency
approach captures both the internal consistency and strong boundary
properties of salient objects.
By incorporating the SVM saliency approach,
the saliency detection results
of only using the hypergraph saliency approach
are further smoothed,
leading to an
improved saliency detection accuracy.
Therefore, we use the best configuration (i.e., combination of
SVM and hypergraph saliency) for performance evaluations
in the following experiments.

\begin{figure}
\begin{center}
\scalebox{0.9}{
\begin{tabular}{@{}c@{}c@{}c@{}c}
\includegraphics[width=0.25\linewidth]{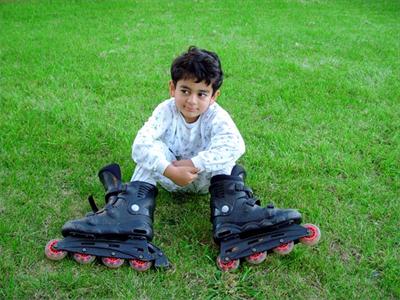} \ &
\includegraphics[width=0.25\linewidth]{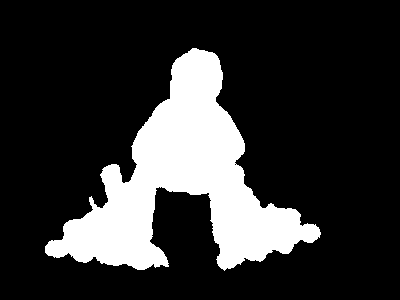} \ &
\includegraphics[width=0.25\linewidth]{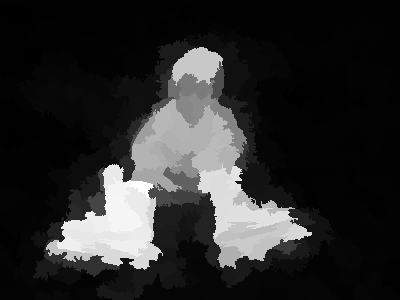} \ &
\includegraphics[width=0.25\linewidth]{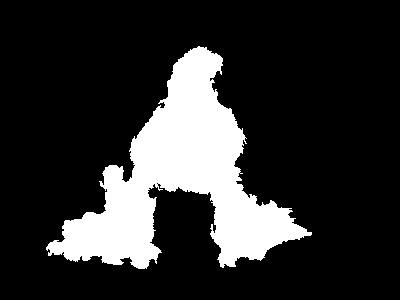} \ \\
\includegraphics[width=0.25\linewidth]{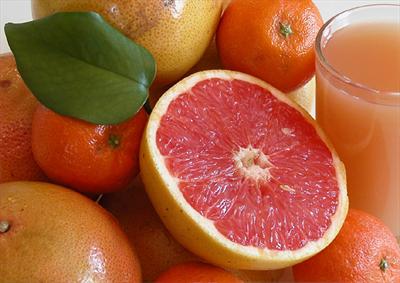} \ &
\includegraphics[width=0.25\linewidth]{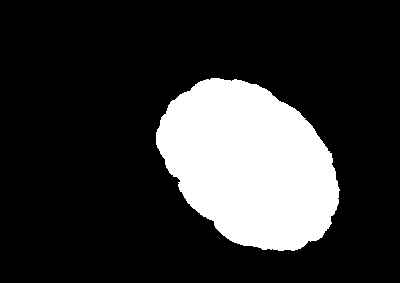} \ &
\includegraphics[width=0.25\linewidth]{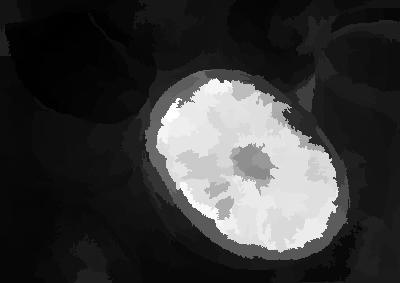} \ &
\includegraphics[width=0.25\linewidth]{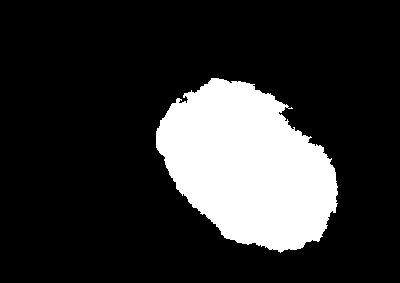} \ \\
\includegraphics[width=0.25\linewidth]{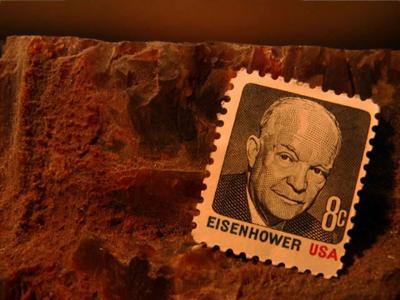} \ &
\includegraphics[width=0.25\linewidth]{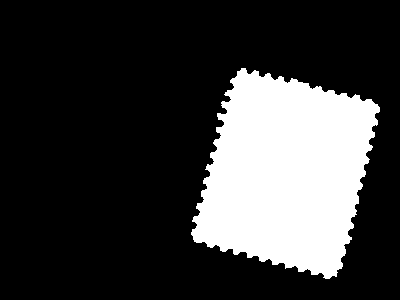} \ &
\includegraphics[width=0.25\linewidth]{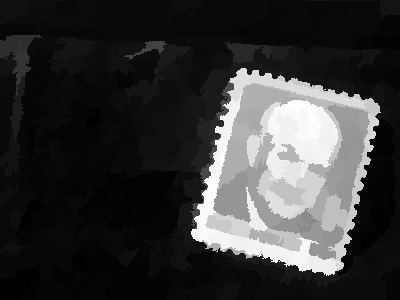} \ &
\includegraphics[width=0.25\linewidth]{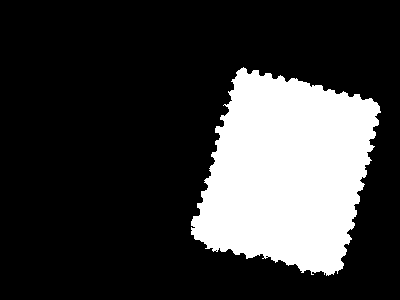} \ \\
\end{tabular}
}
\end{center}
\caption{Examples of salient object segmentation. From left to right: input images, ground truth, saliency maps, segmentation results.
Clearly, our approach obtains  visually consistent segmentation results with ground truth.
}
\label{fig:salient_object_segmentation_sample}
\end{figure}

\subsection{Comparison of saliency detection approaches}
In the experiments,
we qualitatively and quantitatively compare the proposed approach with twelve state-of-the-art approaches,
including
GS\rule[-2pt]{2mm}{0.5pt}SP~\cite{WeiWZ012}, LR~\cite{shen2012unified},
SF~\cite{perazzi2012saliency}, CB~\cite{jiang2011automatic}, SVO~\cite{ChangLCL11},
RC~\cite{cheng2011global}, HC~\cite{cheng2011global}, RA~\cite{RahtuKSH10},
FT~\cite{achanta2009frequency}, CA~\cite{GofermanZT10},
ICL~\cite{DBLP:conf/nips/HouZ08}, and IT~\cite{itti1998model}.
These approaches are implemented using their either
publicly available source code or original saliency detection
results from the authors.

Fig.~\ref{fig:cmparison} shows the quantitative saliency detection performance
of the proposed approach against the twelve competing approaches in
the PR and ROC curves on the four datasets.
From the left half of Fig.~\ref{fig:cmparison}, we see that
the proposed approach achieves the highest precision rate
in most cases when the recall rate is fixed.
Given a fixed false positive rate, the proposed approach
obtains a higher  true positive rate than
the other approaches in most cases, as shown in the right half
of Fig.~\ref{fig:cmparison}.

From Fig.~\ref{fig:cmparison_Fmeasure}, it is observed that the proposed approach
achieves the best F-measure performance on the two popular benchmark
datasets, that is, MSRA-1000 and SOD.
On the SED-100 dataset, GS\rule[-2pt]{2mm}{0.5pt}SP and the proposed
approach obtain the best
results, and the F-measure of
the proposed approach is
slightly lower than GS\rule[-2pt]{2mm}{0.5pt}SP.
On the Imgsal-50 dataset, the proposed approach is
one of the two best approaches,
and achieves a slightly lower
F-measure than CB.
In addition, Fig.~\ref{fig:salmap} shows several
salient object detection examples of all the
thirteen approaches. It is seen from Fig.~\ref{fig:salmap}
that our approach obtain visually more feasible saliency
detection results than the other competing approaches.

Furthermore, Tab.~\ref{tal:overlapratio} shows the corresponding
VOC overlap scores of all the thirteen approaches.
It is seen from Tab.~\ref{tal:overlapratio} that
the proposed approach obtains the highest VOC overlap score
with a low variance in most cases.
Besides, Fig.~\ref{fig:salient_object_segmentation_sample} gives three intuitive
examples of salient object segmentation (i.e., binarization using the
adaptive threshold~\cite{achanta2009frequency}) based on
the proposed approach.
From Fig.~\ref{fig:salient_object_segmentation_sample}, we observe that
the proposed approach achieves the visually consistent
segmentation results with ground truth.

\begin{figure}[t]
\begin{center}
\scalebox{1}{
\begin{tabular}{@{}c@{}c@{}c@{}c@{}c@{}c}
\includegraphics[width=1.3cm,height=1cm]{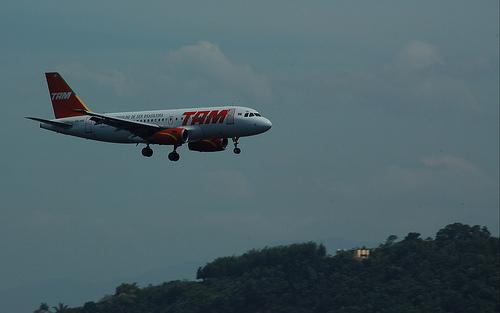} \ &
\includegraphics[width=1.3cm,height=1cm]{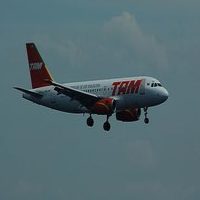} \ &
\includegraphics[width=1.3cm,height=1cm]{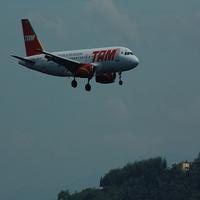} \ &
\includegraphics[width=1.3cm,height=1cm]{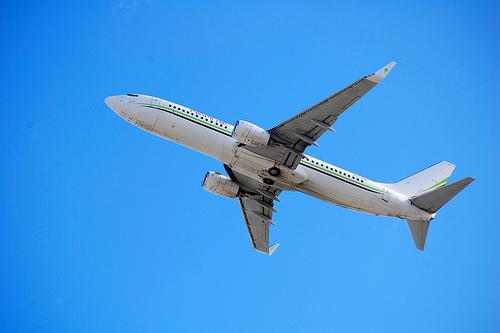} \ &
\includegraphics[width=1.3cm,height=1cm]{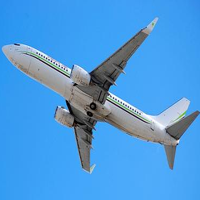} \ &
\includegraphics[width=1.3cm,height=1cm]{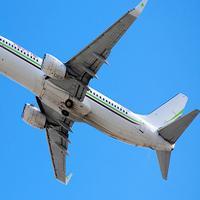} \ \\
\includegraphics[width=1.3cm,height=1cm]{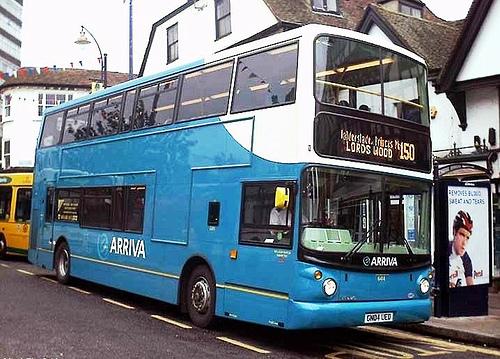} \ &
\includegraphics[width=1.3cm,height=1cm]{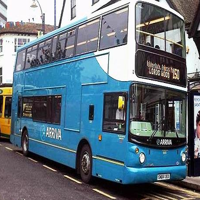} \ &
\includegraphics[width=1.3cm,height=1cm]{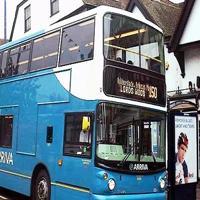} \ &
\includegraphics[width=1.3cm,height=1cm]{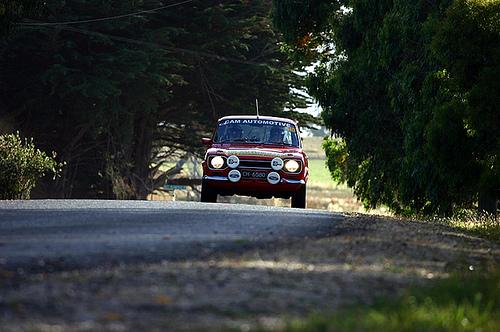} \ &
\includegraphics[width=1.3cm,height=1cm]{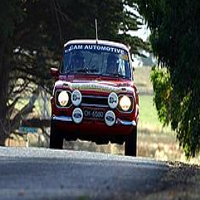} \ &
\includegraphics[width=1.3cm,height=1cm]{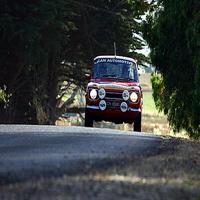} \ \\
\includegraphics[width=1.3cm,height=1cm]{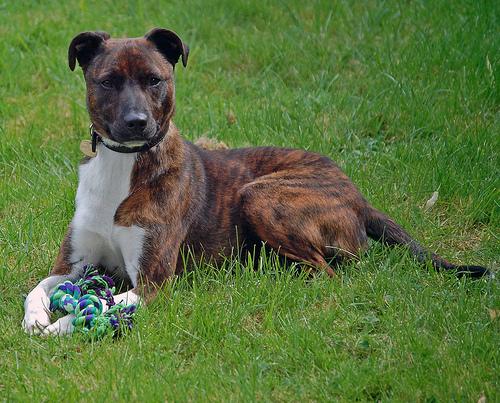} \ &
\includegraphics[width=1.3cm,height=1cm]{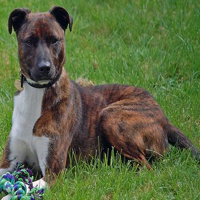} \ &
\includegraphics[width=1.3cm,height=1cm]{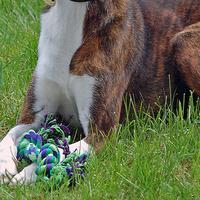} \ &
\includegraphics[width=1.3cm,height=1cm]{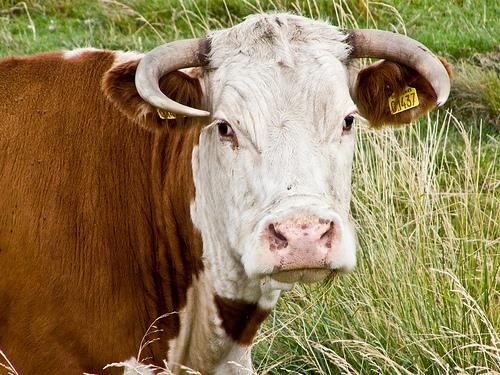} \ &
\includegraphics[width=1.3cm,height=1cm]{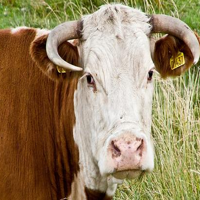} \ &
\includegraphics[width=1.3cm,height=1cm]{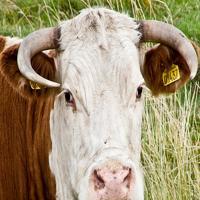} \ \\
\includegraphics[width=1.3cm,height=1cm]{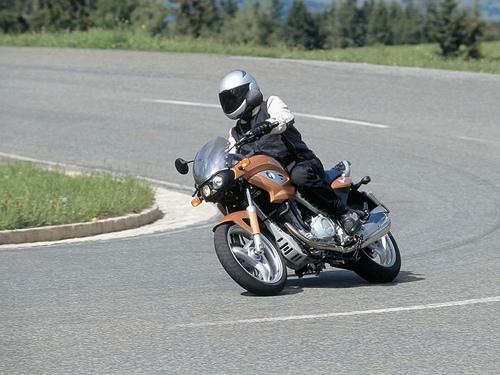} \ &
\includegraphics[width=1.3cm,height=1cm]{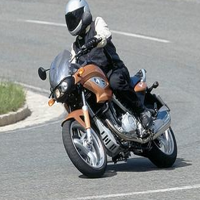} \ &
\includegraphics[width=1.3cm,height=1cm]{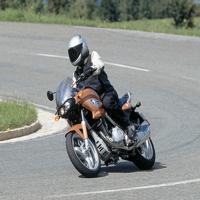} \ &
\includegraphics[width=1.3cm,height=1cm]{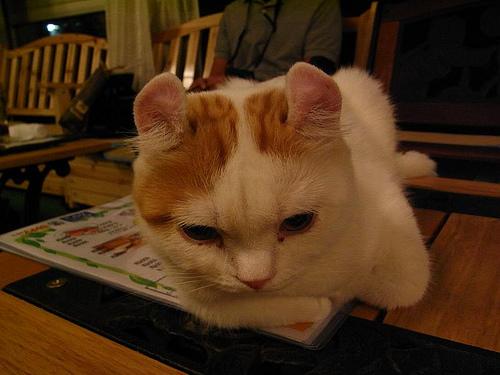} \ &
\includegraphics[width=1.3cm,height=1cm]{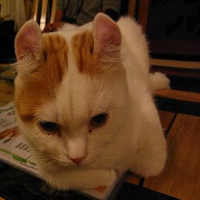} \ &
\includegraphics[width=1.3cm,height=1cm]{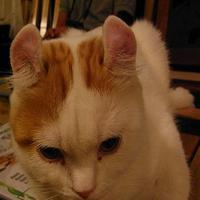} \ \\
\includegraphics[width=1.3cm,height=1cm]{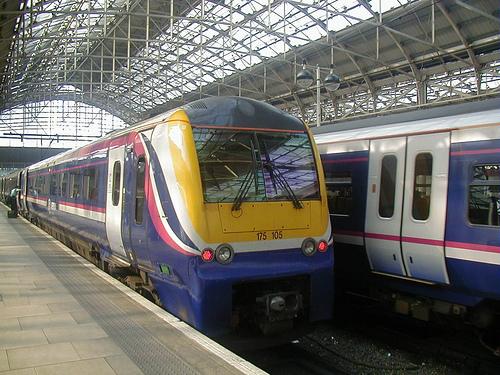} \ &
\includegraphics[width=1.3cm,height=1cm]{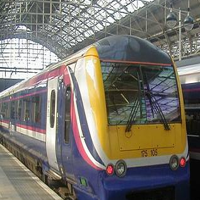} \ &
\includegraphics[width=1.3cm,height=1cm]{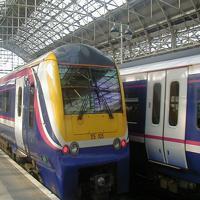} \ &
\includegraphics[width=1.3cm,height=1cm]{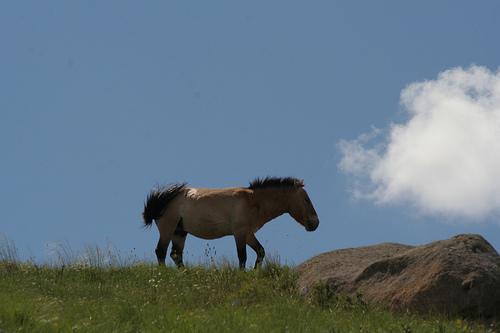} \ &
\includegraphics[width=1.3cm,height=1cm]{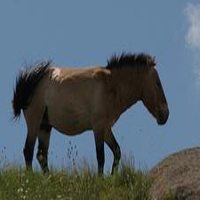} \ &
\includegraphics[width=1.3cm,height=1cm]{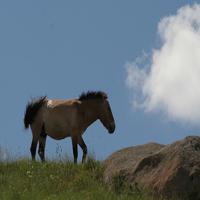} \ \\
\end{tabular}
}
\end{center}
\caption{Qualitative image retargeting performance comparison  between \cite{sun2011scale} and
ours. From left to right: images, our results, results of \cite{sun2011scale}.
Clearly, the performance of our approach is better than that of \cite{sun2011scale}.
}
\label{fig:Retargeting_sample}
\end{figure}

\subsection{Application to image retargeting}

The goal of image retargeting is to reduce
image size while preserving important content.
As shown in~\cite{sun2011scale}, saliency detection
plays an important role in image retargeting.
Following the work of~\cite{sun2011scale}, we directly
replace its saliency detection component with ours
while keeping the other components fixed.
Fig.~\ref{fig:Retargeting_sample} shows some image retargeting examples
of the two approaches (i.e., \cite{sun2011scale} and ours) on
the image retargeting dataset from~\cite{sun2011scale}.
Clearly, our approach obtains  more visually feasible results.
This indicates that our approach is capable of
effectively preserving the intrinsic structural
information on salient objects during image retargeting.

\section{Conclusion}

    In this work, we have proposed two salient object detection approaches
    based on hypergraph modeling and center-versus-surround max-margin
    learning. Specifically, we have designed a hypergraph modeling
    approach that captures the intrinsic contextual saliency
    information on image pixels/superpixels by detecting salient vertices and
    hyperedges in a hypergraph.  Furthermore, we have developed a
    local salient object detection approach based on
    center-versus-surround max-margin learning, which solves an imbalanced
    cost-sensitive SVM optimization problem.
    Compared with the twelve state-of-the-art
    approaches, we have empirically shown that the fusion of our approaches
    is able to achieve more accurate and robust results of salient
    object detection.

\textbf{Acknowledgments}
This work is in part supported by ARC grants
LP120200485 and FT120100969.
Correspondence should be addressed to C. Shen.

{\small
\bibliographystyle{unsrt}
\bibliography{CSRef}
}

\end{document}

%% file: macros.tex
\def\T{{\!\top}}

\def\balpha{ {\boldsymbol{\alpha} } }

\def\bepsilon{{\boldsymbol{\epsilon} } }

\def\bone{ {\bf 1 } }

\def\bx{ {\bf x } }
\def\bp{ {\bf p } }

\def\bw{ {\bf w } }

\def\by{ {\bf y } }
\def\bc{ {\bf c } }

\def\bSigma{ {\boldsymbol{\Sigma} } }

\def\bI{ {\bf I } }

\def\bH{ {\bf H } }
